%% file: main.tex
\documentclass{article}
\usepackage[preprint]{neurips_2026}
\input{preamble}

\title{UI-UG 2: Advancing Mobile UI Understanding and Generation with Fine-Grained Supervision and Reinforcement Learning}
\input{authors}
\date{}

\begin{document}
\maketitle
\input{sections/00_abstract}
\input{sections/01_introduction}
\input{sections/02_related_work}
\input{sections/03_data}
\input{sections/04_method}
\input{sections/05_experiments}
\FloatBarrier
\input{sections/06_discussion}
\input{sections/07_conclusion}

\clearpage
\bibliographystyle{plainnat}
\bibliography{references}

\clearpage
\appendix
\input{appendices/01_taxonomy}
\input{appendices/02_dataset}
\input{appendices/03_dsl_examples}
\input{appendices/04_prompts}
\input{appendices/05_dense_evaluation}
\input{appendices/06_understanding}
\input{appendices/07_generation}
\input{appendices/08_efficiency}
\end{document}

%% file: preamble.tex
\usepackage[utf8]{inputenc}
\usepackage[T1]{fontenc}
\usepackage{amsmath,amssymb}
\usepackage{graphicx}
\usepackage{booktabs,array}
\usepackage{microtype}
\usepackage{xcolor}
\usepackage{placeins,needspace}
\usepackage{xurl}
\usepackage[hidelinks]{hyperref}
\ifdefined\pdfminorversion
\fi
\ifdefined\XeTeXversion
  
\fi
\hypersetup{
  pdftitle={UI-UG 2: Advancing Mobile UI Understanding and Generation},
  pdfsubject={Mobile UI understanding and structured UI generation},
  pdfkeywords={mobile UI, multimodal models, grounding, UI generation, GRPO}
}
\setcitestyle{numbers,square,comma}

%% file: authors.tex
\author{%
  \textbf{Hao Yang}\qquad
  \textbf{Shuoyang Liu}\qquad
  \textbf{Xiaoyu Lin}\\[2pt]
  \textbf{Zhou Fang}\qquad
  \textbf{Maji Huang}\qquad
  \textbf{Zhaosong Huang}\\[2pt]
  \textbf{Ruichao Mao}\qquad
  \textbf{Teng Guo}\qquad
  \textbf{Hai Rao}\\[6pt]
  \normalfont Ant Group
}
\hypersetup{pdfauthor={Hao Yang, Shuoyang Liu, Xiaoyu Lin, Zhou Fang, Maji Huang, Zhaosong Huang, Ruichao Mao, Teng Guo, Hai Rao}}

%% file: sections/00_abstract.tex
\begin{abstract}
Mobile UI understanding and generation require both fine-grained component perception and reliable structured synthesis. Inspired by the unified direction introduced by UI-UG~\cite{uiugv1}, we present \textbf{UI-UG 2}, a compact 4B-parameter multimodal large language model (MLLM) that advances mobile \textbf{UI} \textbf{U}nderstanding and \textbf{G}eneration with broader task coverage, a front-end-aligned taxonomy, and online render-aware reinforcement learning. UI-UG 2 constructs a multi-source understanding corpus from RICO, VINS, Android in the Wild, and Android Control, annotated with an 80-category two-level UI taxonomy aligned with front-end component semantics, and pairs it with over 5,000 requirement--screenshot--JSON tuples for progressive structured UI generation through a domain-specific language (DSL). For understanding, UI-UG 2 introduces decomposed rewards for category prediction, OCR fidelity, color recognition, dense grounding, and GUI grounding.  Unlike the offline Direct Preference Optimization (DPO)-style generation optimization in UI-UG, UI-UG 2 optimizes generation through end-to-end online Group Relative Policy Optimization (GRPO): sampled DSL outputs are rendered with a Playwright-based server-side pipeline and evaluated using LLM-as-a-Judge and contrastive preference ranking. Experiments show that UI-UG 2 achieves the strongest referring and dense-grounding results among the evaluated models on our 80-category benchmark, and achieves the best composite generation score among compared models at a compact scale (4B). More importantly, we observe a consistent understanding-to-generation transfer: adding fine-grained UI perception supervision improves generation quality under a fixed training budget, suggesting that component-level semantics provide a shared representational foundation for both tasks.
\end{abstract}

%% file: sections/01_introduction.tex
\begin{figure}[htbp]
\centering

\includegraphics[width=1.0\linewidth]{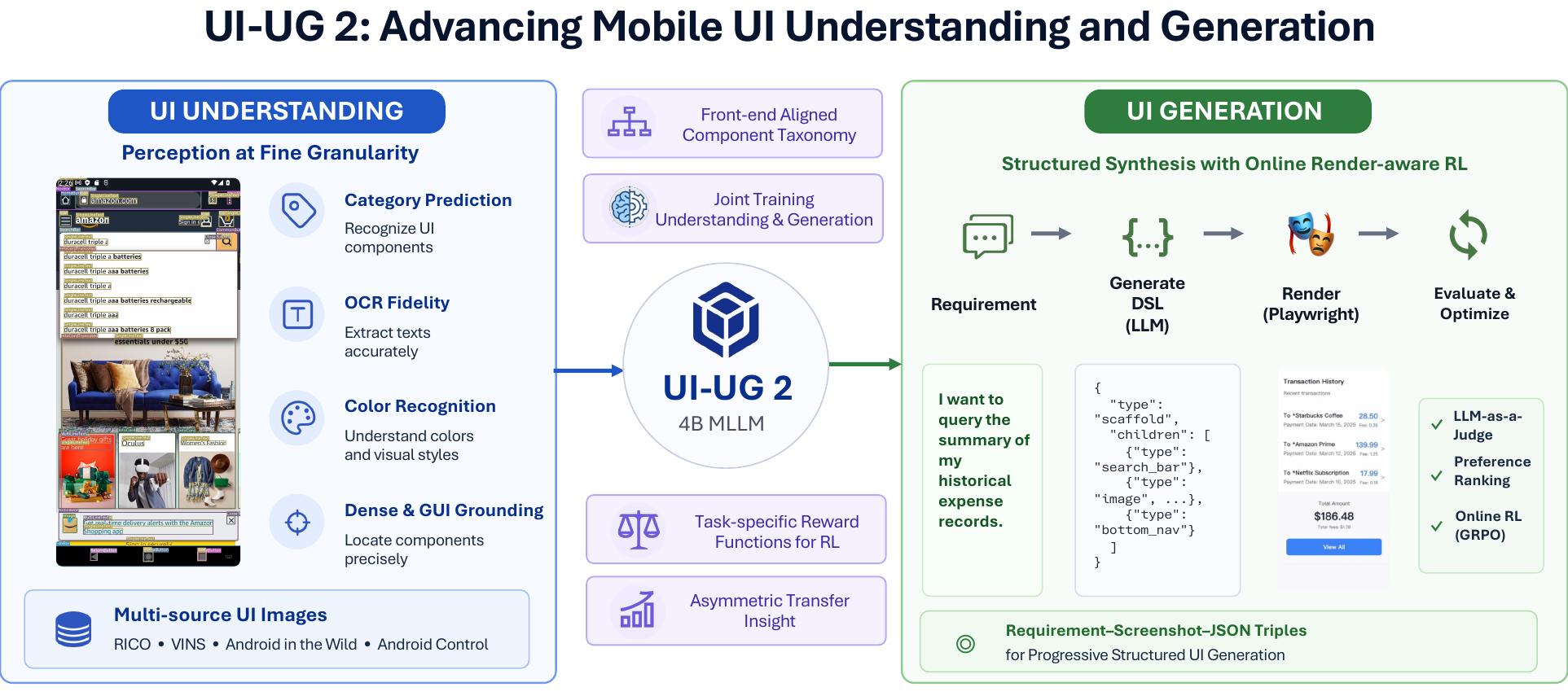}

\caption{Overview of UI-UG 2. UI-UG 2 unifies fine-grained mobile UI understanding and structured UI generation with a front-end-aligned taxonomy, joint training, task-specific rewards, and online render-aware optimization.}

\label{fig2}
\end{figure}

\section{Introduction}
Mobile user interfaces are the primary bridge between users and digital services. As applications grow in complexity, the ability to accurately \emph{understand} and \emph{generate} UIs underpins a broad spectrum of downstream applications, from screen parsing and GUI agent interaction to personalized interface synthesis and LLM-driven front-end development. These two capabilities, however, have largely evolved along separate tracks, with separate models, separate data pipelines, and separate evaluation protocols.

General-purpose MLLMs such as GPT~\cite{gpt5}, Claude~\cite{Claude}, Gemini~\cite{gemini}, and Qwen-VL~\cite{bai2025qwen3vltechnicalreport} can nominally perform both tasks, yet their accuracy remains insufficient for precision-critical mobile scenarios. Mobile screens are fundamentally different from natural images: they are dense compositions of small, semantically loaded components---buttons, icons, text fields, navigation bars---whose interaction semantics require fine-grained categorical reasoning. Distinguishing a \emph{CloseButton} from a \emph{ReturnButton}, or a \emph{SearchBar} from a generic \emph{TextInput}, is essential for reliable automation and high-quality generation, but lies beyond the granularity that general-purpose models reliably achieve. Specialized approaches address one side of the problem---Ferret-UI~\cite{ferret-ui} and Ferret-UI2~\cite{ferret-ui2} for grounding and referring, DCGen~\cite{dcgen} and DeclarUI~\cite{declareUI} for structured generation---while UI-UG~\cite{uiugv1} took an initial step toward a unified model. However, important limitations remain: its labels were not explicitly organized around front-end component semantics, its understanding coverage focused mainly on referring and dense grounding, its generation optimization relied on offline Direct Preference Optimization (DPO) over a fixed preference dataset, and its evidence for understanding-to-generation transfer remained preliminary.

We present \textbf{UI-UG 2}, a compact 4B MLLM for mobile UI Understanding and Generation, with an overview illustrated in Figure~\ref{fig2}. UI-UG 2 keeps the central goal of unified UI understanding and generation while strengthening the data, training, and analysis pipeline. On the data side, we construct a large-scale understanding benchmark from four public sources (RICO, VINS, Android in the Wild, and Android Control), covering over 30{,}000 screenshots with 80 fine-grained element categories organized in a two-level hierarchy aligned to front-end conventions, enriched with element-level, action-level, OCR, and color annotations. We pair this with a curated generation corpus of over 5{,}000 requirement--screenshot--JSON triples whose compact domain-specific language (DSL) supports cross-platform rendering and progressive streaming.

On the optimization side, we adopt a two-stage pipeline of supervised fine-tuning followed by task-specific Group Relative Policy Optimization (GRPO). For understanding, we design decomposed rewards targeting format compliance, category prediction, OCR fidelity, color recognition, dense grounding, and GUI grounding. For generation, UI-UG 2 performs online render-aware RL: each sampled DSL is rendered by a Playwright-based server-side pipeline, then evaluated through LLM-as-a-Judge and Contrastive Preference Ranking (CPR). This closes the loop between structured generation, actual rendering, and reward feedback, making optimization better aligned with deployment-time UI quality.

Experiments demonstrate that UI-UG 2 achieves the strongest referring and dense-grounding results among the compared models on our benchmark, strong GUI grounding on ScreenSpot mobile benchmarks, and the best overall generation quality among compared models at a compact 4B scale. Through data-mixing ablations, reward transfer analysis, and attention probing, we further observe an asymmetric synergy: \emph{understanding supervision improves generation quality at an appropriate data-mixing ratio}. This finding---that component-level perceptual representations form a transferable foundation for generative tasks---offers a concrete design principle for future mobile UI systems.

\Needspace{5\baselineskip}
Our contributions are:
\begin{itemize}

    \item \textbf{UI-UG 2.} We propose UI-UG 2, a compact unified MLLM that jointly supports mobile UI understanding and generation, advancing the UI-UG direction~\cite{uiugv1} with broader task coverage, stronger efficiency, and more rigorous optimization. UI-UG 2 achieves the best results among the compared models on our newly constructed referring and dense-grounding benchmark, and obtains the best composite generation score among all compared baselines despite using only a compact 4B backbone.

    \item \textbf{Front-end-aligned data and tasks.} We contribute an 80-category hierarchical UI understanding dataset with taxonomy aligned to front-end conventions, supporting referring, dense grounding, and GUI grounding, together with a high-quality UI generation corpus based on a compact, renderable JSON DSL.

    \item \textbf{Task-specific online GRPO.} We introduce decomposed reward functions for category, OCR, color, dense grounding, GUI grounding, format validity, renderability, LLM-judged UI quality, and CPR-based preference, enabling online render-aware optimization for generation.

    \item \textbf{Understanding-to-generation synergy.} We provide systematic evidence that fine-grained UI perception consistently benefits generation quality, supported by data-mixing ablations, reward transfer analysis, and attention probing.
\end{itemize}

%% file: sections/02_related_work.tex
\section{Related Work}

\noindent\textbf{UI Understanding.}
UI understanding is central to screen parsing, interface analysis and intelligent automation. Early advances were enabled by datasets such as VINS~\cite{bunian2021vins}, ERICA~\cite{erica}, WebSRC~\cite{chen2021websrc}, and RICO~\cite{rico}. Methodologically, the field has evolved from relying on structured signals such as Document Object Models (DOMs) or view hierarchies~\cite{cai2003vips} to parsing interfaces directly from pixels~\cite{wu2021screenparsing, lee2023pix2struct, omniparser, omniparserv2}. More recently, MLLM-based models such as Ferret-UI~\cite{ferret-ui} and Ferret-UI2~\cite{ferret-ui2} have unified multiple UI understanding tasks, while agent-oriented grounding models such as SeeClick~\cite{cheng2024seeclick}, ShowUI~\cite{lin2025showui}, UI-TARS~\cite{qin2025ui-tars}, GUI-Owl~\cite{guiowl} and UI-Venus~\cite{gu2025ui} have improved fine-grained localization on benchmarks such as ScreenSpot Series~\cite{cheng2024seeclick,li2025screenspotpro}. Compared with prior work, we introduce a richer front-end-aligned taxonomy and support a more comprehensive set of fine-grained UI understanding tasks.

\noindent\textbf{UI Generation.}
Automated UI generation aims to translate visual designs or user requirements into executable interface code. Early approaches such as pix2code~\cite{beltramelli2018pix2code}, Sketch2Code~\cite{jain2019sketch2code}, and REMAUI~\cite{nguyen2015remaUI} focused on image-to-code prediction or explicit component reconstruction. More recent MLLM-based methods, including DCGen~\cite{dcgen} and DeclarUI~\cite{declareUI}, improve structured UI generation through stronger multimodal reasoning. Benchmarks such as Web2Code~\cite{yun2024web2code} and LiveCodeBench~\cite{jain2024livecodebench} have also helped standardize evaluation. In contrast, we study MLLM-based UI generation from both user requirements and reference images, with additional support for interactive prototyping.

\noindent\textbf{Unified UI Understanding and Generation.}
UI-UG~\cite{uiugv1} first explored jointly training mobile UI understanding and structured UI generation in a single MLLM, using a renderable DSL and task-aware post-training. This line of work suggests that UI perception and UI synthesis can share a common representation, but several questions remain open, including how to define front-end-aligned supervision, how to optimize generation under rendered visual feedback, and how to analyze transfer between understanding and generation more systematically.

\noindent\textbf{Reinforcement Learning and Multi-task MLLMs.}
Post-training of MLLMs has evolved from RLHF~\cite{christiano2017rlhf} to preference optimization such as DPO~\cite{dpo} and efficient policy optimization such as GRPO~\cite{deepseek-r1}, with recent multimodal extensions including Visual-RFT~\cite{liu2025visual-rft}, R1-VL~\cite{zhang2025r1-vl}, and VLM-R1~\cite{shen2025vlm-r1}. We apply GRPO to UI understanding and generation with task-specific rewards for structured perception and renderable synthesis. Related multi-task studies show that heterogeneous objectives can cause imbalance and asymmetric transfer~\cite{liu2019multitask,vandenhende2021multitask,rao2026understanding}, motivating our sequential optimization and transfer analysis.

%% file: sections/03_data.tex
\section{Data and Structured Representation}\label{sec:data}
\subsection{Understanding Corpus and Annotation}

UI-UG 2 uses structured supervision for both UI understanding and generation. For understanding, we curate over \textbf{30,000} deduplicated mobile screenshots from RICO~\cite{rico}, VINS~\cite{bunian2021vins}, Android in the Wild~\cite{rawles2023androidinthewild}, and Android Control~\cite{li2024androidcontrol}. The data combine element-level supervision from RICO/VINS with action-level interaction semantics from AITW/AC, which we convert into instruction-conditioned GUI grounding samples.

We annotate screenshots with an \textbf{80-category} two-level taxonomy (\textbf{9 parent classes + 71 child classes}) aligned with common front-end component semantics, drawing on naming and interaction patterns found in libraries such as Ant Design~\cite{ant_design}, Material UI~\cite{material_ui}, and Vant~\cite{vant}. A human-in-the-loop bootstrapping pipeline expands an annotated seed set with a post-trained detection model~\cite{yolov8_ultralytics}, OCR, and color labels, followed by human refinement; agreement reaches Cohen's $\kappa > 0.82$ on a 500-image held-out set. Each element receives a category and a bounding box; text-like elements additionally receive OCR text and text-color annotations. The taxonomy distinguishes functional roles (e.g., \texttt{ReturnButton} versus \texttt{CloseButton}), component states, and truncated elements near screen boundaries. The CLIP-deduplicated corpus draws approximately 7.5K screenshots from each source. The complete taxonomy and distribution analyses appear in Appendices~\ref{app:taxonomy} and~\ref{app:dataset_stats}.

\subsection{Generation Corpus and Rendering Pipeline}

\begin{figure}[htbp]
\centering
\includegraphics[width=0.78\linewidth]{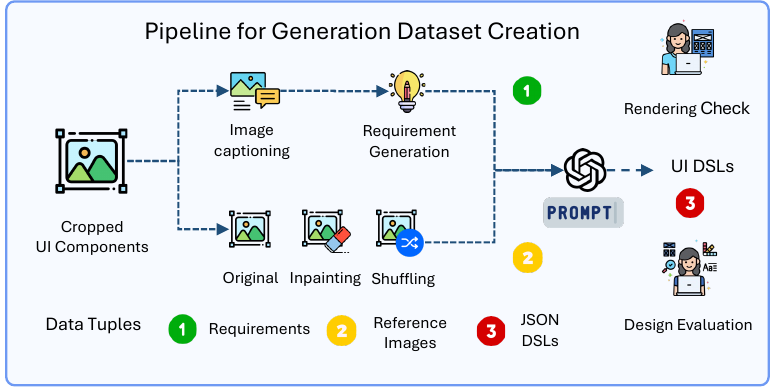}
\caption{Pipeline for constructing the UI generation dataset. Each tuple consists of a requirement, a reference image, and a corresponding renderable UI JSON DSL.}
\label{fig3}
\end{figure}

For generation, we retain the renderable JSON DSL used by UI-UG~\cite{uiugv1} instead of raw HTML/CSS, as it offers a compact, structured, and cross-platform interface for progressive UI rendering, following the broader industrial trend of DSL-to-UI systems such as Google/A2UI~\cite{google_a2ui} and Vercel/json-renderer~\cite{vercel_json_renderer}.
 We improve the corpus with stronger MLLM distillation, richer requirements, and stricter rendering-based filtering. The final generation data contain over \textbf{5,000} high-quality requirement--image--JSON tuples; the DSL reduces token usage by \textbf{16.7\%} on average compared with equivalent HTML. Figure~\ref{fig3} shows the corpus construction pipeline. Each accepted tuple couples a natural-language UI requirement, a reference screenshot, and a JSON specification that passes schema and rendering checks. The shared renderer also executes sampled DSL outputs during generation GRPO.

\subsection{Renderable JSON Representation}

Our JSON-based DSL describes UI content and styling features. The basic node attributes include:

\begin{itemize}
    \item \texttt{type}: Node type, either \texttt{"Component"} or \texttt{"Tag"}.
    \item \texttt{name}: If \texttt{type} is \texttt{"Component"}, the component name from the available components list; if \texttt{"Tag"}, an HTML tag name.
    \item \texttt{className}: Tailwind CSS class names using relative layouts. Width/height can be absolute (e.g., \texttt{w-[100px]}) or relative (e.g., \texttt{w-full}).
    \item \texttt{params}: Content features of the node. Must include \texttt{bindType} (\texttt{"Static"} or \texttt{"Data"}) to indicate whether the value is hardcoded or dynamically bound.
    \item \texttt{children}: A list of child nodes.
\end{itemize}

The representation separates component identity, layout/styling, content binding, and child structure. It supports progressive rendering as the model produces the structured specification. Figure~\ref{fig:dsl_complex} in Appendix~\ref{app:dsl} gives a complete rendered example. A minimal button can be expressed as:
\begin{quote}\small\ttfamily
\{\quad"type": "Tag", "name": "button",\\
\quad"className": "rounded-xl bg-blue-600 px-4 py-2",\\
\quad"params": \{"textContent":\\
\qquad\{"bindType": "Static", "value": "Continue"\}\}\quad\}
\end{quote}

%% file: sections/04_method.tex
\section{Method}

Given the data and representation in Section~\ref{sec:data}, we train UI-UG 2 with unified supervised fine-tuning followed by task-specific GRPO. Compared with UI-UG~\cite{uiugv1}, UI-UG 2 uses a smaller 4B backbone and replaces offline generation preference learning with online render-aware GRPO.

\subsection{Supervised Fine-tuning}

\textbf{SFT Data Construction.}
We construct a unified SFT dataset for \textbf{referring}, \textbf{grounding}, and \textbf{generation}, with all targets expressed in JSON for reliable parsing. Referring predicts category, OCR text, and color from a given box; grounding covers dense full-screen localization and instruction-conditioned captioned grounding; generation maps a requirement and optional reference screenshot to a structured JSON DSL. Dense-grounding boxes are sorted by category frequency and top-left coordinates to expose repeated layout regularities.

\noindent\textbf{Model Fine-tuning.}
We initialize \textbf{UI-UG 2-4B-SFT} from Qwen3-VL-4B-Instruct, freeze the ViT encoder, and train the LLM and VL adapter. The training mixture and hyperparameters are reported in Section~\ref{sec:training}.

\subsection{Reinforcement Learning via GRPO}

\label{sec:grpo}

While SFT provides a strong initialization, it does not directly optimize task-specific quality metrics. We therefore further optimize UI-UG 2 with GRPO, using rewards tailored to UI understanding and generation.

\noindent\textbf{UI Understanding Rewards.}

\noindent\textbf{(1) Referring Reward.}
The referring reward decomposes element description into category, OCR, and color prediction:
\begin{equation}
  R_{\text{ref}} = R_{\text{fmt}} + R_{\text{cls}}
  + \mathbf{1}[t^* \in \mathcal{T}_{\text{txt}}]
  \cdot (R_{\text{ocr}} + R_{\text{col}})
  \label{eq:ref_reward}
\end{equation}

where $t^*$ is the target element type, $R_{\text{fmt}}\in\{0,1\}$ and $R_{\text{cls}}\in\{0,1\}$
denote format validity and category correctness, and
$\mathcal{T}_{\text{txt}} = \{$SingleLineText, MultiLineText, MoneyLineText$\}$
activates OCR/color rewards only for text elements.

The \textbf{OCR reward} uses normalized edit distance (NED) to provide smooth partial credit:
\begin{equation}
  R_{\text{ocr}} = 1 - \frac{\text{ED}(\hat{s}, s^*)}
  {\max(|\hat{s}|, |s^*|)}
  \label{eq:ocr_reward}
\end{equation}
where $\hat{s}$ and $s^*$ denote predicted and ground-truth strings; if both are empty, $R_{\text{ocr}}=1$. NED provides smooth partial credit for common UI OCR ambiguities.

The color reward is computed in perceptually uniform CIE~L*a*b* space:
\begin{equation}
  R_{\text{col}} = \max\!\left(0,\ 1 -
  \frac{\Delta E_{ab}(\hat{\mathbf{c}}, \mathbf{c}^*)}{50}\right)
  \label{eq:col_reward}
\end{equation}
where $\hat{\mathbf{c}}$ and $\mathbf{c}^*$ denote the predicted and ground-truth colors, $\Delta E_{ab}$ denotes the CIE76 color difference, and 50 normalizes the practical perceptual range.

\noindent\textbf{(2) Grounding Reward.}
The grounding task contains dense grounding and captioned grounding. For dense full-screen localization, we adopt Dual-IoU~\cite{wang2025traceableground}, which captures both recall-oriented coverage and precision-oriented fidelity:
\begin{equation}
  R_{\text{rec}} = \frac{1}{N}\sum_{i=1}^{N}
  \max_{j:\hat{c}_j=c_i}\,\text{IoU}(\hat{b}_j, b_i), \quad
  R_{\text{pre}} = \frac{1}{M}\sum_{j=1}^{M}
  \max_{i:c_i=\hat{c}_j}\,\text{IoU}(b_i, \hat{b}_j)
\end{equation}
\begin{equation}
  R_{\text{dense}} = R_{\text{rec}} + R_{\text{pre}}
  + R_{\text{fmt}}
\end{equation}
where $N$ and $M$ are the numbers of ground-truth and predicted boxes, respectively, and $c_i$ and $\hat{c}_j$ are their categories. Empty category-matched sets yield 0; the corresponding average is defined as 0 when its box set is empty.
This dual formulation prevents degenerate solutions: optimizing recall alone encourages over-detection, whereas optimizing precision alone encourages under-detection.

Captioned grounding localizes a specific element from a natural-language instruction:
\begin{equation}
  R_{\text{cap}} = \text{IoU}(\hat{b}, b^*) + R_{\text{fmt}}
\end{equation}
where $\hat{b}$ and $b^*$ denote the predicted and ground-truth bounding boxes, respectively.

Since dense and captioned grounding are triggered by different prompts, we use a task-conditional grounding reward:
\begin{equation}
  R_{\text{ground}} = \mathbf{1}[\text{dense}]\cdot R_{\text{dense}}
  + \mathbf{1}[\text{cap}]\cdot R_{\text{cap}}
\end{equation}

\noindent\textbf{UI Generation Reward.}

We design a three-part generation reward combining \textbf{format validity}, \textbf{renderability}, and \textbf{perceptual quality}.

\noindent\textbf{(1) Format Reward.}
$R_{\text{fmt}}\in\{0,1\}$ checks JSON parseability and conformance to the generation schema. For referring, format checks require category, text, and text-color fields; grounding uses the task's coordinate schema, with category labels required for dense predictions.

\noindent\textbf{(2) Rendering Reward.}
$R_{\text{render}}$ checks whether the DSL can be rendered without execution or layout errors. UI-UG 2 renders each sampled output online with a server-side Playwright pipeline before computing visual rewards, aligning optimization with inference-time artifacts.

\noindent\textbf{(3) Perceptual Quality Reward.}
We model perceptual quality with \textbf{LLM-as-a-Judge} and \textbf{Contrastive Preference Ranking (CPR)}, which provide complementary absolute and relative signals. Following Web2Code~\cite{yun2024web2code}, rendered UIs are scored by Qwen3-VL-235B-A22B along four dimensions: (1)~Color \& Aesthetic Design, (2)~Visual Structure \& Alignment, (3)~Textual \& Content Consistency, and (4)~Interface Usability \& Interactivity:

\begin{equation}
R_{\text{judge}} = \frac{1}{4}\sum_{k=1}^{4}
\frac{s_k}{10}
\end{equation}
where $s_k \in [0,10]$ is the judge score on dimension $k$.
This online render-and-judge loop reduces the mismatch between offline preference data and inference-time UI artifacts.

\textit{CPR} mitigates calibration drift in absolute scoring by adding round-robin pairwise comparisons within each GRPO rollout group. For $G$ sampled responses and $K{=}4$ quality dimensions:
\begin{equation}
  R_{\text{cpr}}^{(i)} = \frac{1}{(G-1) \cdot K}
  \sum_{j \neq i}\sum_{k=1}^{K}
  w_k(y^{(i)}, y^{(j)})
\end{equation}
where $y^{(i)}$ and $y^{(j)}$ are the $i$-th and $j$-th sampled responses, and $w_k \in \{0, 0.5, 1\}$ indicates lose/tie/win on dimension $k$.

The perceptual quality reward combines both signals:
\begin{equation}
  R_{\text{perc}} = \alpha \cdot R_{\text{judge}} +
  (1-\alpha) \cdot R_{\text{cpr}}, \quad \alpha = 0.5
\end{equation}

The final generation reward integrates all three components:
\begin{equation}
  R_{\text{gen}} = R_{\text{fmt}} + R_{\text{render}}
  + R_{\text{perc}}
  \label{eq:gen_reward}
\end{equation}
where $R_{\text{perc}}$ is computed only when both $R_{\text{fmt}} = 1$ and $R_{\text{render}} = 1$.

\noindent\textbf{Sequential GRPO Training.}
We use sequential task optimization, as supported by the schedule ablation in Section~\ref{sec:seqab}. The selected curriculum optimizes generation, then referring, then grounding, with one GRPO epoch per stage. Training settings are reported below.

\subsection{Training Configuration}\label{sec:training}
\label{app:hyperparams}

We train the policy on 4 A100 80GB GPUs. SFT uses 120K instruction samples (approximately 0.6B tokens) with a referring:dense/GUI-grounding:generation ratio of 3:3:4. GRPO uses 6,000 referring samples, 6,000 grounding samples, and 2,000 generation samples, with eight rollouts per prompt. The policy-training time is approximately two days for SFT and two days for GRPO; it does not include a separate accounting of judge-service compute.

Table~\ref{tab:hyperparams} summarizes the reported SFT and GRPO settings.

\begin{table}[htbp]
\centering
\footnotesize
\caption{Training hyperparameters for SFT and GRPO stages.}
\label{tab:hyperparams}
\begin{tabular}{lcc}
\toprule
\textbf{Hyperparameter} & \textbf{SFT} & \textbf{GRPO} \\
\midrule
Base Model & \multicolumn{2}{c}{Qwen3-VL-4B-Instruct} \\
Frozen Modules & \multicolumn{2}{c}{ViT Encoder} \\
Trainable Modules & \multicolumn{2}{c}{LLM + VL Adapter} \\
\midrule
Max Learning Rate & 1e-5 & 1e-6 \\
LR Schedule & Warmup-Stable-Decay & Warmup-Stable-Decay \\
Warmup Ratio & 0.05 & 0.03 \\
Optimizer & AdamW & AdamW \\
Weight Decay & 0.01 & 0.01 \\
Max Sequence Length & 8192 & 8192 \\
Global Batch Size & 16 & 64 \\
Epochs & 2 & 1 per stage \\
\midrule
\# SFT Samples & 120K ($\sim$0.6B tokens) & --- \\
Task Ratio (Ref:Grd:Gen) & 3:3:4 & --- \\
\midrule
\# Rollouts per Prompt & --- & 8 \\
KL Penalty Coefficient & --- & 0.01 \\
GRPO Samples (Referring) & --- & 6,000 \\
GRPO Samples (Grounding) & --- & 6,000 \\
GRPO Samples (Generation) & --- & 2,000 \\
\midrule
Hardware & \multicolumn{2}{c}{4$\times$ A100 80GB} \\
Training Time & $\sim$2 days & $\sim$2 days \\
\bottomrule
\end{tabular}
\end{table}

The hardware above refers to policy-model training. The Qwen3-VL-235B-A22B judge and the Playwright rendering service are deployed separately and queried asynchronously during generation GRPO.

\FloatBarrier

%% file: sections/05_experiments.tex
\section{Experiments}\label{sec:experiments}

We compare UI-UG 2-4B against both general-purpose MLLMs
(GPT-5.2, Claude~Sonnet~4.5, Gemini~3.1~Pro, Kimi~K2.5~\cite{kimi},
GLM-4.5V~\cite{glm45}, Mistral Large~3~\cite{Mistral}, and the Qwen3-VL series)
and UI-specific models (Ferret-UI2, OmniParser~V2,
ShowUI, UI-TARS, GUI-Owl, and UI-Venus) on UI understanding and UI generation.
Compared with UI-UG~\cite{uiugv1}, the evaluation here is broader in both task coverage and evidence: we add GUI grounding on ScreenSpot mobile benchmarks, use confidence-free dense-grounding metrics tailored to MLLM outputs, and include broader ablations on online RL, reward transfer, and understanding-generation interaction.

\subsection{UI Understanding}

UI understanding consists of Referring, Dense Grounding, and GUI Grounding. Referring and Dense Grounding are evaluated on our in-house dataset, while GUI Grounding is evaluated on public mobile benchmarks. The former assesses the proposed annotation space within the curated corpus; ScreenSpot provides complementary external single-target localization evidence.

\noindent\textbf{Evaluation.}
We evaluate \textbf{Referring} by predicting category, OCR text, and color attribute from a ground-truth box, using Category Accuracy (Cat), OCR-Acc (Eq.~\ref{eq:ocr_reward}), and Color-Acc (Eq.~\ref{eq:col_reward}). \textbf{Dense Grounding} is evaluated by F1, Precision, Recall, and mIoU; because MLLMs do not provide confidence scores, we use set-matching F1 rather than AP-style metrics. \textbf{GUI Grounding} is evaluated by click accuracy on the mobile splits of ScreenSpot V1~\cite{cheng2024seeclick} and ScreenSpot-V2~\cite{wu2024atlas}, yielding 502 and 449 samples after mobile-only filtering. All MLLMs use standardized prompts with temperature~0 and model-specific coordinate adaptations when needed. For models whose interfaces do not support certain tasks, we mark the corresponding entries as unsupported in Table~\ref{tab:understanding_gui}; UI-UG-7B~\cite{uiugv1} is mapped to our taxonomy for comparison.

\noindent\textbf{Dense set matching.}
After parsing and discarding malformed or out-of-taxonomy predictions, we match predicted and reference boxes with a Hungarian assignment whose cost is $1-\mathrm{IoU}$ for equal categories and $1$ otherwise. A true positive requires equal categories and IoU $\geq 0.5$. Precision, recall, and F1 are computed per image and then averaged; mIoU is computed over category-correct true-positive matches. Consequently, the reported mean F1 need not equal the harmonic mean of mean precision and mean recall. Appendix~\ref{app:dense_eval} provides the step-by-step protocol.

\noindent\textbf{Results.}
Table~\ref{tab:understanding_gui} shows that UI-UG 2-4B achieves the best overall performance on Referring and Dense Grounding.

On \textbf{Referring}, general-purpose MLLMs remain inconsistent: the strongest closed-source baseline, Gemini~3.1~Pro, reaches 0.535 category accuracy, and the best open-source general model, Kimi~K2.5, reaches 0.514. Relative to UI-UG-7B, UI-UG 2 pushes the unified setting to a substantially stronger operating point: UI-UG 2-4B-SFT already reaches 0.904/0.930/0.727 on Cat/OCR/Color, and GRPO further improves this to \textbf{0.906/0.938/0.739}. The same-backbone SFT-to-GRPO comparison isolates the additional post-training gain. The comparison to UI-UG-7B also changes the backbone, label space, and data, so it does not isolate any single design choice.

On \textbf{Dense Grounding}, the margin is larger: the best general-purpose model, Gemini~3.1~Pro, achieves 0.471 F1, while UI-UG 2-4B-SFT reaches 0.742 F1 and 0.865 mIoU, and GRPO improves this further to \textbf{0.777 F1} and \textbf{0.894 mIoU} ($+$0.035 F1 and $+$0.029 mIoU over SFT). Relative to UI-UG-7B, the new model again advances the unified setting by a wide margin, with the difference most visible in recall because the earlier 20-class label space cannot fully cover our 80-category protocol. Among UI-specific models, ShowUI and UI-TARS expose point/action-based single-target interfaces and do not provide the exhaustive typed boxes or referring attributes required by this protocol, whereas GUI-Owl and UI-Venus-Ground retain grounding ability.

On \textbf{GUI Grounding}, UI-UG 2-4B attains 0.902/0.929 on ScreenSpot-V1/V2 mobile, outperforming most open-source general MLLMs and competitive 7B specialists such as UI-TARS-7B (0.886/0.911). Several larger general-purpose models and task-specific baselines remain higher, including UI-Venus-Ground-7B (0.910/0.951) and UI-Venus-Ground-72B (0.922/0.971), likely because they are optimized only for single-target localization. Still, UI-UG 2-4B clearly surpasses its base model Qwen3-VL-4B, validating the effectiveness of the unified training recipe.

\begin{table}[htbp]
\centering
\scriptsize
\renewcommand{\arraystretch}{0.95}
\caption{UI understanding results. Best results are in \textbf{bold}; second-best are \underline{underlined}. $-$: unknown; $\dagger$: unsupported by the native output interface. Unsupported entries are excluded from task rankings.}
\setlength{\tabcolsep}{1.2pt}
\begin{tabular*}{\linewidth}{@{\extracolsep{\fill}}
  l c
  ccc
  cccc
  cc}
\toprule
& & \multicolumn{3}{c}{\textbf{Referring}}
& \multicolumn{4}{c}{\textbf{Dense Grounding}}
& \multicolumn{2}{c}{\textbf{GUI Grounding}} \\
\cmidrule(lr){3-5}\cmidrule(lr){6-9}\cmidrule(lr){10-11}
\textbf{Model} & \textbf{Params}
  & Category & OCR & Color
  & F1 & Precision & Recall & mIoU
  & SS-V1-Mobile & SS-V2-Mobile \\
\midrule
\multicolumn{11}{l}{\textit{Closed-source}} \\
GPT-5.2~\cite{gpt5}
  & $-$
  & 0.415 & 0.551 & 0.611 & 0.048 & 0.045 & 0.052 & 0.656 & 0.492 & 0.352 \\
Claude~Sonnet~4.5~\cite{Claude}
  & $-$
  & 0.484 & 0.830 & 0.628 & 0.368 & 0.415 & 0.330 & 0.757 & 0.187 & 0.232 \\
Gemini~3.1~Pro~\cite{gemini}
  & $-$
  & 0.535 & 0.690 & 0.501 & 0.471 & 0.461 & 0.482 & 0.759 & 0.900 & 0.951 \\
\midrule
\multicolumn{11}{l}{\textit{Open-source}} \\
Qwen3-VL-4B~\cite{bai2025qwen3vltechnicalreport}
  & 4B
  & 0.310 & 0.775 & 0.417 & 0.213 & 0.354 & 0.152 & 0.783 & 0.861 & 0.886 \\
Qwen3-VL-8B~\cite{bai2025qwen3vltechnicalreport}
  & 8B
  & 0.463 & 0.830 & 0.614 & 0.289 & 0.378 & 0.234 & 0.729 & 0.466 & 0.918 \\
Qwen3-VL-30B~\cite{bai2025qwen3vltechnicalreport}
  & 30B (A3B)
  & 0.407 & 0.812 & 0.571 & 0.221 & 0.315 & 0.170 & 0.775 & 0.912 & 0.940 \\
Qwen3-VL-235B~\cite{bai2025qwen3vltechnicalreport}
  & 235B (A22B)
  & 0.449 & 0.855 & 0.484 & 0.295 & 0.385 & 0.239 & 0.766 & \textbf{0.926} & \underline{0.958} \\
Kimi~K2.5~\cite{kimi}
  & 1T (A32B)
  & 0.514 & 0.845 & 0.689 & 0.292 & 0.312 & 0.274 & 0.727 & 0.910 & 0.953 \\
GLM-4.5V~\cite{glm45}
  & 106B (A12B)
  & 0.352 & 0.799 & 0.552 & 0.264 & 0.315 & 0.227 & 0.742 & 0.616 & 0.933 \\

Mistral Large~3~\cite{Mistral}
  & 675B (A41B)
  & 0.321 & 0.628 & 0.552 & 0.179 & 0.192 & 0.180 & 0.723 & 0.161 & 0.261 \\
\midrule
\multicolumn{11}{l}{\textit{UI-specific}} \\
Ferret-UI2-2B~\cite{ferret-ui2}
  & 2B
  & 0.470 & 0.220 & 0.312 & 0.031 & 0.036 & 0.027 & 0.667 & 0.380 & 0.479 \\
Ferret-UI2-8B~\cite{ferret-ui2}
  & 8B
  & 0.493 & 0.284 & 0.312 & 0.031 & 0.036 & 0.027 & 0.667 & 0.538 & 0.521 \\
OmniParser~V2~\cite{omniparserv2}
  & 0.02B
  & $\dagger$ & $\dagger$ & $\dagger$
  & 0.342 & 0.299 & 0.401 & 0.715 & 0.612 & 0.635 \\
ShowUI-2B~\cite{lin2025showui}
  & 2B
  & $\dagger$ & $\dagger$ & $\dagger$
  & $\dagger$ & $\dagger$ & $\dagger$ & $\dagger$
  & 0.823 & 0.853 \\
UI-TARS-7B~\cite{qin2025ui-tars}
  & 7B
  & $\dagger$ & $\dagger$ & $\dagger$
  & $\dagger$ & $\dagger$ & $\dagger$ & $\dagger$
  & 0.886 & 0.911 \\
UI-TARS-72B~\cite{qin2025ui-tars}
  & 72B
  & $\dagger$ & $\dagger$ & $\dagger$
  & $\dagger$ & $\dagger$ & $\dagger$ & $\dagger$
  & 0.890 & 0.902 \\
GUI-Owl-7B~\cite{guiowl}
  & 7B
  & 0.174 & 0.561 & 0.390 & 0.013 & 0.007 & 0.044 & 0.855 & 0.898 & 0.931 \\
GUI-Owl-32B~\cite{guiowl}
  & 32B
  & 0.176 & 0.561 & 0.392 & 0.014 & 0.008 & 0.045 & 0.855 & 0.898 & 0.940 \\
UI-Venus-Ground-7B~\cite{gu2025ui}
  & 7B
  & 0.268 & 0.821 & 0.480 & 0.081 & 0.128 & 0.059 & 0.811 & 0.910 & 0.951 \\
UI-Venus-Ground-72B~\cite{gu2025ui}
  & 72B
  & 0.451 & 0.867 & 0.627 & 0.342 & 0.511 & 0.257 & 0.778 & \underline{0.922} & \textbf{0.971} \\
UI-UG-7B~\cite{uiugv1}
  & 7B
  & 0.738 & 0.759 & 0.583 & 0.447 & 0.653 & 0.340 & 0.840 & 0.637 & 0.656 \\
\midrule
\multicolumn{11}{l}{\textit{Ours}} \\
UI-UG 2-4B-SFT
  & 4B
  & \underline{0.904} & \underline{0.930} & \underline{0.727} & \underline{0.742} & \underline{0.798} & \underline{0.693} & \underline{0.865} & 0.869 & 0.889 \\
\textbf{UI-UG 2-4B}
  & 4B
  & \textbf{0.906} & \textbf{0.938} & \textbf{0.739} & \textbf{0.777} & \textbf{0.832} & \textbf{0.729} & \textbf{0.894} & 0.902 & 0.929 \\
\bottomrule
\end{tabular*}
\label{tab:understanding_gui}
\end{table}

\FloatBarrier
\subsection{UI Generation}

\noindent\textbf{Evaluation.}
We evaluate generation by JSON format compliance, render success rate, and four MLLM-scored dimensions: Color \& Aesthetic Design, Visual Structure \& Alignment, Textual \& Content Consistency, and Interface usability \& Interactivity. The four reported judge dimensions are normalized to $[0,1]$. The reported composite adds the two validity rates and the four aggregate judge dimensions:
\begin{equation}
S_{\mathrm{gen}} = p_{\mathrm{JSON}} + p_{\mathrm{render}} + \sum_{k=1}^{4}\bar{q}_k,\qquad S_{\mathrm{gen}}\in[0,6],
\end{equation}
where $p_{\mathrm{JSON}}$ and $p_{\mathrm{render}}$ are the reported JSON-validity and rendering-success rates, and $\bar{q}_k$ is the reported mean normalized score on judge dimension $k$. This composite sums the four visual dimensions, whereas the training reward in Eq.~\ref{eq:gen_reward} uses their average together with CPR. Training and evaluation share the quality rubric; the generation score should therefore be interpreted together with format/rendering checks and the human-reference study rather than as an independent measure of all aspects of UI quality.

\noindent\textbf{Human-reference judge validation.}
On 100 generated UI cases scored independently by three professional UI designers, the Qwen3-VL-235B-A22B judge has a mean absolute deviation of 0.17 and residual variance of 0.046 against the averaged human reference. Gemini~3.1~Pro has the lowest deviation (0.08) in this comparison; Qwen is used for the training reward for deployment and repeated-evaluation practicality. The complete comparison and scoring prompts are in Appendices~\ref{app:judge_validity} and~\ref{app:judge_prompt}. This post-hoc study does not establish human alignment at every RL update.

\noindent\textbf{Results.}
Table~\ref{tab:generation} shows that UI-UG 2-4B achieves the best overall generation quality and improves the average total score by \textbf{5.8\%} over the SFT model. The comparison demonstrates a stronger compact operating point than UI-UG-7B, while changes in backbone, data, and optimization prevent attributing the full difference to GRPO alone. In particular, UI-UG used offline DPO over a fixed preference dataset, whereas UI-UG 2 uses online render-aware GRPO, aligning optimization with the final executable DSL after actual rendering. This advantage is also reflected in comparisons to strong baselines: UI-UG 2-4B surpasses Claude~Sonnet~4.5 (4.70), GPT-5.2 (4.33), and Gemini~3.1~Pro (3.89), and also outperforms Qwen3-VL-235B, which uses roughly \textbf{59$\times$} as many total parameters and \textbf{5.5$\times$} as many active parameters. Larger general models also do not reliably produce valid DSL outputs: GLM-4.5V and Qwen3-VL-30B show notably reduced format compliance (0.60 and 0.54), indicating that general instruction tuning alone is insufficient for structured UI generation.

\noindent\textbf{Representation efficiency.}
The JSON DSL reduces token count by about \textbf{16.7\%} relative to equivalent HTML in the reported corpus comparison. Appendix~\ref{app:inference} separates the measured UI-UG 2 latency from active-parameter-based estimates for larger models; these estimates are not same-hardware measurements of those models.

\begin{table}[htbp]
\scriptsize
\centering
\caption{UI generation results. JSON: format compliance; Rend: render success; Col/Str/Txt/Int: four judge dimensions; Sum: composite score (maximum 6). Best results are in \textbf{bold}; second-best are \underline{underlined}.}
\setlength{\tabcolsep}{6pt}
\renewcommand{\arraystretch}{0.95}
\begin{tabular}{l cc cccc c}
\toprule
\textbf{Model}
  & \textbf{JSON} & \textbf{Rend}
  & \textbf{Col} & \textbf{Str}
  & \textbf{Txt} & \textbf{Int}
  & \textbf{Sum} \\
\midrule
\multicolumn{8}{l}{\textit{Closed-source}} \\
GPT-5.2~\cite{gpt5}
  & \underline{0.99} & 0.96 & 0.58 & 0.62 & 0.58 & 0.60 & 4.33 \\
Claude~Sonnet~4.5~\cite{Claude}
  & 0.96 & 0.94 & 0.69 & \underline{0.70} & 0.71 & \underline{0.70} & \underline{4.70} \\
Gemini~3.1~Pro~\cite{gemini}
  & \textbf{1.00} & \underline{0.97} & 0.42 & 0.53 & 0.48 & 0.49 & 3.89 \\
\midrule
\multicolumn{8}{l}{\textit{Open-source}} \\
Qwen3-VL-4B~\cite{bai2025qwen3vltechnicalreport}
  & 0.98 & 0.96 & 0.59 & 0.59 & 0.60 & 0.58 & 4.30 \\
Qwen3-VL-8B~\cite{bai2025qwen3vltechnicalreport}
  & 0.67 & 0.66 & 0.61 & 0.62 & 0.59 & 0.61 & 3.76 \\
Qwen3-VL-30B~\cite{bai2025qwen3vltechnicalreport}
  & 0.54 & 0.48 & 0.64 & 0.63 & 0.55 & 0.62 & 3.46 \\
Qwen3-VL-235B~\cite{bai2025qwen3vltechnicalreport}
  & \textbf{1.00} & \textbf{0.98} & 0.66 & 0.68 & 0.61 & 0.66 & 4.59 \\
Kimi~K2.5~\cite{kimi}
  & 0.86 & 0.86 & 0.66 & 0.68 & 0.61 & 0.65 & 4.32 \\
GLM-4.5V~\cite{glm45}
  & 0.60 & 0.59 & 0.64 & 0.66 & 0.62 & 0.64 & 3.75 \\

Mistral Large~3~\cite{Mistral}
  & \underline{0.99} & 0.95 & 0.56 & 0.54 & 0.55 & 0.54 & 4.13 \\
\midrule
\multicolumn{8}{l}{\textit{UI-specific}} \\
UI-UG-7B~\cite{uiugv1}
  & 0.93 & 0.89 & 0.62 & 0.57 & 0.63 & 0.56 & 4.20 \\
\midrule
\multicolumn{8}{l}{\textit{Ours}} \\
UI-UG 2-4B-SFT
  & 0.93 & 0.93 & \underline{0.71} & 0.67 & \underline{0.74} & 0.67 & 4.65 \\
\textbf{UI-UG 2-4B}
  & \underline{0.99} & \textbf{0.98} & \textbf{0.73} & \textbf{0.72} & \textbf{0.79} & \textbf{0.71} & \textbf{4.92} \\
\bottomrule
\end{tabular}
\label{tab:generation}
\end{table}

\FloatBarrier
\subsection{Ablation Study}
\label{sec:ablation}

\begin{figure}[htbp]
\centering
\includegraphics[width=0.98\linewidth]{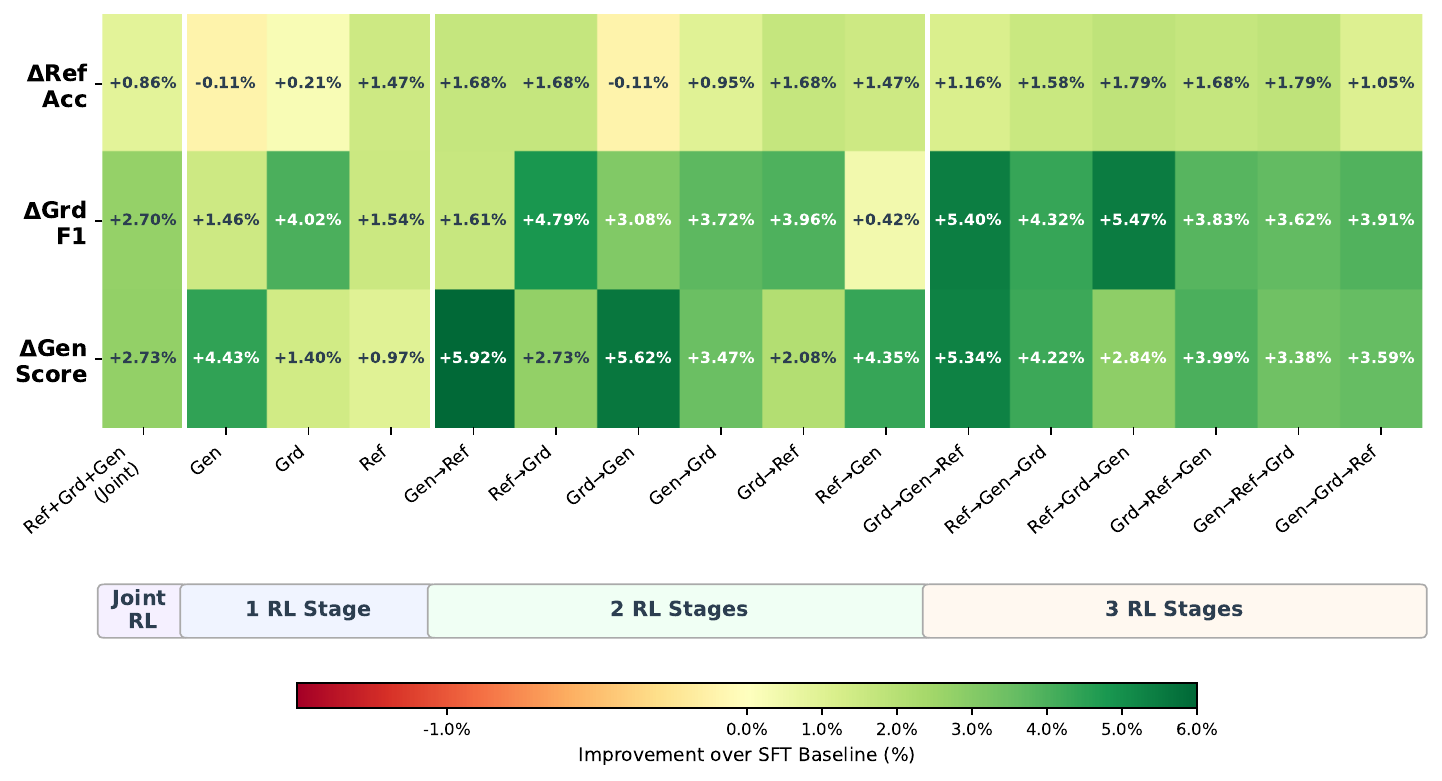}

\caption{RL strategy ablation. Each cell shows improvement over SFT; sequential multi-stage training yields larger gains.}

\label{fig:heat}
\end{figure}

\noindent\textbf{RL Training Strategy.}
We ablate RL objectives and training schedules. Figure~\ref{fig:heat} reports improvement over the SFT baseline in referring accuracy ($\Delta$Ref Acc), grounding F1 ($\Delta$Grd F1), and generation score ($\Delta$Gen) across joint training and 1/2/3-stage sequential RL configurations.

\textbf{Finding 1: RL improves over SFT.}
Task-aligned RL consistently improves each target capability over the SFT baseline: single-stage GenerationRL raises generation score by +4.43\%, GroundingRL improves grounding F1 by +4.02\%, and ReferringRL increases referring accuracy by +1.47\%. Notably, understanding-oriented RL also benefits generation, suggesting that stronger element-level understanding transfers to UI generation. The gains further scale with the number of training stages, as 2-stage and 3-stage sequential configurations progressively improve all three metrics.

\label{sec:seqab}
\textbf{Finding 2: Sequential outperforms joint optimization.}
Joint RL yields Gen +2.73\%, Grd F1 +2.70\%, and Ref Acc +0.86\%, below the selected sequential configurations at the same reported training budget. The tasks differ in response length and reward distribution; this pattern is consistent with greater interference during simultaneous optimization. The pattern is consistent with prior work showing that heterogeneous rewards are difficult to combine into a single stable objective~\cite{wu2026imbalanced}.

\subsubsection{Dense Grounding Reward Components}
\label{app:dual_iou}

The dense grounding reward $R_{\text{dense}} = R_{\text{fmt}} + R_{\text{pre}} + R_{\text{rec}}$ comprises three components. We compare combinations of these terms: (1) a format reward $R_{\text{fmt}}$ enforcing valid JSON with parseable bounding boxes; (2) a precision-oriented term $R_{\text{pre}} = \frac{1}{M}\sum_{j=1}^{M}\max_{i}\,\text{IoU}(b_i, \hat{b}_j)$ to suppress spurious predictions; and (3) a recall-oriented term $R_{\text{rec}} = \frac{1}{N}\sum_{i=1}^{N}\max_{j}\,\text{IoU}(\hat{b}_j, b_i)$ to improve coverage. Table~\ref{tab:dual_iou} reports an additive ablation from the SFT baseline.

\begin{table}[htbp]
\centering
\small
\caption{Ablation of dense grounding reward components. Each row specifies the reward combination applied to the SFT baseline. Precision, Recall, and F1 are averaged over samples; therefore, the reported F1 is not necessarily the harmonic mean of the averaged Precision and Recall.}
\label{tab:dual_iou}
\setlength{\tabcolsep}{4pt}
\begin{tabular}{lcccc}
\toprule
\textbf{Configuration} & \textbf{Precision} & \textbf{Recall} & \textbf{F1} & \textbf{mIoU} \\
\midrule
SFT (no RL) & 0.798 & 0.693 & 0.742 & 0.865 \\
\midrule
$+R_{\text{fmt}}$ & 0.803 & 0.697 & 0.745 & 0.868 \\
$+R_{\text{fmt}}+R_{\text{pre}}$ & \textbf{0.836} & 0.712 & 0.763 & 0.877 \\
$+R_{\text{fmt}}+R_{\text{rec}}$ & 0.754 & \textbf{0.742} & 0.727 & 0.836 \\
$+R_{\text{fmt}}+R_{\text{pre}}+R_{\text{rec}}$ & 0.832 & 0.729 & \textbf{0.777} & \textbf{0.894} \\
\bottomrule
\end{tabular}
\end{table}

From Table~\ref{tab:dual_iou}, adding $R_{\text{fmt}}$ first gives a small but consistent improvement over SFT (F1: 0.742 $\rightarrow$ 0.745; mIoU: 0.865 $\rightarrow$ 0.868), mainly by reducing invalid outputs. Adding $R_{\text{pre}}$ on top of $R_{\text{fmt}}$ further improves overall quality (Precision: 0.836; F1: 0.763; mIoU: 0.877), showing that explicit precision control effectively suppresses noisy boxes.

An important phenomenon appears when training with recall emphasis alone: the $R_{\text{fmt}}+R_{\text{rec}}$ setting increases Recall to 0.742, but Precision drops noticeably to 0.754, and both F1 (0.727) and mIoU (0.836) become worse than the SFT baseline. This indicates a reward-hacking pattern, where the model tends to output more boxes to gain recall reward. Therefore, bidirectional IoU supervision is necessary. The full objective $R_{\text{fmt}}+R_{\text{pre}}+R_{\text{rec}}$ (UI-UG 2-4B) restores the balance and achieves the best final performance (F1: 0.777, mIoU: 0.894), with gains over SFT of +0.035 F1 and +0.029 mIoU.

\subsection{Understanding--Generation Transfer Analysis}
\label{sec:transfer_analysis}

We analyze the interaction between UI understanding and generation from three perspectives.

\textbf{Experiment 1: Effect of Understanding-Generation Data Mixing Ratio during SFT.}
We vary the proportion of understanding data (referring + grounding) and generation data during SFT under a fixed SFT budget of 120K samples (approximately 0.6B tokens).

\begin{figure}[htbp]
\centering
\includegraphics[width=0.95\linewidth]{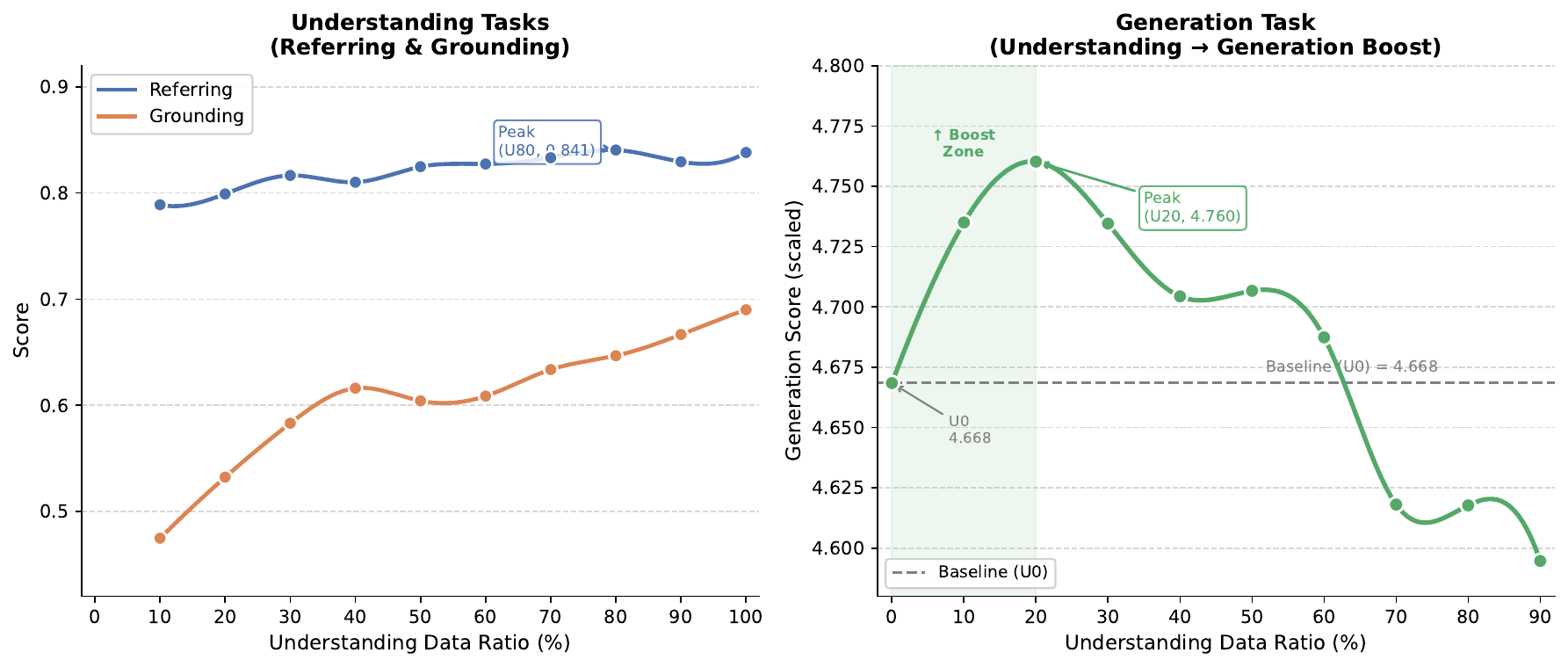}

\caption{Effect of understanding-data ratio during SFT. Generation peaks at U20, while understanding metrics increase monotonically.}

\label{fig:mix_ratio}
\end{figure}

Figure~\ref{fig:mix_ratio} shows a non-monotonic pattern. U$xx$ denotes that $xx\%$ of training data comes from understanding tasks. Generation peaks at \textbf{U20 (4.760)}, improving over the generation-only setting U0 (4.668). Because the total training budget is fixed, this gain comes from complementary supervision rather than larger scale. Beyond U20, generation quality gradually declines as the model receives less generation-specific supervision. In contrast, understanding improves monotonically with understanding ratio: referring accuracy rises from 0.789 (U10) to 0.838 (U100), and grounding F1 from 0.475 to 0.690. This supports an asymmetric transfer pattern in this fixed-budget mixture: moderate understanding supervision benefits generation, while replacing understanding supervision with generation does not improve the measured understanding metrics.

\textbf{Experiment 2: Dimension-Specific Understanding Rewards
for Generation.}
We test whether different understanding rewards improve their semantically aligned generation dimensions. Each reward is applied independently during RL, and the resulting change in generation score is measured relative to the SFT baseline.

Table~\ref{tab:reward_ablation} shows a clear alignment between understanding skills and generation dimensions. $+R_{\text{col}}$ gives the notable gain on Color \& Aesthetic Design ($+$1.60\%), $+R_{\text{ocr}}$ gives the largest gain on Textual \& Content Consistency ($+$2.48\%), and $+R_{\text{dense}}$ produces the largest overall benefit, especially on Visual Structure \& Alignment ($+$4.72\%). In contrast, $+R_{\text{cap}}$ yields only marginal changes, suggesting that single-element GUI grounding transfers less effectively to holistic generation. The transfer is selective rather than uniform: several off-dimension changes are negative, and GUI-grounding reward alone gives little benefit. The combined model's improvement does not imply that each understanding objective improves every generation dimension.

\begin{table}[htbp]
\scriptsize
\centering
\caption{Relative change in generation scores after applying each understanding reward independently in RL training. Notable Gain is shown
in \textbf{bold}(>1\%).}
\setlength{\tabcolsep}{7pt}
\renewcommand{\arraystretch}{0.95}
\begin{tabular}{l rrrr}
\toprule
\textbf{Reward}
  & \textbf{$\Delta$Col}
  & \textbf{$\Delta$Str}
  & \textbf{$\Delta$Txt}
  & \textbf{$\Delta$Int} \\
\midrule
$+R_{\text{cls}}$ (Category)
  & \textbf{$+$1.31\%} & $+$0.86\% & \textbf{$+$1.31\%} & $-$1.49\% \\
$+R_{\text{col}}$ (Color)
  & \textbf{$+$1.60\%} & $-$0.39\% & $-$0.18\% & $+$0.41\% \\
$+R_{\text{ocr}}$ (Text)
  & $-$0.11\% & $-$1.45\% & \textbf{$+$2.48\%} & $-$0.13\% \\
$+R_{\text{dense}}$ (Dense Gnd.)
  & \textbf{$+$1.64\%} & \textbf{$+$4.72\%} & $+$0.25\% & \textbf{$+$1.74\%} \\
$+R_{\text{cap}}$ (GUI Gnd.)
  & $+$0.20\% & $-$0.27\% & $-$0.25\% & $-$0.17\% \\
\bottomrule
\end{tabular}
\label{tab:reward_ablation}
\end{table}

\textbf{Experiment 3: Cross-Task Attention Similarity Analysis.}
To probe the mechanism behind the understanding--generation synergy, we compare text-to-vision attention patterns across tasks. For sample \(b\) at layer \(l\), we aggregate text-to-vision attention over heads, extract the visual-token slice, and apply min--max normalization to obtain a heatmap \(\hat{\mathbf{V}}_{b,l} \in \mathbb{R}^{H \times W}\), whose flattened form is \(\hat{\mathbf{v}}_{b,l}\). Averaging \(\hat{\mathbf{v}}_{b,l}\) over samples within each task gives the layer-wise task representation \(\hat{\mathbf{v}}_l^{\,p}\).
We then define task similarity as the mean cosine similarity over the \(L\) layers:
\begin{equation}
\mathrm{Sim}_{\mathrm{cos}}(p,q)
=
\frac{1}{L}
\sum_{l=1}^{L}
\frac{\hat{\mathbf{v}}_{l}^{\,p} \cdot \hat{\mathbf{v}}_{l}^{\,q}}
{\|\hat{\mathbf{v}}_{l}^{\,p}\|_2\;\|\hat{\mathbf{v}}_{l}^{\,q}\|_2}.
\end{equation}

\begin{figure}[htbp]
\centering
\includegraphics[width=0.65\linewidth]{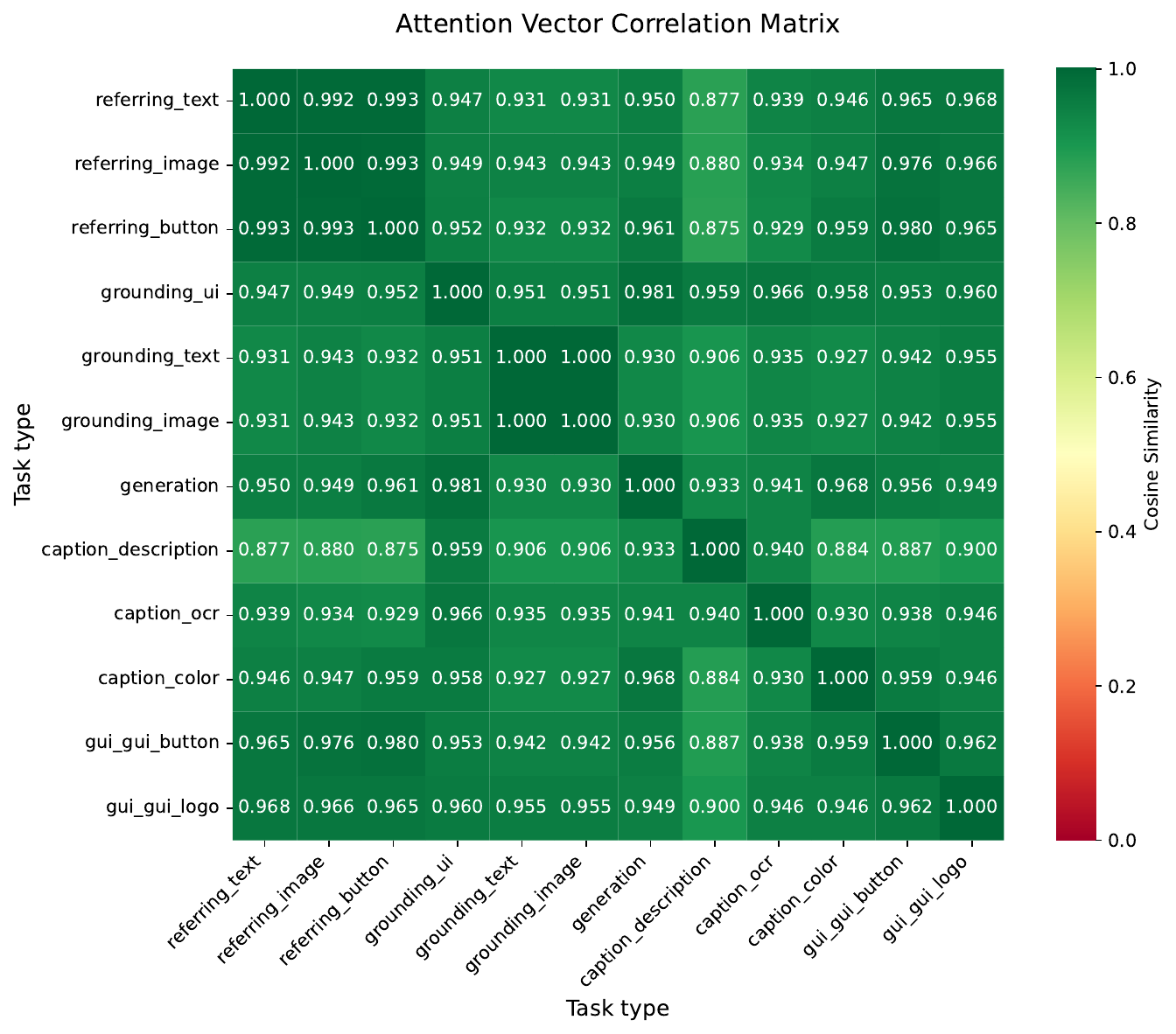}
\caption{Cosine similarity of text-to-vision attention patterns across understanding and generation tasks. Generation and dense grounding show the strongest alignment.}
\label{app:attn_similarity}
\end{figure}

Figure~\ref{app:attn_similarity} shows high similarity across understanding--generation pairs, with generation and dense grounding exhibiting the strongest alignment. This is consistent with the reward-transfer results, but attention similarity is correlational evidence: it does not by itself establish the causal mechanism of transfer.

\FloatBarrier
\subsection{Qualitative Generation Results}

\begin{figure}[htbp]
\centering
\includegraphics[width=0.95\linewidth]{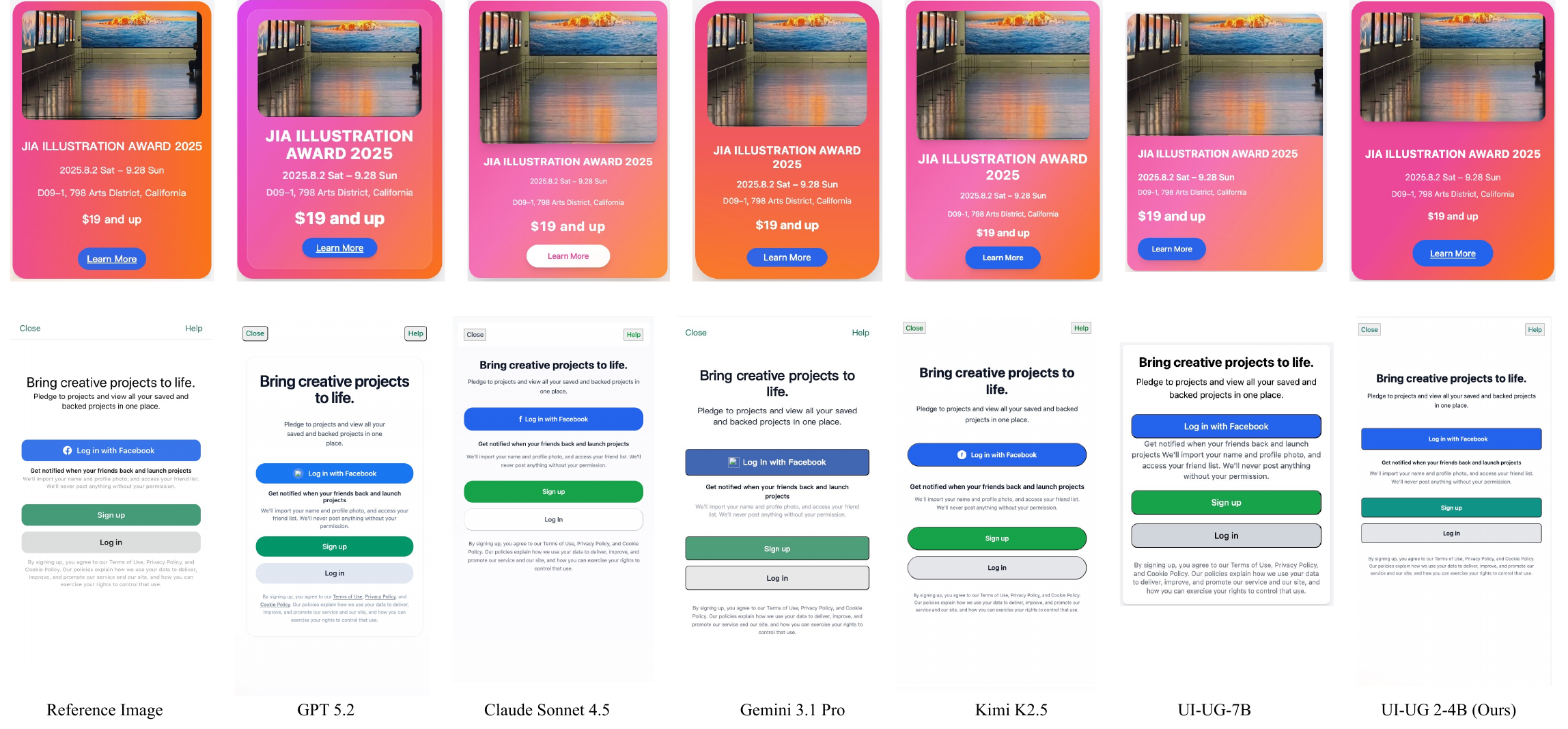}
\caption{Visual comparison of UI generation quality on two representative cases (poster card and login interface). Each column corresponds to one model under the same prompt condition. UI-UG 2-4B (rightmost) shows better layout stability, hierarchy consistency, and style-control fidelity than representative MLLMs.}
\label{fig:gen_compare}
\end{figure}

Figure~\ref{fig:gen_compare} compares generated interfaces under the same requirements: an event-poster card and a mobile login page. UI-UG 2-4B produces more consistent typography, spacing, component geometry, and global style across these two cases. The examples illustrate the layout and content differences reflected in the aggregate scores; they are qualitative cases rather than a substitute for broader functional testing.

\FloatBarrier

%% file: sections/06_discussion.tex
\section{Discussion and Limitations}
\label{sec:limitations}

\noindent\textbf{Benchmark scope and comparisons.}
The largest understanding gains are measured on the proposed 80-category corpus and should be interpreted as in-domain results. ScreenSpot mobile evaluates external single-target localization, where several larger or specialized models remain stronger. Native point/action-based specialists do not expose the typed exhaustive predictions required by our dense/referring protocol; unsupported entries indicate an interface mismatch, not a measured failure on their native task. The UI-UG-7B comparison changes the backbone, data, and taxonomy simultaneously, whereas the Qwen3-VL-4B and UI-UG 2-4B-SFT rows provide same-backbone controls.

\noindent\textbf{Generation evaluation and reward alignment.}
Generation rewards and reported scores share four visual-quality dimensions, so optimization against the rubric may overstate improvements outside that rubric. The human-reference study is limited to 100 cases and is post-hoc rather than a continuous audit during GRPO. Screenshot-based judgments of usability assess visual affordances; they do not replace interactive functional testing, accessibility evaluation, or end-user studies. Online rendering and large-model judging also add service compute beyond the reported policy-training budget.

\noindent\textbf{Transfer and deployment.}
The fixed-budget mixture and reward ablations support a benefit from selected understanding signals, but some isolated rewards reduce other generation dimensions. Attention alignment is descriptive evidence rather than causal identification. Rare UI types outside the taxonomy, distribution shifts, and environments beyond the studied mobile setting remain open challenges. Finally, the appendix's latency extrapolations use active parameter count as a rough scaling assumption; they are not direct deployment benchmarks of every model.

%% file: sections/07_conclusion.tex
\section{Conclusion}

We presented \textbf{UI-UG 2}, a unified 4B MLLM for mobile UI understanding and generation. With a front-end-aligned taxonomy, broader grounding supervision, a compact renderable DSL, and online render-aware GRPO, UI-UG 2 achieves strong results across referring, dense grounding, GUI grounding, and UI generation. Our ablations further show asymmetric transfer from fine-grained understanding to generation, suggesting that component-level UI perception is a useful foundation for structured UI synthesis.

%% file: appendices/01_taxonomy.tex
\section{Complete UI Element Taxonomy}
\label{app:taxonomy}

This section provides a full taxonomy reference used throughout our UI understanding and generation pipeline. We include the complete parent-child structure together with front-end correspondence for reproducibility.

Our taxonomy follows the structure in ``Mobile UI Component Taxonomy: Category Definitions and Library Mappings,'' with \textbf{80 categories} in total, organized as \textbf{9 parent classes + 71 fine-grained subcategories}. The parent classes are annotation super-categories designed for stable dense labeling rather than one-to-one modules in any single front-end library. In this appendix, Ant Design Mobile is used as a practical reference for front-end correspondence. Table~\ref{tab:taxonomy_compare} further situates our taxonomy against representative public UI datasets.

Compared with coarse UI labels, this taxonomy explicitly models three properties that are critical for UI understanding: (1) \textbf{state-level variants} (e.g., checked/unchecked, fold/unfold), (2) \textbf{functional variants} (e.g., different button and card roles), and (3) \textbf{boundary truncation variants} for partially visible elements near screen edges. These design choices improve semantic coverage and annotation usability for dense grounding and referring tasks. Table~\ref{tab:full_taxonomy} provides the complete 9-parent hierarchy and mapping rationale, with subcategory counts summing strictly to 71.

\begin{table}[htbp]
\centering
\scriptsize
\caption{Complete UI taxonomy summary (80 classes = 9 parents + 71 subcategories). AntD Mobile is used here as the reference for front-end component correspondence.}
\label{tab:full_taxonomy}
\renewcommand{\arraystretch}{1.0}
\setlength{\tabcolsep}{3pt}
\begin{tabular}{p{0.16\linewidth} p{0.39\linewidth} p{0.35\linewidth}}
\toprule
\textbf{Parent Class} & \textbf{Subcategories (complete list)} & \textbf{AntD Correspondence} \\
\midrule

\textbf{Card} (9)
  & InfoCard, CommodityCard, MenuCard, InfoCardTruncated, CommodityCardTruncated,
    Banner, BannerTruncated, NoticeBar, Swiper
  & {\raggedright AntD Card (functional/truncation variants);
    AntD NoticeBar; AntD Swiper;
    custom Banner pattern\par} \\
\midrule

\textbf{Collapse} (9)
  & CollapseFold, CollapseUnFold,
    Popup, FloatingPanel, List, ListItem, Table, Form, FormItem
  & {\raggedright AntD Collapse (state-level split);
    AntD Popup/FloatingPanel;
    AntD List/List.Item;
    AntD Form/Form.Item;
    custom Table\par} \\
\midrule

\textbf{Chart} (7)
  & LineChart, HorizontalBarChart, VerticalBarChart, CircleChart,
    RadarChart, DashBoardChart, OtherChart
  & {\raggedright Not in AntD; referenced from AntV/F2, ECharts, and Recharts;
    custom OtherChart for uncommon types\par} \\
\midrule

\textbf{Navigation} (7)
  & Tabs, NavBar, TabBar, SideBar, IndexBar,
    TabsItemSelected, TabsItemUnselected
  & {\raggedright AntD Tabs/NavBar/TabBar/SideBar/IndexBar;
    custom item-state split (Selected/Unselected)\par} \\
\midrule

\textbf{Progress} (6)
  & ProgressBar, ProgressCircle,
    Selector, Steps, StepsItem, Scroll
  & {\raggedright AntD ProgressBar/ProgressCircle;
    AntD Selector; AntD Steps/Steps.Step;
    custom Scroll indicator\par} \\
\midrule

\textbf{Button} (10)
  & CommonButton, SideButton, CustomButton, TextButton,
    CloseButton, MoreButton, HomeButton, ReturnButton,
    Switch, Dropdown
  & {\raggedright AntD Button (functional variants);
    AntD Switch; AntD Dropdown\par} \\
\midrule

\textbf{Text} (16)
  & SingleLineText, MultiLineText, MoneyLineText,
    Popover, FloatingBubble, Icon, Avatar, Tag, Badge, Stepper,
    Input, TextArea, SearchBar, Slider, Calendar, Keyboard
  & {\raggedright AntD Ellipsis (typography);
    domain-specific MoneyLineText;
    AntD Popover, FloatingBubble, Avatar, Tag, Badge, Stepper;
    AntD Input, TextArea, SearchBar, Slider;
    AntD CalendarPicker, NumberKeyboard\par} \\
\midrule

\textbf{Image} (3)
  & Image, BackgroundImage, QRCode
  & {\raggedright AntD Image;
    custom BackgroundImage pattern;
    AntD QRCode (desktop, via qrcode.react)\par} \\
\midrule

\textbf{Checkbox} (4)
  & CheckboxChecked, CheckboxUnchecked, Radio, Rate
  & {\raggedright AntD Checkbox (state-level split);
    AntD Radio; AntD Rate\par} \\

\bottomrule
\end{tabular}
\end{table}

\begin{table}[htbp]
\centering
\footnotesize
\caption{UI element taxonomy coverage comparison. FW-Aligned indicates whether the labels are aligned with front-end component libraries such as Ant Design, Material UI, and Vant.}
\label{tab:taxonomy_compare}
\setlength{\tabcolsep}{2pt}
\begin{tabular}{l c c c c r}
\toprule
\textbf{Dataset} & \textbf{Categories} & \textbf{OCR~/~Color} & \textbf{Level} & \textbf{FW-Aligned} & \textbf{\#~Images} \\
\midrule
RICO~\cite{rico}                        & 24 & \texttimes~/~\texttimes & Element & \texttimes & $\sim$72k \\
VINS~\cite{bunian2021vins}              & 11 & \texttimes~/~\texttimes & Element & \texttimes & $\sim$4k \\
AITW~\cite{rawles2023androidinthewild}  & --- & \texttimes~/~\texttimes & Action & \checkmark & $\sim$715k \\
AC~\cite{li2024androidcontrol}          & --- & \texttimes~/~\texttimes & Action & \texttimes & $\sim$1.5M \\
\midrule
\textbf{Ours} & \textbf{80~(9\,+\,71)} & \checkmark & Element+Action & \checkmark & \textbf{>30k} \\
\bottomrule
\end{tabular}
\end{table}

\FloatBarrier

%% file: appendices/02_dataset.tex
\section{UI Image Dataset Statistics}
\label{app:dataset_stats}

This section summarizes how the UI understanding dataset is composed and why its distribution is suitable for robust training and evaluation.

\subsection{Data Sources and Scale}

Following the data setup in the main text (Data Construction section), our curated \textbf{UI understanding dataset} is built from four sources: RICO, VINS, Android in the Wild, and Android Control. After CLIP-based deduplication, the merged corpus contains \textbf{over 30,000 unique screenshots}, with an almost balanced contribution of roughly ${\sim}$7.5k images per source. All samples are annotated under our \textbf{80-category taxonomy} (9 parent classes + 71 child classes), providing unified supervision for UI understanding across heterogeneous mobile UI domains.

\subsection{Annotation Visualization}

\begin{figure}[htbp]
\centering
\begin{minipage}[t]{0.235\textwidth}
\centering
\includegraphics[height=1.55in]{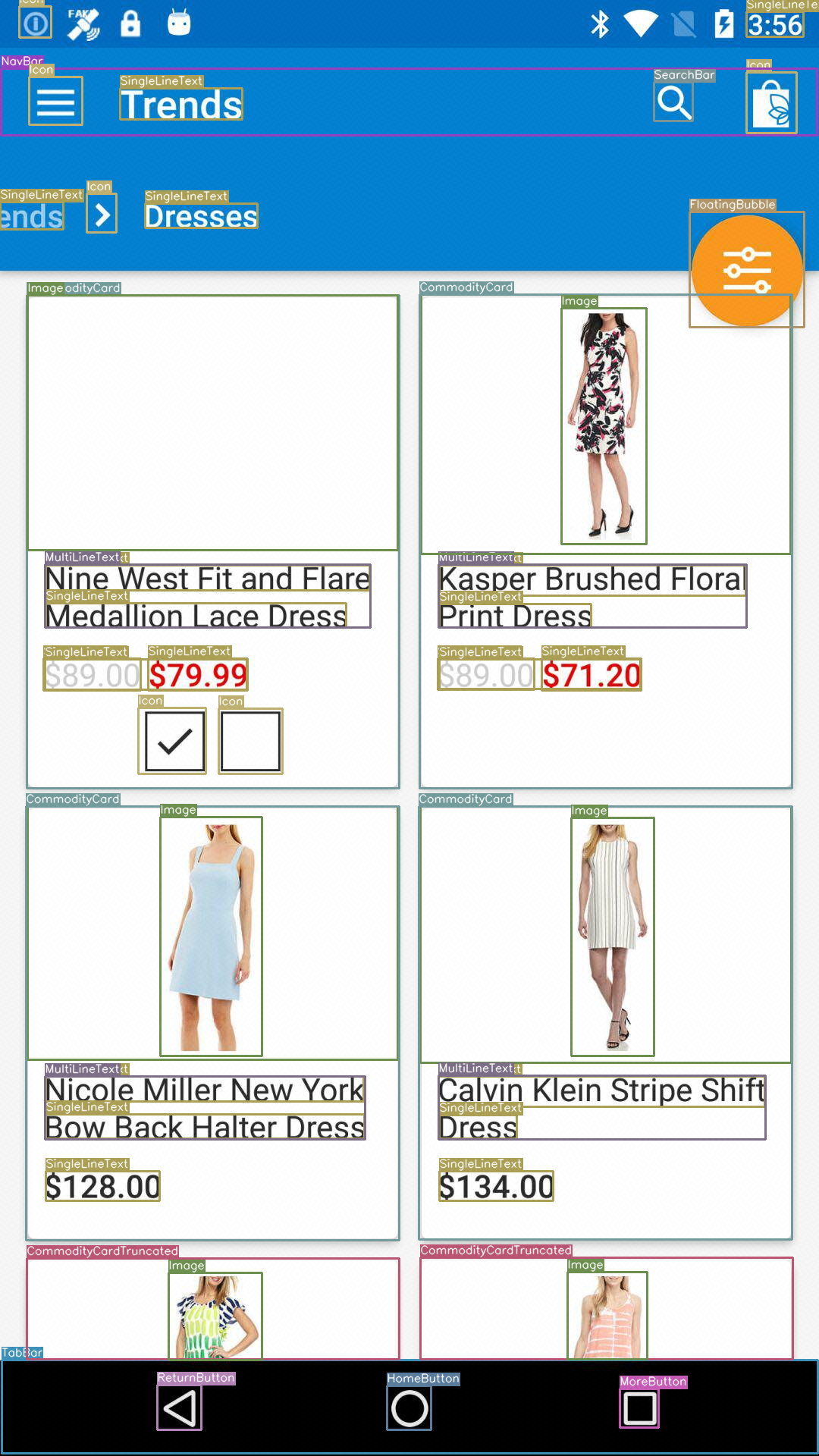}
\\[-0.2em]\normalsize RICO
\end{minipage}\hfill
\begin{minipage}[t]{0.235\textwidth}
\centering
\includegraphics[height=1.55in]{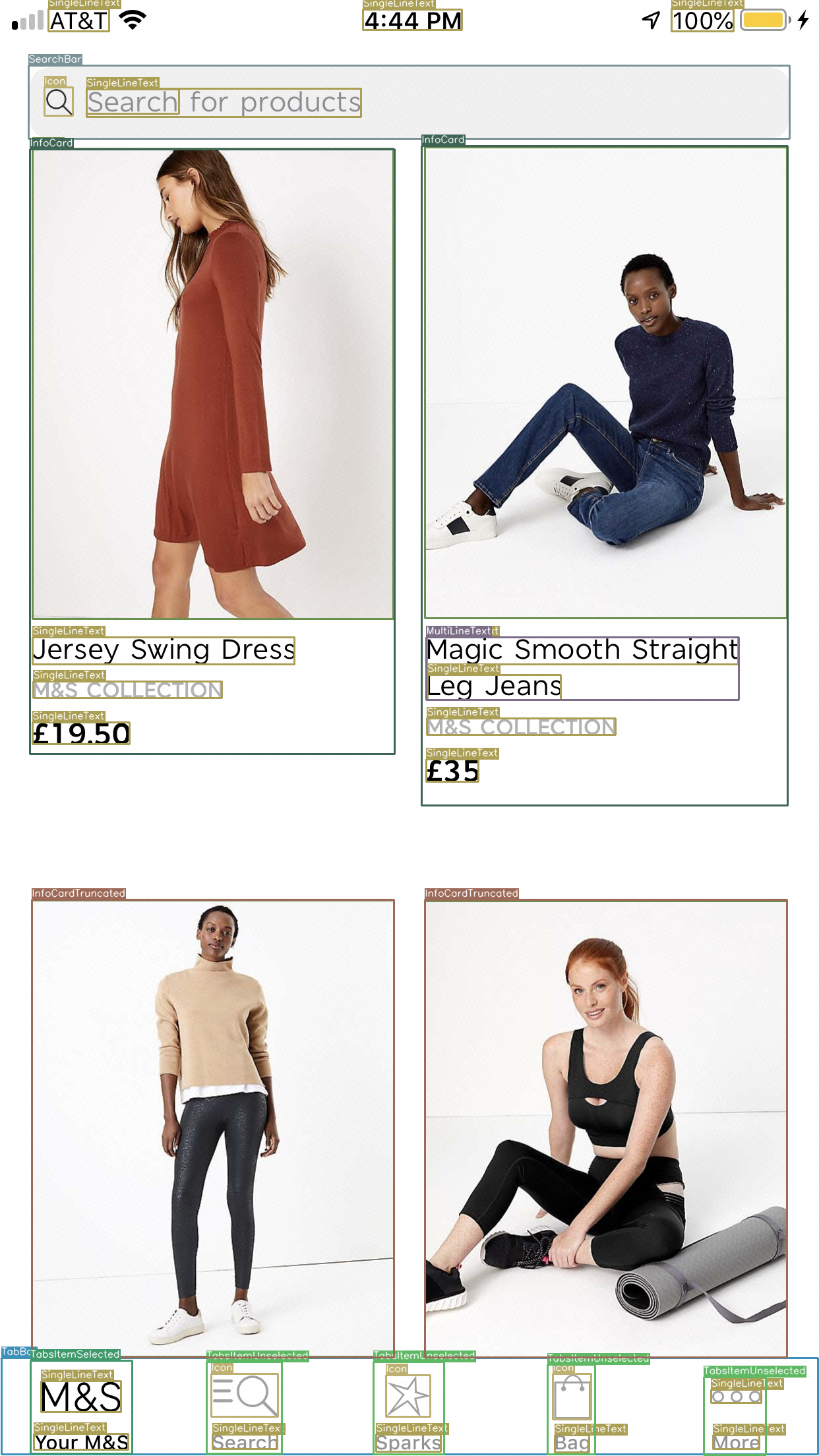}
\\[-0.2em]\normalsize VINS
\end{minipage}\hfill
\begin{minipage}[t]{0.235\textwidth}
\centering
\includegraphics[height=1.55in]{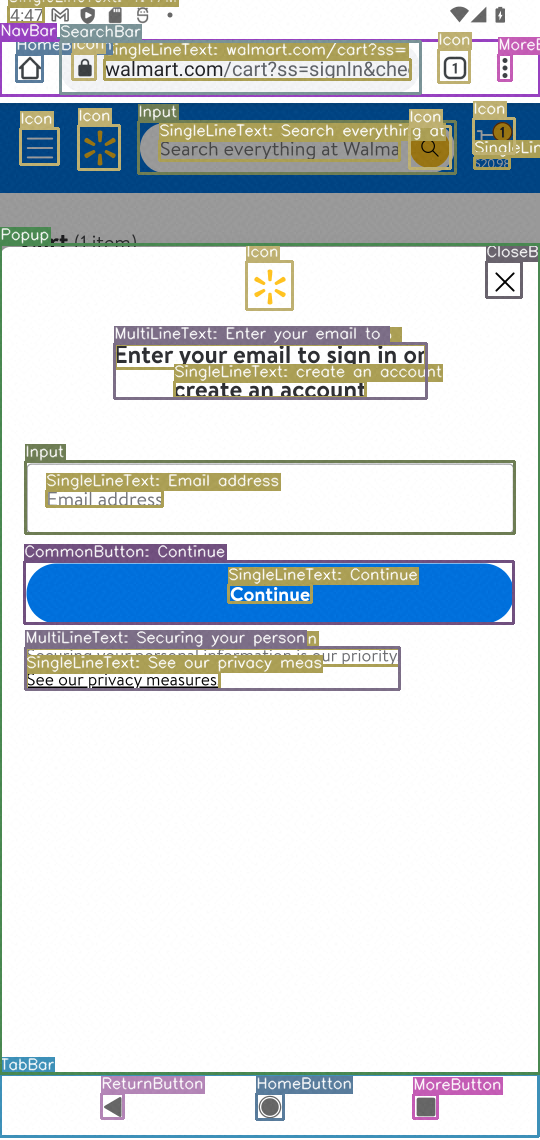}
\\[-0.2em]\normalsize AITW
\end{minipage}\hfill
\begin{minipage}[t]{0.235\textwidth}
\centering
\includegraphics[height=1.55in]{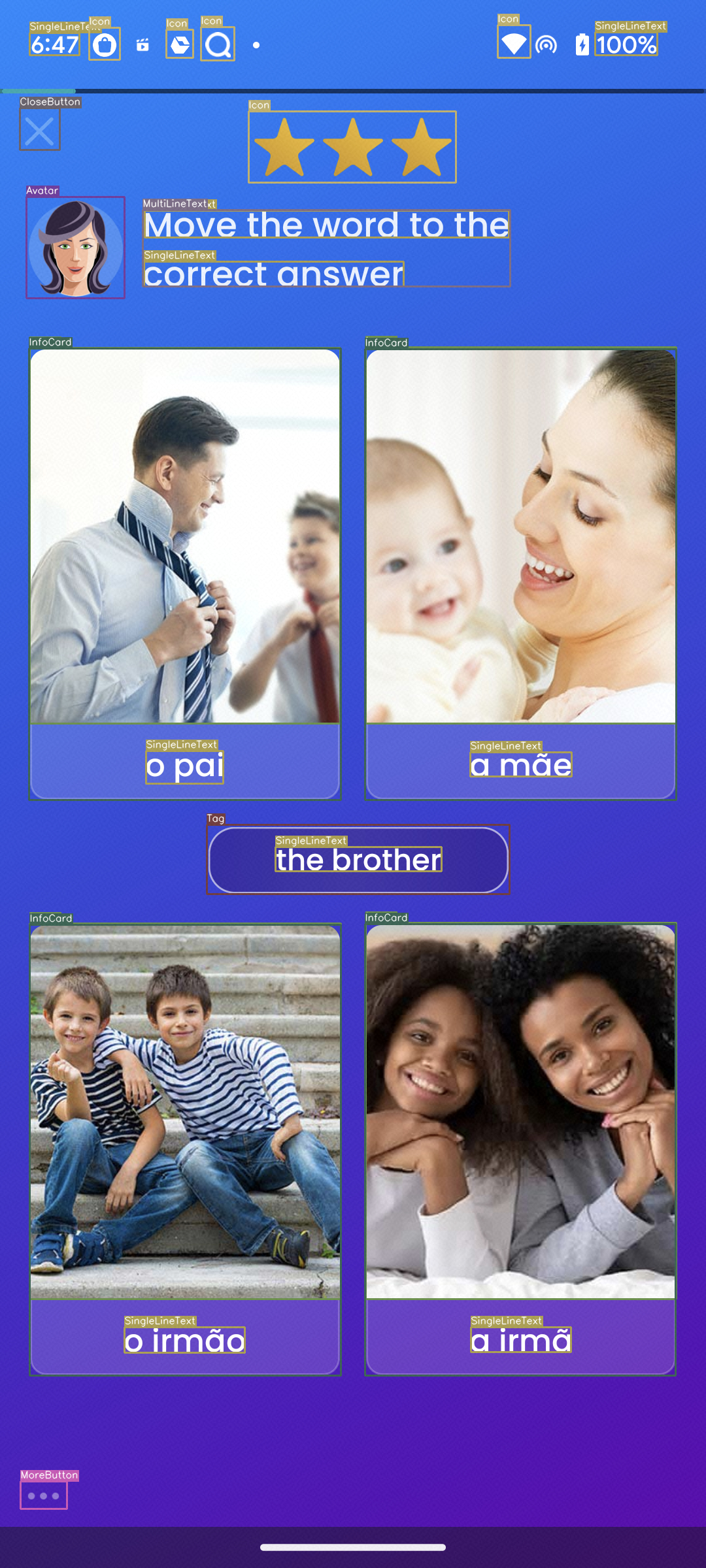}
\\[-0.2em]\normalsize AC
\end{minipage}

\vspace{0.4em}

\begin{minipage}[t]{0.235\textwidth}
\centering
\includegraphics[height=1.55in]{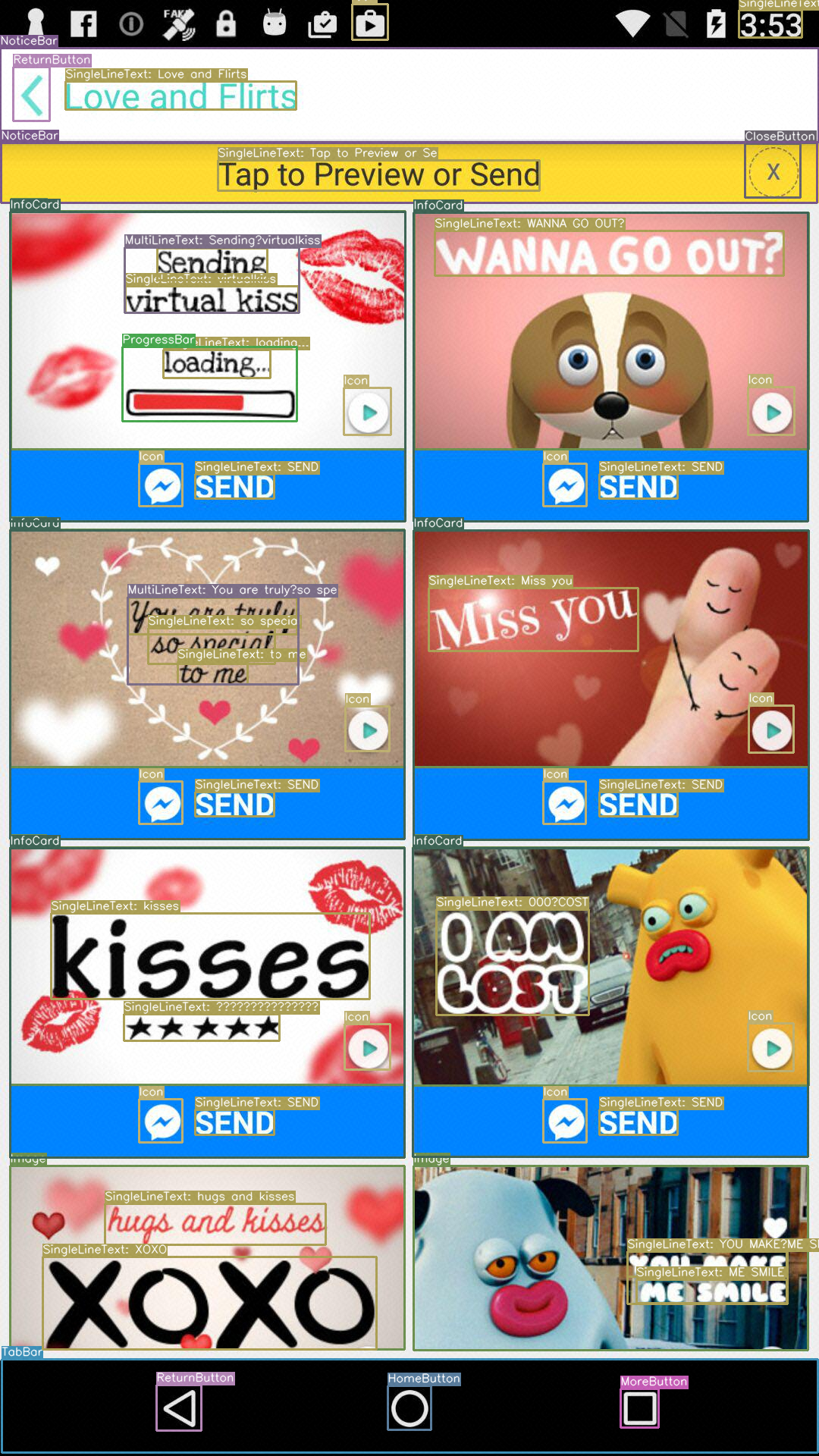}
\\[-0.2em]\normalsize RICO
\end{minipage}\hfill
\begin{minipage}[t]{0.235\textwidth}
\centering
\includegraphics[height=1.55in]{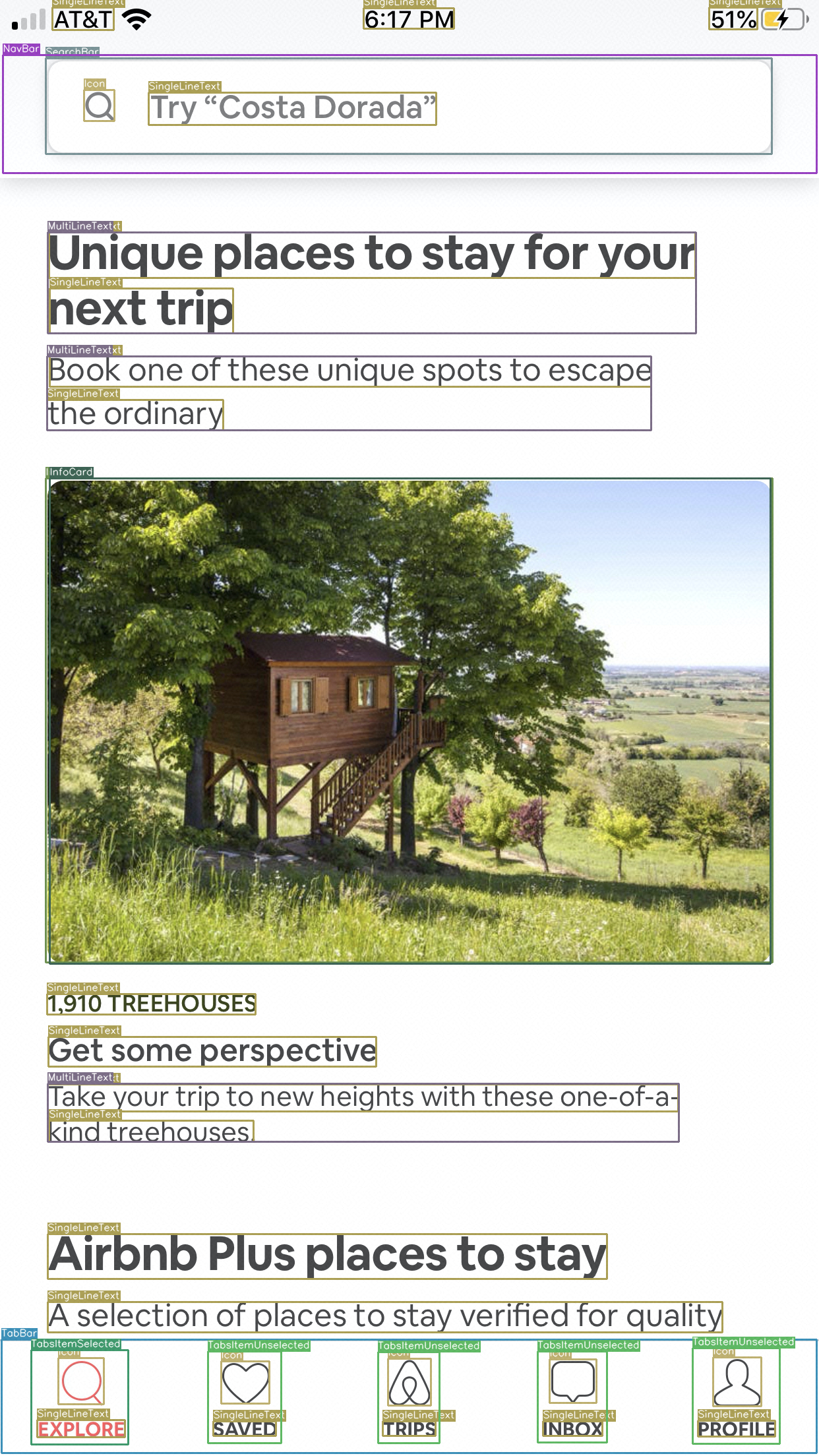}
\\[-0.2em]\normalsize VINS
\end{minipage}\hfill
\begin{minipage}[t]{0.235\textwidth}
\centering
\includegraphics[height=1.55in]{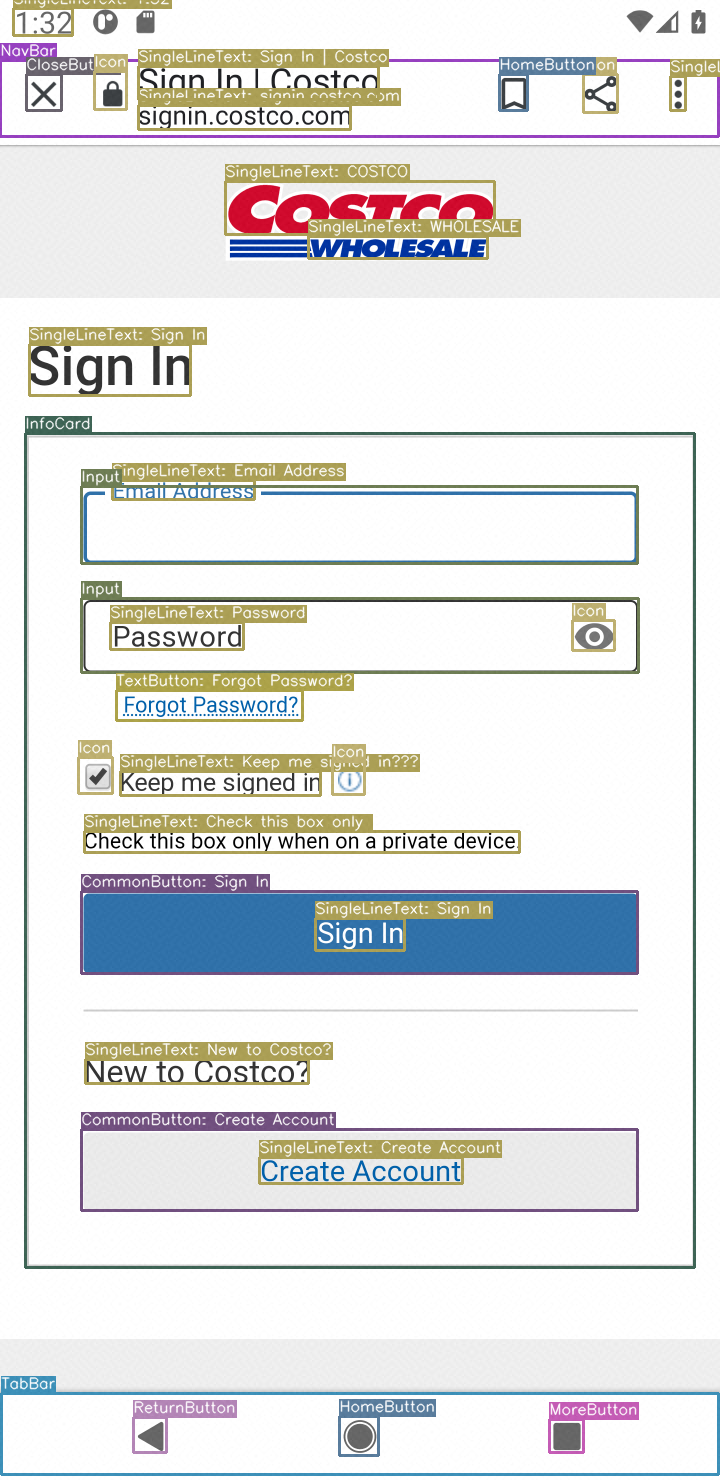}
\\[-0.2em]\normalsize AITW
\end{minipage}\hfill
\begin{minipage}[t]{0.235\textwidth}
\centering
\includegraphics[height=1.55in]{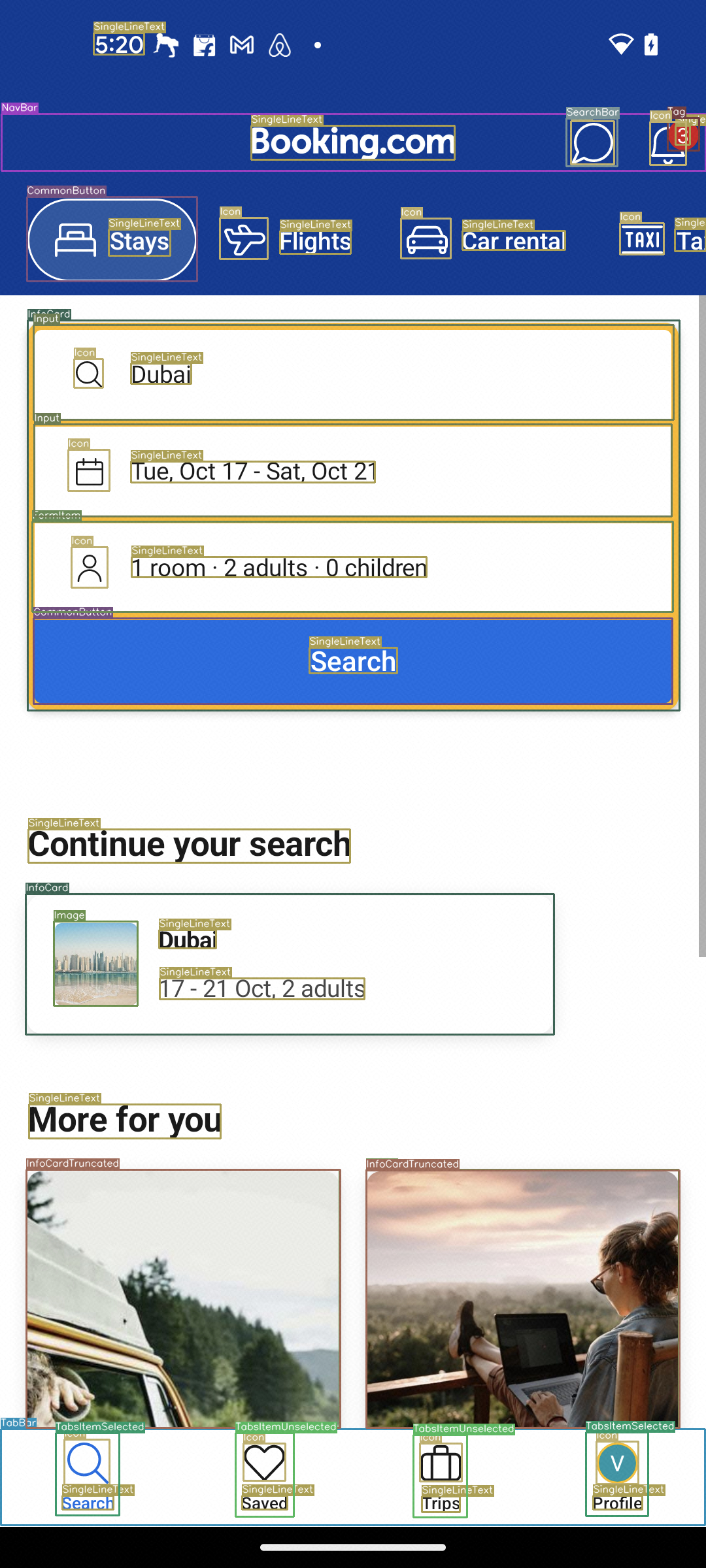}
\\[-0.2em]\normalsize AC
\end{minipage}
\caption{Visualization examples of dataset annotations (2$\times$4). The order in each row is RICO, VINS, AITW, and AC. Overlaid annotations illustrate both UI element categories and bounding-box locations.}
\label{fig:dataset_demo_grid}
\end{figure}

To illustrate annotation quality, Figure~\ref{fig:dataset_demo_grid} shows representative visualization results from the four data sources (RICO, VINS, AITW, and AC). Each sample overlays \textbf{element category} and \textbf{bounding-box location} annotations, demonstrating consistent fine-grained type and position labeling across diverse UI styles.

\subsection{Distribution Balance Analysis}

\begin{figure}[htbp]
\centering
\includegraphics[width=0.7\linewidth]{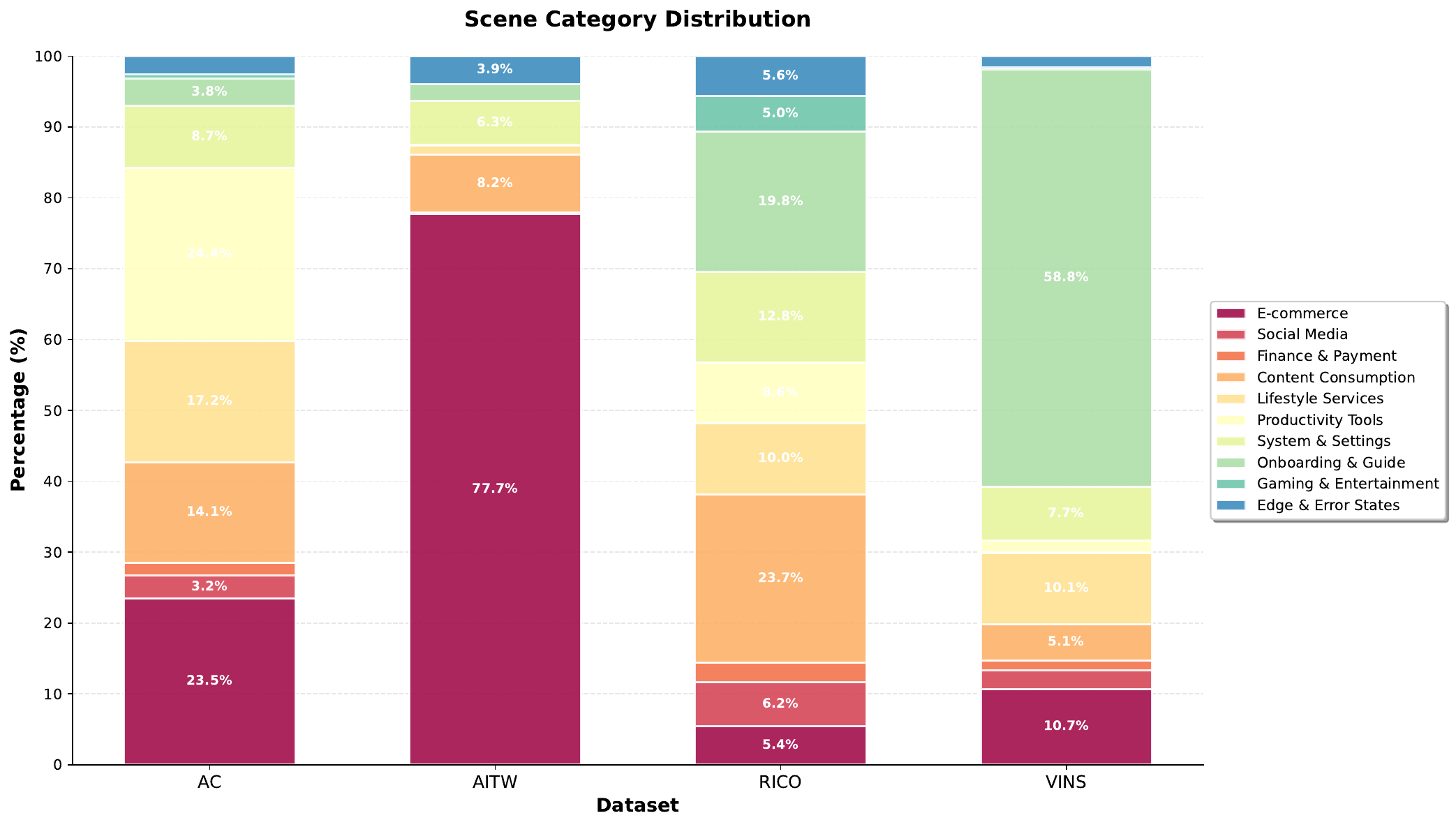}
\caption{Scene category distribution across AC, AITW, RICO, and VINS. The four sources exhibit complementary scene biases that improve overall diversity.}
\label{fig:scene_distribution_appendix}
\end{figure}

Our final dataset is constructed by \textbf{mixing screenshots from all four sources} (RICO, VINS, AITW, and AC), rather than relying on any single dataset. We further analyze the merged data distribution from two complementary perspectives: \textbf{scene category composition} and \textbf{sample complexity}. Figure~\ref{fig:scene_distribution_appendix} shows that the four sources have clearly different scene priors (e.g., AITW is dominated by social-media-like pages, while VINS contains a large proportion of onboarding/guide pages). This heterogeneity is precisely why multi-source mixing is beneficial: it broadens semantic coverage and reduces the risk that the model overfits to one dominant app scenario.

\begin{figure}[htbp]
\centering
\includegraphics[width=0.7\linewidth]{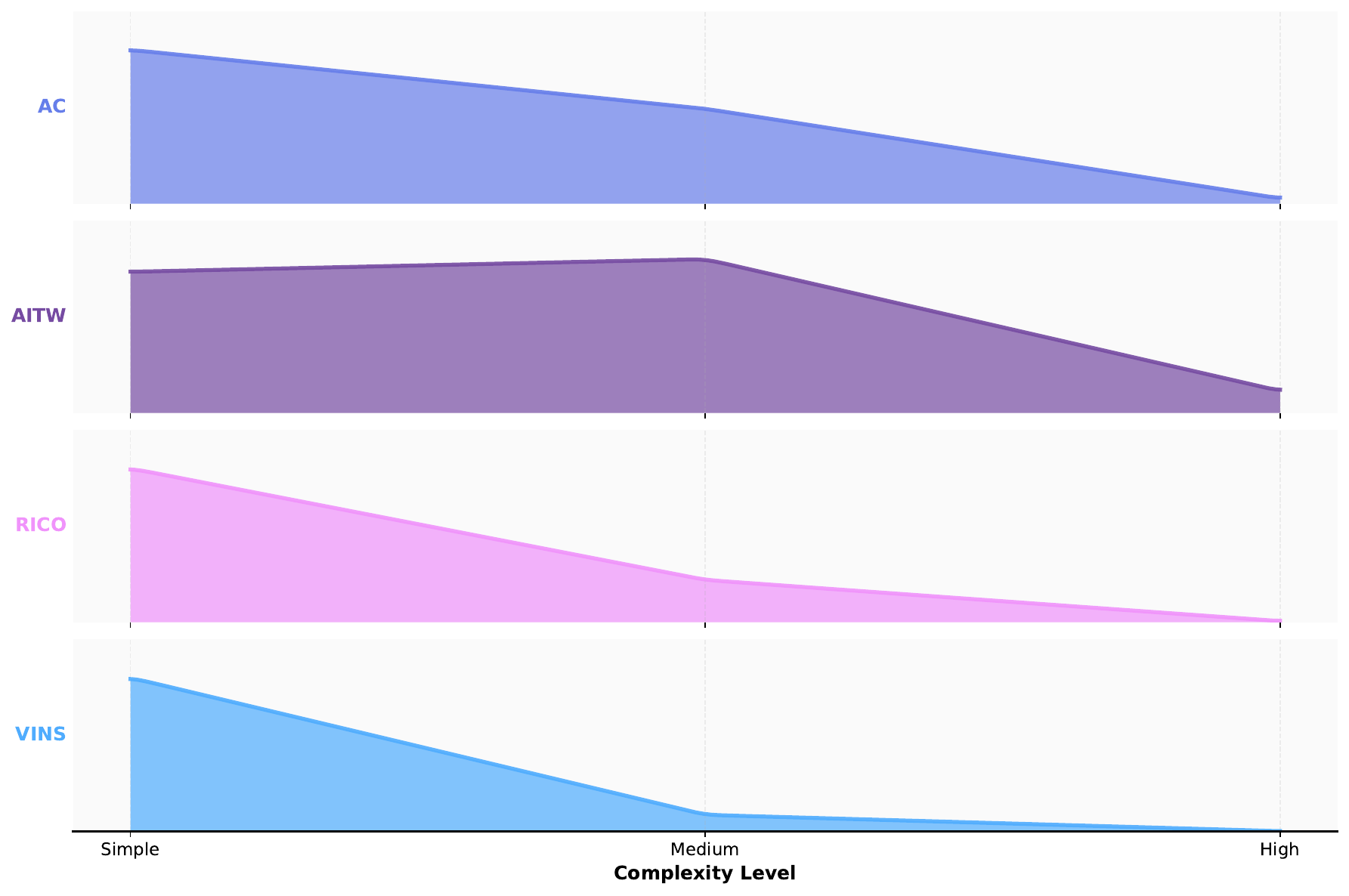}
\caption{Complexity-level distribution (Simple/Medium/High) for each source dataset. Most samples are simple/medium, with source-dependent medium/high tails.}
\label{fig:complexity_distribution_appendix}
\end{figure}

Figure~\ref{fig:complexity_distribution_appendix} presents complexity-level statistics (Simple/Medium/High) across datasets. The single-source distributions are imbalanced: some datasets are skewed toward simpler layouts, while others retain a broader medium/high tail. After mixing, the resulting corpus better covers the full complexity range, from clean low-density screens to dense and cluttered interfaces. In practice, this helps mitigate the complexity bias of individual datasets and provides more stable supervision for both easy cases (format and basic localization consistency) and hard cases (fine-grained category discrimination and dense grounding under crowded layouts).

In summary, the four-dataset mixture serves as a practical balancing strategy: it compensates for \textbf{scene-type imbalance} and \textbf{complexity imbalance} in any single source, yielding a more diverse and better-conditioned training distribution for unified UI understanding and generation.

\FloatBarrier

%% file: appendices/03_dsl_examples.tex
\section{Additional JSON DSL Examples}
\label{app:dsl}
The corpus construction pipeline and node schema are given in Section~\ref{sec:data}. This appendix provides a rendered dataset example and a longer textual specification.

\subsection{DSL Examples}

\noindent\textbf{Simple example.}
A minimal single-button node:
\begin{small}
\begin{quote}
\ttfamily
\{ "type": "Tag", "name": "a",\\
\ \ "className": "inline-flex items-center justify-center font-medium bg-blue-600 text-white rounded-3xl py-2 px-4",\\
\ \ "params": \{ "textContent": \{ "bindType": "Static", "value": "Learn More" \} \}\\
\}
\end{quote}
\end{small}

\begin{figure}[htbp]
\centering
\includegraphics[width=0.95\linewidth]{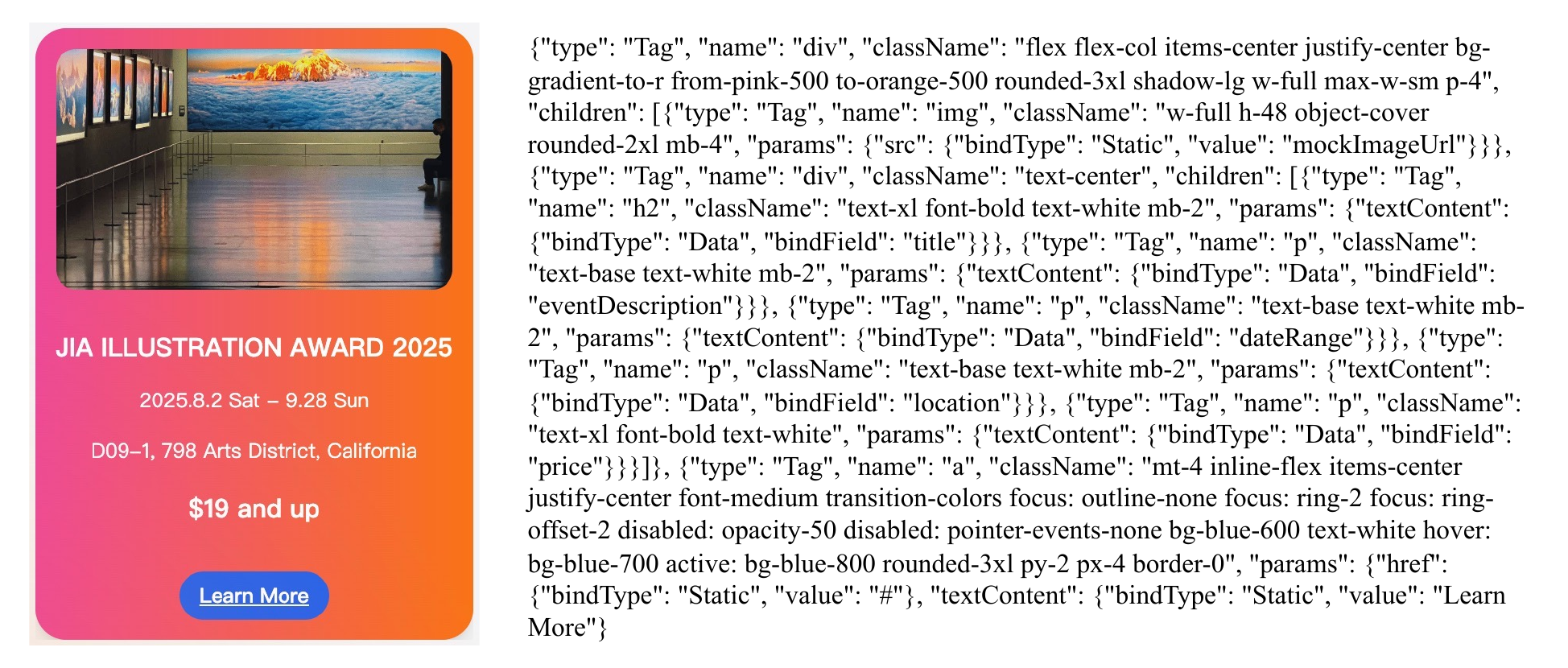}
\caption{A real example from our generation dataset. Left: rendered UI card. Right: corresponding JSON DSL (truncated). The full DSL contains nested layout, dynamic data bindings, and Tailwind styling tokens.}
\label{fig:dsl_complex}
\end{figure}

\noindent\textbf{Complex example from dataset.}
Figure~\ref{fig:dsl_complex} presents a real example from our generation dataset. The left panel shows the rendered UI card, and the right panel shows the corresponding DSL (truncated for readability). The full DSL comprises a root \texttt{div} with nested children including an image container, text hierarchy, and an action button, totaling 499 tokens (counted with OpenAI's tokenizer).

\FloatBarrier

%% file: appendices/04_prompts.tex
\section{Prompt Templates}
\label{app:prompts}

For completeness, we list the prompt templates used in SFT, evaluation, and reward-related judging.

\subsection{SFT Task Prompts}

We provide the prompt templates used during supervised fine-tuning for each task.

\noindent\textbf{Referring prompts.}
\begin{small}
\begin{quote}
\begin{itemize}
  \item ``Describe the region \{box\}.''
  \item ``Describe the categories and properties of the UI in area \{box\}.''
  \item ``List the types and characteristics of UI elements in region \{box\}.''
  \item ``Provide details on the UI types and their attributes within area \{box\}.''
  \item ``Explain the categories and features of the UI components in region \{box\}.''
  \item ``Summarize the types and properties of UI present in area \{box\}.''
\end{itemize}
\end{quote}
\end{small}

\noindent\textbf{Dense grounding prompts.}
\begin{small}
\begin{quote}
\begin{itemize}
  \item ``Detect all UI elements in the image and return their categories and bounding boxes in JSON format.''
  \item ``List all the \{category\} items.''
  \item ``Please list all the \{category\} in the image.''
  \item ``Identify every \{category\} present in this picture.''
  \item ``List all \{category\} shown in the image.''
  \item ``Enumerate each \{category\} in the provided photo.''
  \item ``What are all the \{category\} in this image? Please enumerate.''
  \item ``Can you list every \{category\} appearing in this image?''
  \item ``Give a detailed list of every \{category\} in the picture.''
\end{itemize}
\end{quote}
\end{small}

\noindent\textbf{GUI grounding prompts.}
\begin{small}
\begin{quote}
\begin{itemize}
  \item ``Output the bounding box corresponding to the following instruction: \{instruction\}. The output should just be the coordinates of a box, in the format [x0, y0, x1, y1]. If the task is infeasible, output [-1, -1, -1, -1].''
\end{itemize}
\end{quote}
\end{small}

\noindent\textbf{Generation prompts.}
\begin{small}
\begin{quote}
\begin{itemize}
  \item ``Design a UI card to showcase the details of \{ui\_description\}.''
  \item ``Design a concise, clear, and visually appealing UI card to display \{ui\_description\}.''
  \item ``Please design a UI card displaying \{ui\_description\}.''
  \item ``Please generate a UI card to guide users to the \{ui\_description\}.''
  \item ``Design a UI card to display \{ui\_description\}. The following is the mock data for the UI card. \{mock\_data\}.''
\end{itemize}
\end{quote}
\end{small}

\subsection{LLM-as-a-Judge Scoring Prompt}
\label{app:judge_prompt}

The following prompt is used for absolute quality scoring in our generation reward:

\begin{small}
\begin{quote}
You are an advanced AI model equipped with OCR and image understanding capabilities, capable of analyzing visual elements in detail.

Your task is to assess two UI images and output a score between 0 and 10 for each of the following questions.

If the answer to a question is a definite YES, output a score of 10, signifying perfect similarity. Conversely, a definite NO should yield a score of 0, indicating no similarity.

For answers that fall in between, assign a score accordingly, where a higher number indicates a greater degree of similarity. Only provide the numerical score for each question, without any additional text.

Example contexts are provided for clarity. Examples provide the idea, but you can output any number in 0--10 range accordingly. Only output a comma separated list containing 4 numbers. DO NOT give score of 10 for any category unless otherwise the two images are identical.

1. \textbf{Color \& Aesthetic Design} (Score: 0--10):
How closely do the color schemes of the two UI images align in terms of background and text colors? Evaluate the similarity in hues, saturation, and overall color aesthetics. Is the overall aesthetic appeal (modern, minimalistic, traditional, etc.) similar on both UI images?
(e.g., A score of 10 for perfectly matching color schemes and identical aesthetics, including identical hues and saturation levels, 6 for similar color palettes and styles with minor variations, and 0 for starkly different color schemes and aesthetics that create entirely different visual impacts.)

2. \textbf{Visual Structure \& Alignment} (Score: 0--10):
Does the placement of elements like images, buttons, and text boxes aligned similarly on both images?
Do the sizes and aspect ratios of images, buttons, and text boxes appear consistent across both images?
Do both UI images exhibit a similar level of visual harmony and balance in their design?
(e.g., A score of 10 for identical visual structure and alignment, 5 for similar but not exact consistent, and 0 for completely different structure.)

3. \textbf{Textual and Content Consistency} (Score: 0--10):
Do the font type, size, style, and weight of two UI images similar?
Do the words and sentences match between the two UI images?
(e.g., A score of 10 for complete uniformity in font type, size, style, weight across both UI images and identical text, 5 for consistency in font type and size but variations in style or weight, and 0 for wide disparities in font type, size, style, or weight, leading to a distinctly different textual appearance and content.)

4. \textbf{Interface Usability \& Interactivity} (Score: 0--10):
Do the user interface elements (like menus, buttons, and forms) share a clear and usable design language?
Are interactive elements such as buttons, inputs, and navigation affordances visually identifiable and appropriately sized?
(e.g., A score of 10 for clear and usable interactive elements, 6 for minor usability issues, and 0 for unclear or unusable UI designs.)
\end{quote}
\end{small}

\subsection{Contrastive Preference Ranking (CPR) Prompt}
\label{app:cpr_prompt}

For pairwise comparison in CPR, we present two rendered UI images side-by-side and use the following prompt. To control for positional bias, we evaluate both orderings (A-left/B-right and B-left/A-right) and aggregate results.

\begin{small}
\begin{quote}
You are an expert UI designer. You are shown two UI images, Image A (left) and Image B (right), both generated from the same requirement. Compare them along the following four dimensions and for each dimension output which is better: A, B, or Tie.

1. \textbf{Color \& Aesthetic Design}: Which UI has a more appealing and harmonious color scheme? Consider hues, saturation, gradients, and overall visual warmth.

2. \textbf{Visual Structure \& Alignment}: Which UI has better element alignment, spacing, and layout balance? Consider the hierarchical organization and visual flow.

3. \textbf{Textual \& Content Consistency}: Which UI has more appropriate text styling (font type, size, weight) and more accurate content relative to the requirement?

4. \textbf{Interface Usability \& Interactivity}: Which UI has clearer interactive elements and better overall usability? Consider button clarity, tap target sizes, and navigational affordances.

Output format: a comma-separated list of 4 values, each being ``A'', ``B'', or ``Tie''. Example: A, Tie, B, A
\end{quote}
\end{small}

\subsection{General-Purpose MLLM Evaluation Prompts}
\label{app:eval_prompts}

For evaluating general-purpose MLLMs on our benchmark, we use standardized prompts with the full 80-category list.

\noindent\textbf{Referring evaluation.}
\begin{small}
\begin{quote}
Your task is to describe the given region in the image from three perspectives. First, decide the category of the given region from the following 80 categories: \{full\_category\_list\}. Second, describe the text content in the given region. Third, output the color of the text in hexadecimal format (e.g., \#414141).

Output in JSON format: \{"category": "...", "text": "...", "text\_color": ["..."]\}. Do not return any other content.
\end{quote}
\end{small}

\noindent\textbf{Dense grounding evaluation.}
\begin{small}
\begin{quote}
Your task is to detect all UI elements in the image. Classify each element using the following 80 categories: \{full\_category\_list\}. Output the bounding box in [x\_min, y\_min, x\_max, y\_max] format.

Return in JSON format: [\{"type": "...", "box": [x\_min, y\_min, x\_max, y\_max]\}, ...]. Do not create new categories and do not return any other content.
\end{quote}
\end{small}

\noindent\textbf{GUI grounding evaluation.}
\begin{small}
\begin{quote}
Output the bounding box corresponding to the following instruction: \{instruction\}. The output should just be the coordinates of a box, in the format [x0, y0, x1, y1]. If the task is infeasible, output [-1, -1, -1, -1].
\end{quote}
\end{small}

\FloatBarrier

%% file: appendices/05_dense_evaluation.tex
\section{Dense Grounding Evaluation Protocol}
\label{app:dense_eval}

Because dense grounding outputs are variable-length structured predictions, we use a dedicated set-matching evaluation protocol rather than confidence-based detection metrics.

Dense grounding evaluation requires matching a variable-length set of predicted bounding boxes to a variable-length set of ground-truth boxes, which is fundamentally a set-matching problem. Unlike standard object detection benchmarks that rely on confidence-score-based AP metrics, MLLMs do not produce per-box confidence scores. We therefore adopt a set-matching protocol:

\noindent\textbf{Step 1: Parsing and filtering.}

We parse the model's structured JSON output to extract predicted bounding boxes and their associated category labels. Predictions with malformed coordinates or categories outside the 80-class taxonomy are discarded.

\noindent\textbf{Step 2: Hungarian matching.}

For each image, we compute a cost matrix $\mathbf{C} \in \mathbb{R}^{M \times N}$ between $M$ predicted boxes and $N$ ground-truth boxes, where each entry is:
\begin{equation}
C_{j,i} = \begin{cases}
1 - \text{IoU}(\hat{b}_j, b_i) & \text{if } \hat{c}_j = c_i \\
1.0 & \text{otherwise}
\end{cases}
\end{equation}
Here $\hat{c}_j$ and $c_i$ denote the predicted and ground-truth category labels. We then apply the Hungarian algorithm to find the optimal one-to-one assignment that minimizes total cost.

\noindent\textbf{Step 3: IoU thresholding.}

A matched pair $(\hat{b}_j, b_i)$ is counted as a true positive only if both (a) the categories match ($\hat{c}_j = c_i$) and (b) $\text{IoU}(\hat{b}_j, b_i) \geq \tau$, where we use $\tau = 0.5$ as the default threshold. Unmatched predictions are false positives; unmatched ground-truth boxes are false negatives.

\noindent\textbf{Step 4: Metric computation.}

We compute:
\begin{itemize}
    \item \textbf{Precision} $= \text{TP} / (\text{TP} + \text{FP})$
    \item \textbf{Recall} $= \text{TP} / (\text{TP} + \text{FN})$
    \item \textbf{F1}: computed per sample from its precision and recall, then averaged over the evaluation set.
    \item \textbf{mIoU}: computed over category-correct true-positive matched pairs, reflecting localization quality after semantic matching.
\end{itemize}

This protocol ensures fair comparison across models with different output formats.
\FloatBarrier

%% file: appendices/06_understanding.tex
\section{Additional Understanding Results}
\label{app:understanding_results}

This section supplements the main paper with additional analyses for referring and grounding, including quantitative diagnostics and qualitative case studies.

\subsection{Referring: Analysis Across Models}
\label{app:referring_analysis}

\begin{figure}[htbp]
\centering
\includegraphics[width=0.95\linewidth]{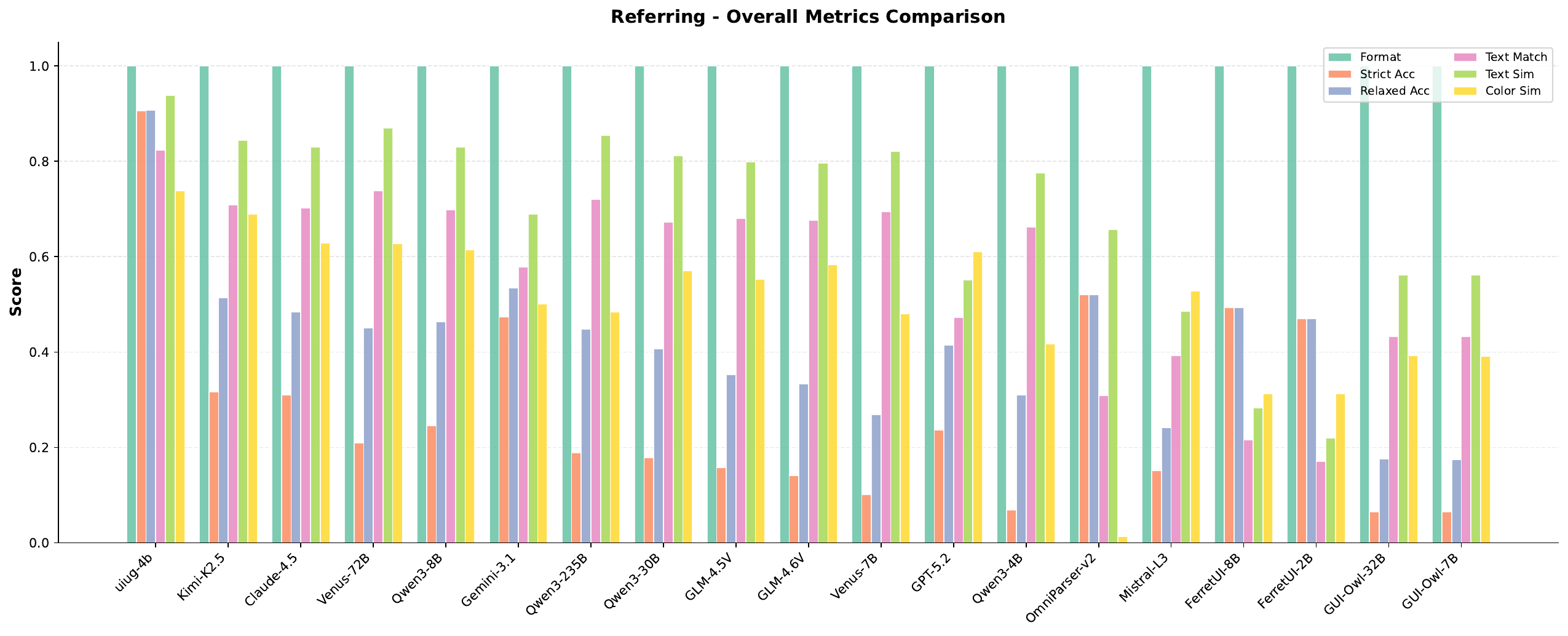}
\caption{Overall metric comparison on the referring task across models.}
\label{fig:referring_overall_appendix}
\end{figure}

Figure~\ref{fig:referring_overall_appendix} compares overall referring metrics across representative open/closed-source and UI-specific models. In addition to strict category accuracy, we report format validity, relaxed category accuracy, text matching/similarity, and color similarity to provide a more complete picture of model behavior.

\subsection{Referring: Strict vs.\ Relaxed Type Matching}
\label{app:strict_relaxed_referring}

Although all evaluated MLLMs are prompted with the full 80-category taxonomy, strict type matching remains challenging in practice. Two factors are prominent: (1) limited instruction-following consistency for constrained category output, and (2) limited fine-grained visual discrimination between semantically adjacent UI types. Therefore, besides strict matching, we also evaluate a \textbf{relaxed matching} mode via post-processing, where visually and semantically similar predicted types are tolerated.

\begin{figure}[htbp]
\centering
\includegraphics[width=0.95\linewidth]{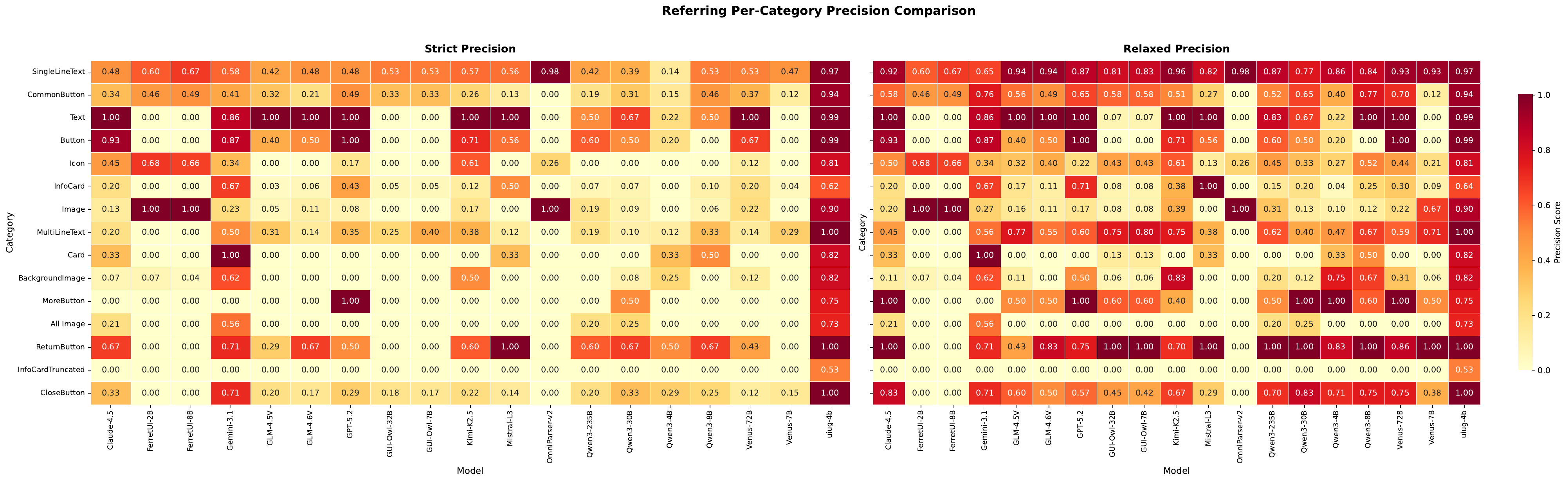}
\caption{Per-category precision comparison under strict and relaxed type matching protocols.}
\label{fig:referring_strict_relaxed_appendix}
\end{figure}

Figure~\ref{fig:referring_strict_relaxed_appendix} reports per-category precision under strict and relaxed settings. The relaxed protocol yields noticeable precision improvements for several general-purpose MLLMs, indicating that many strict-mode errors are near-miss confusions among closely related categories rather than complete perception failures. This analysis helps disentangle category granularity mismatch from core UI understanding capability.

\subsection{Grounding: Analysis Across Models}
\label{app:grounding_analysis}

\begin{figure}[htbp]
\centering
\includegraphics[width=0.95\linewidth]{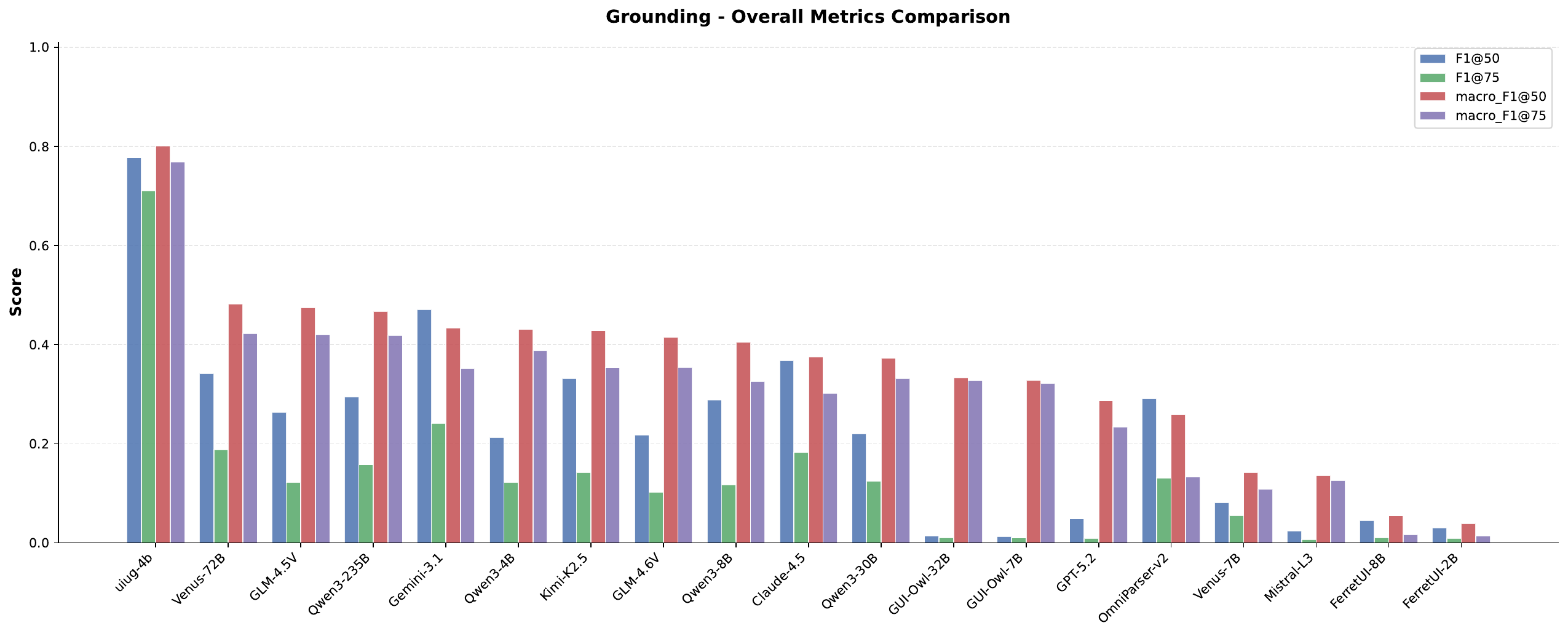}
\caption{Overall grounding metric comparison across models under strict and macro-label protocols (F1@50/F1@75 and macro\_F1@50/macro\_F1@75).}
\label{fig:grounding_overall_appendix}
\end{figure}

We further compare grounding performance across models under both strict fine-grained labels and a relaxed macro-label protocol. Figure~\ref{fig:grounding_overall_appendix} reports four aggregate metrics: \textbf{F1@50}, \textbf{F1@75}, \textbf{macro\_F1@50}, and \textbf{macro\_F1@75}. Here, macro mode uses the broader 9-category taxonomy in our dataset, which is more tolerant to confusions among semantically similar fine-grained types.

Compared with strict fine-grained matching, macro-level evaluation raises scores for multiple general-purpose MLLMs, indicating that part of their errors come from fine-grained type ambiguity rather than complete localization failure. This is consistent with the practical difficulty of dense UI grounding: models may localize a region reasonably but still miss the exact fine category.

\begin{figure}[htbp]
\centering
\includegraphics[width=0.95\linewidth]{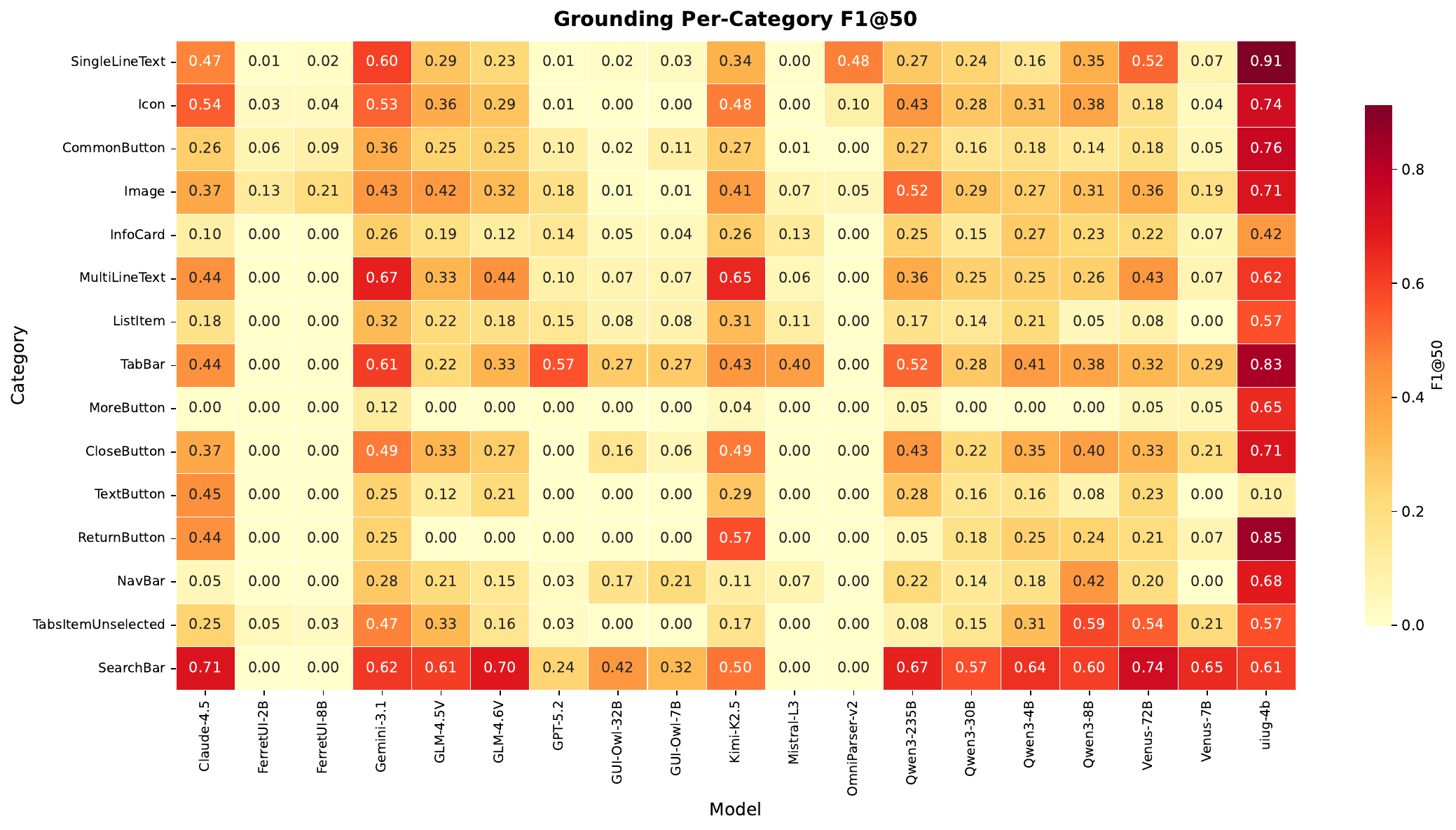}
\caption{Per-category grounding F1@50 comparison across models. Common categories are generally easier, while complex or confusable UI categories remain challenging for most baselines.}
\label{fig:grounding_category_appendix}
\end{figure}

Figure~\ref{fig:grounding_category_appendix} presents per-category F1@50 differences. Stronger grounding models (e.g., Claude~Sonnet~4.5, Gemini~3.1~Pro, and Kimi~K2.5) generally achieve better detection on frequent and visually salient categories such as \textit{Icon}, \textit{Button}, and \textit{Text}. However, performance degrades substantially on complex or easily confused UI categories (e.g., nested card-like structures, fine-grained functional controls, and visually similar component variants). This suggests that the primary bottleneck of current MLLMs is still \textbf{fine-grained UI semantic understanding and category discrimination} under dense layouts, rather than only coarse object localization.

\subsection{Grounding: Visual Comparison of Results}

\begin{figure}[htbp]
\centering
\begin{minipage}[t]{0.136\textwidth}
\centering
\includegraphics[width=\linewidth]{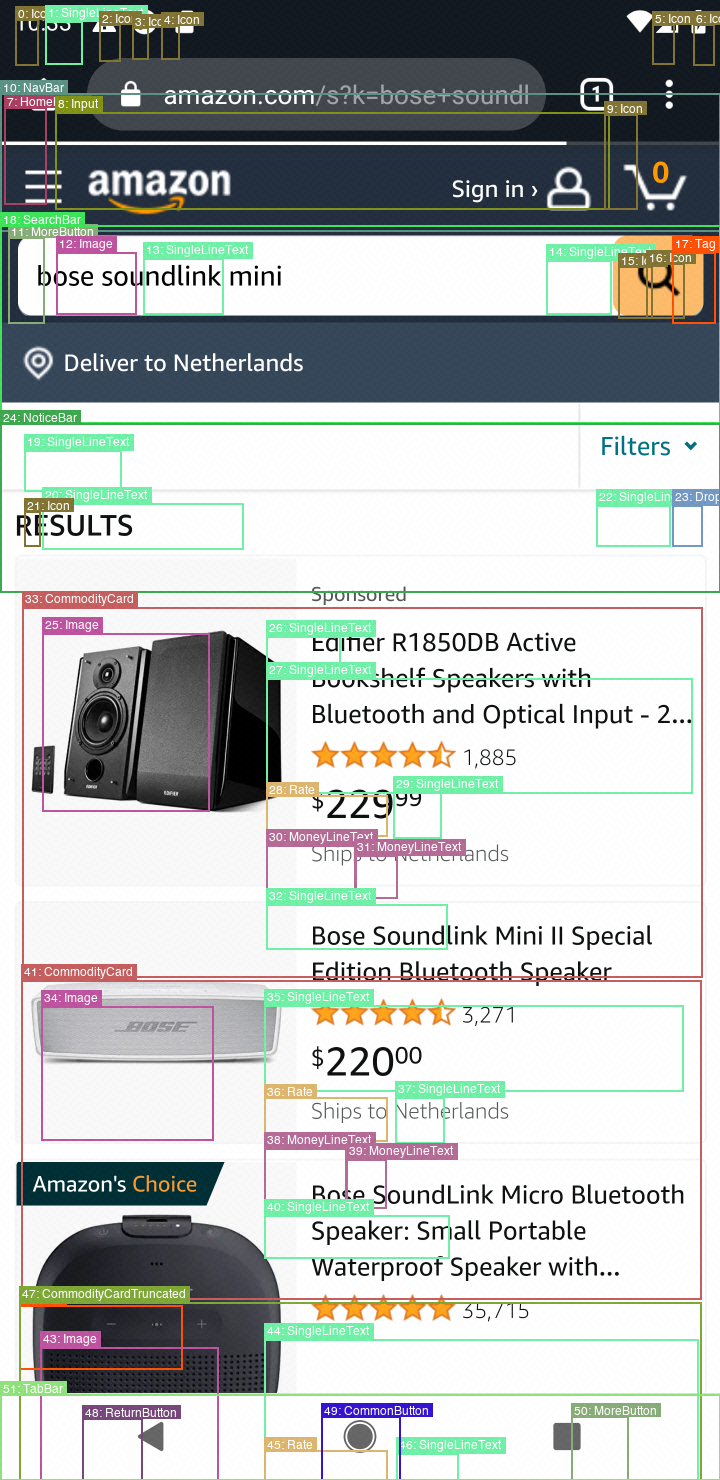}\\[-0.2em]
\scriptsize GPT-5.2
\end{minipage}\hfill
\begin{minipage}[t]{0.136\textwidth}
\centering
\includegraphics[width=\linewidth]{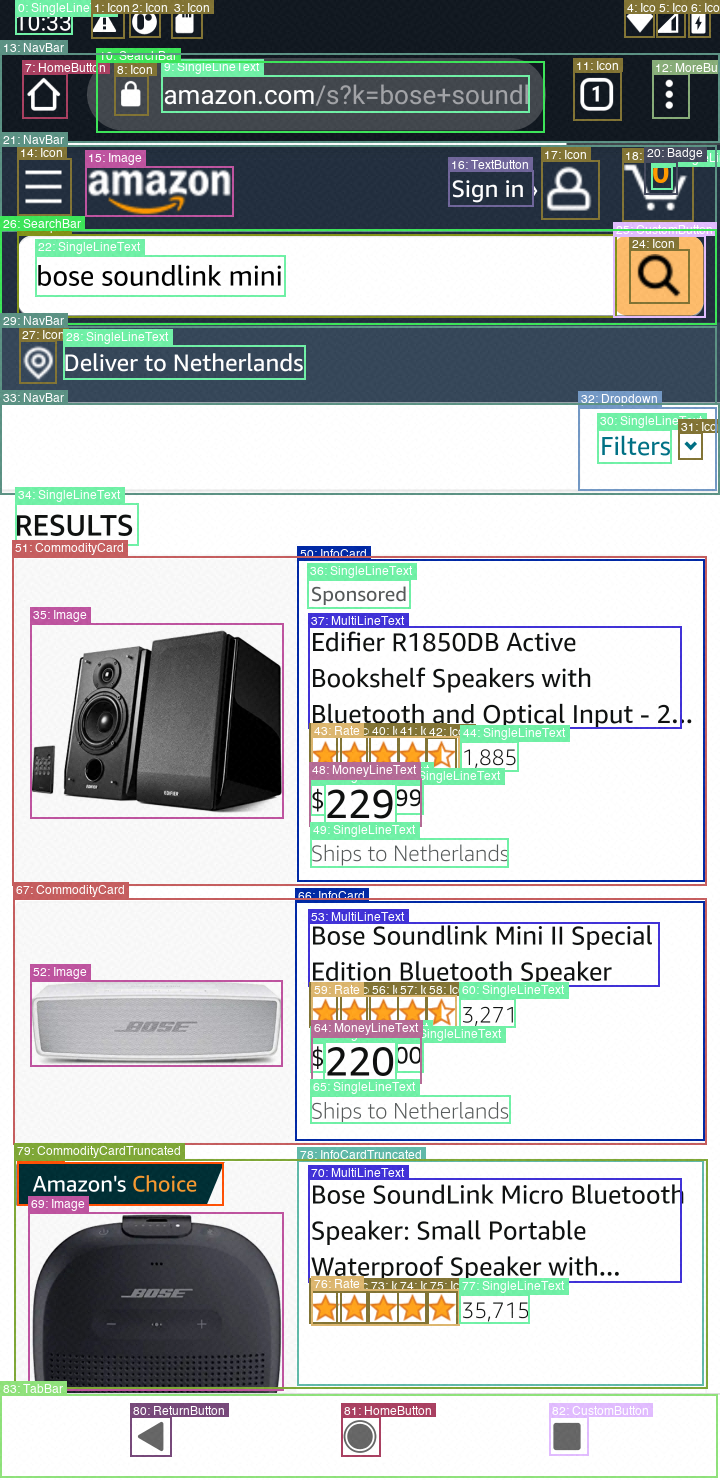}\\[-0.2em]
\scriptsize Gemini~3.1~Pro
\end{minipage}\hfill
\begin{minipage}[t]{0.136\textwidth}
\centering
\includegraphics[width=\linewidth]{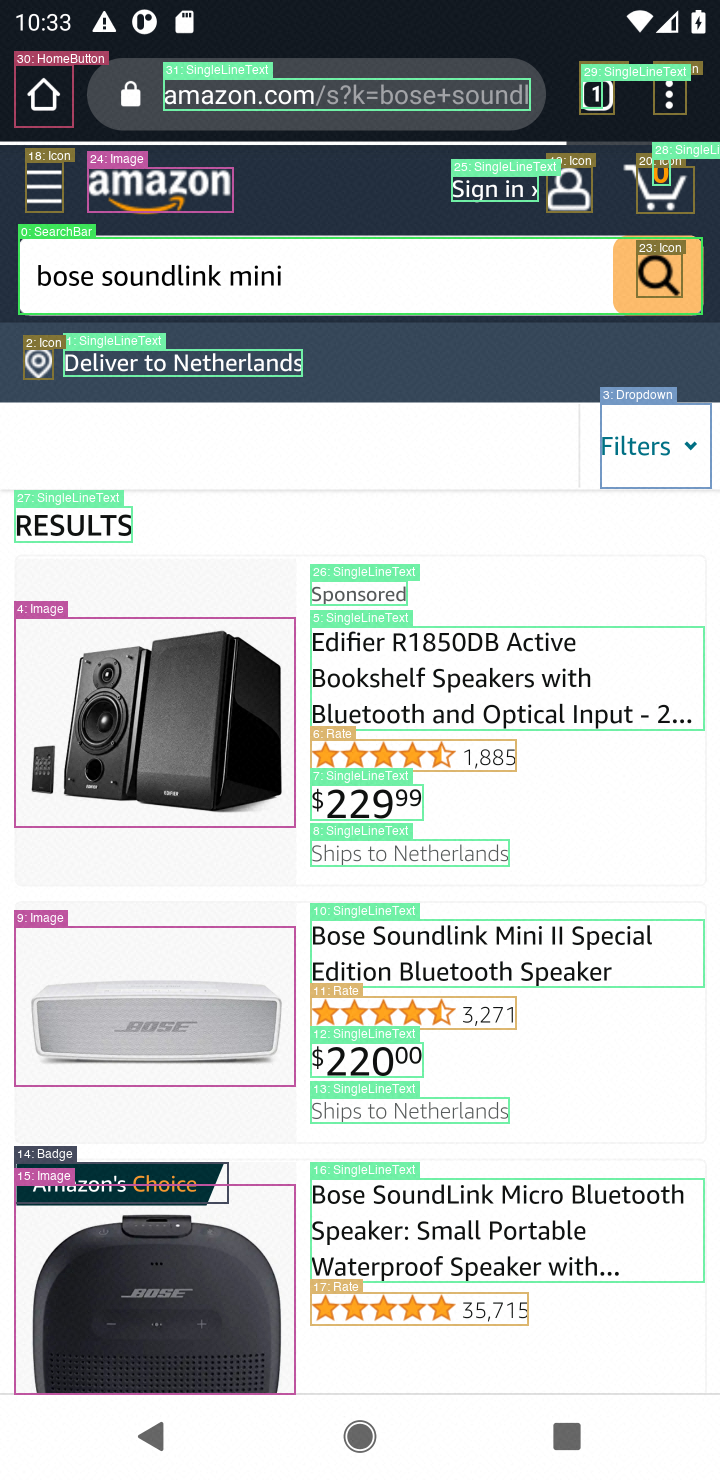}\\[-0.2em]
\scriptsize Claude~Sonnet~4.5
\end{minipage}\hfill
\begin{minipage}[t]{0.136\textwidth}
\centering
\includegraphics[width=\linewidth]{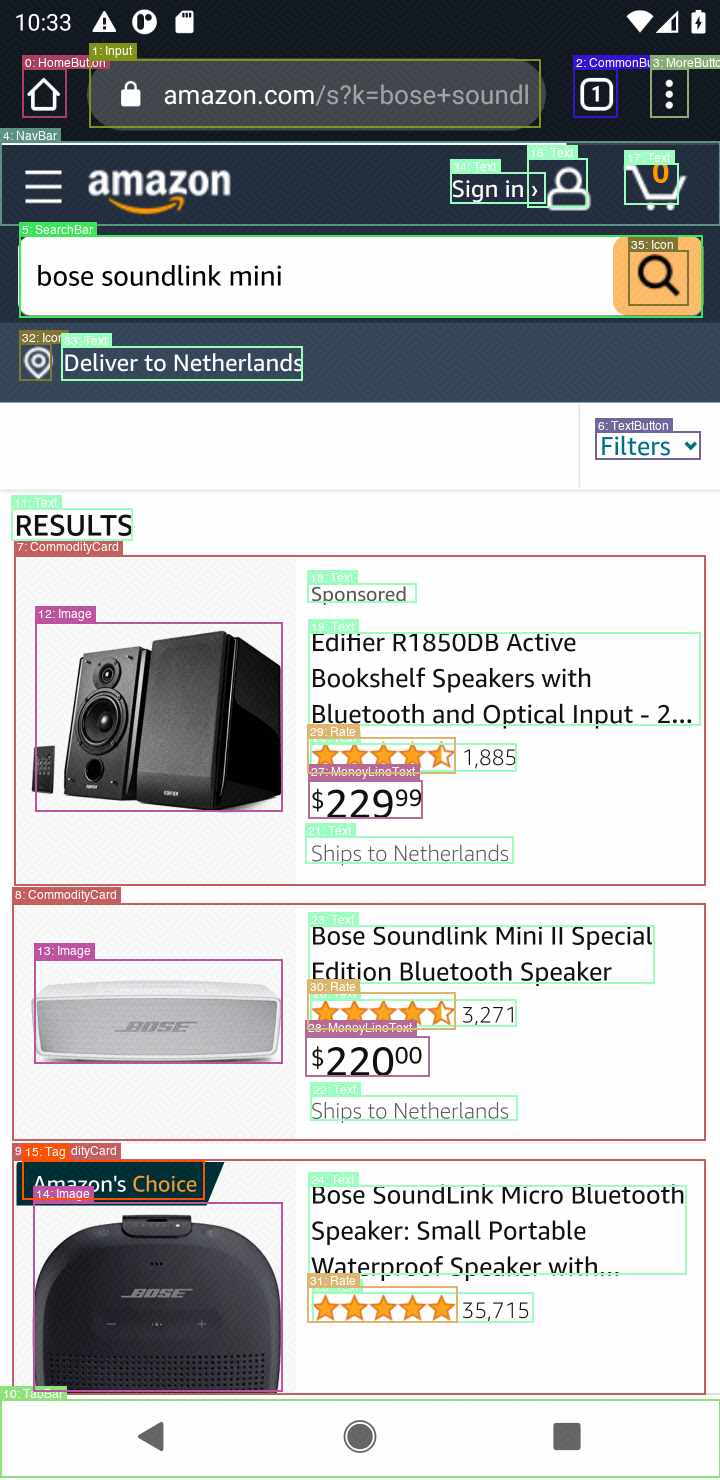}\\[-0.2em]
\scriptsize Qwen3-VL-235B-A22B
\end{minipage}\hfill
\begin{minipage}[t]{0.136\textwidth}
\centering
\includegraphics[width=\linewidth]{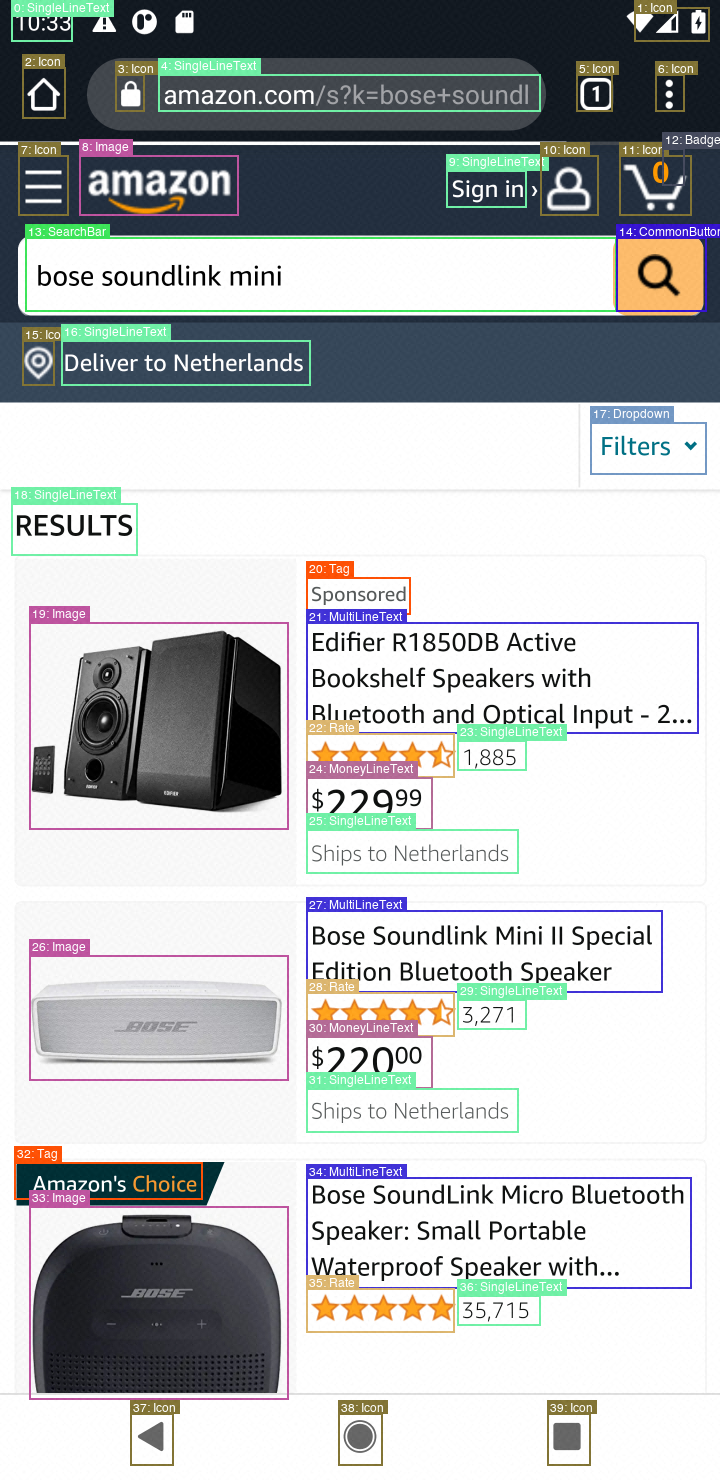}\\[-0.2em]
\scriptsize Kimi~K2.5
\end{minipage}\hfill
\begin{minipage}[t]{0.136\textwidth}
\centering
\includegraphics[width=\linewidth]{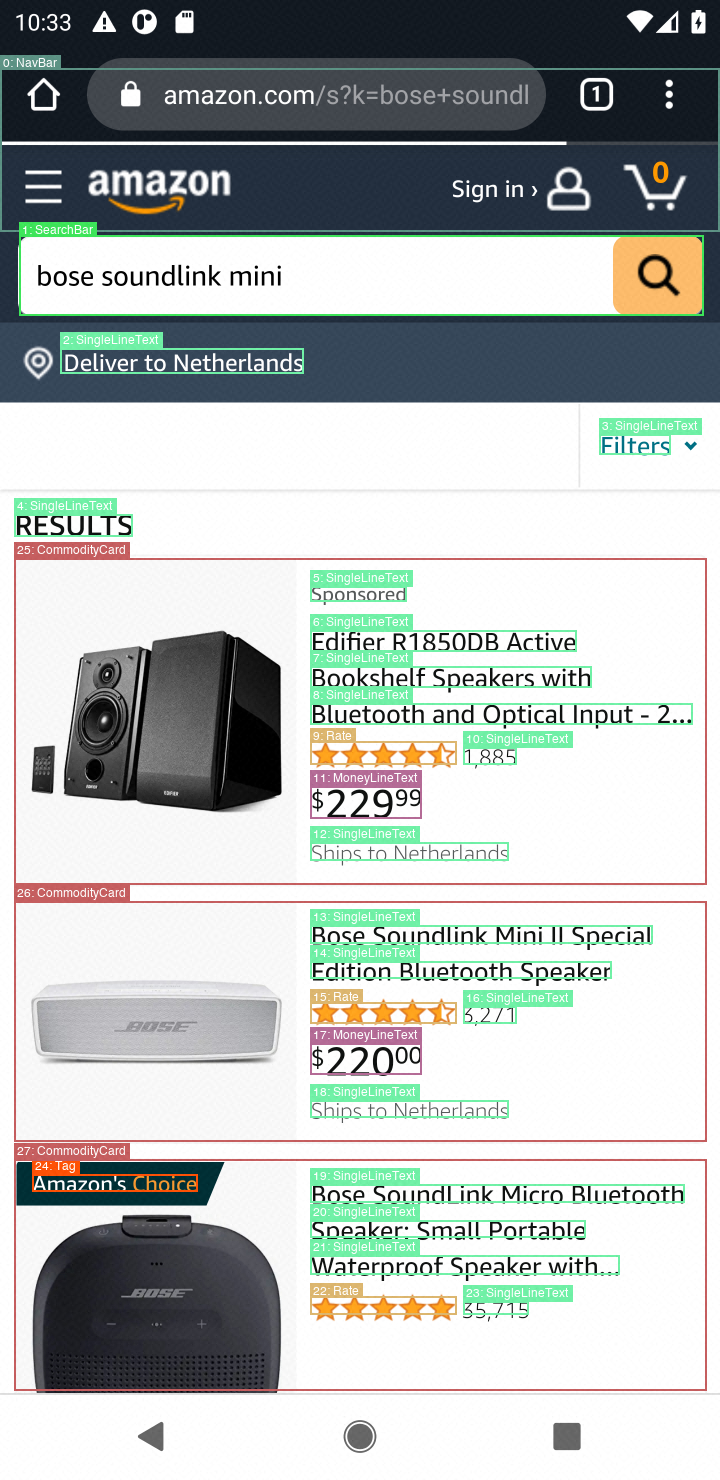}\\[-0.2em]
\scriptsize UI-Venus-Ground-72B
\end{minipage}\hfill
\begin{minipage}[t]{0.136\textwidth}
\centering
\includegraphics[width=\linewidth]{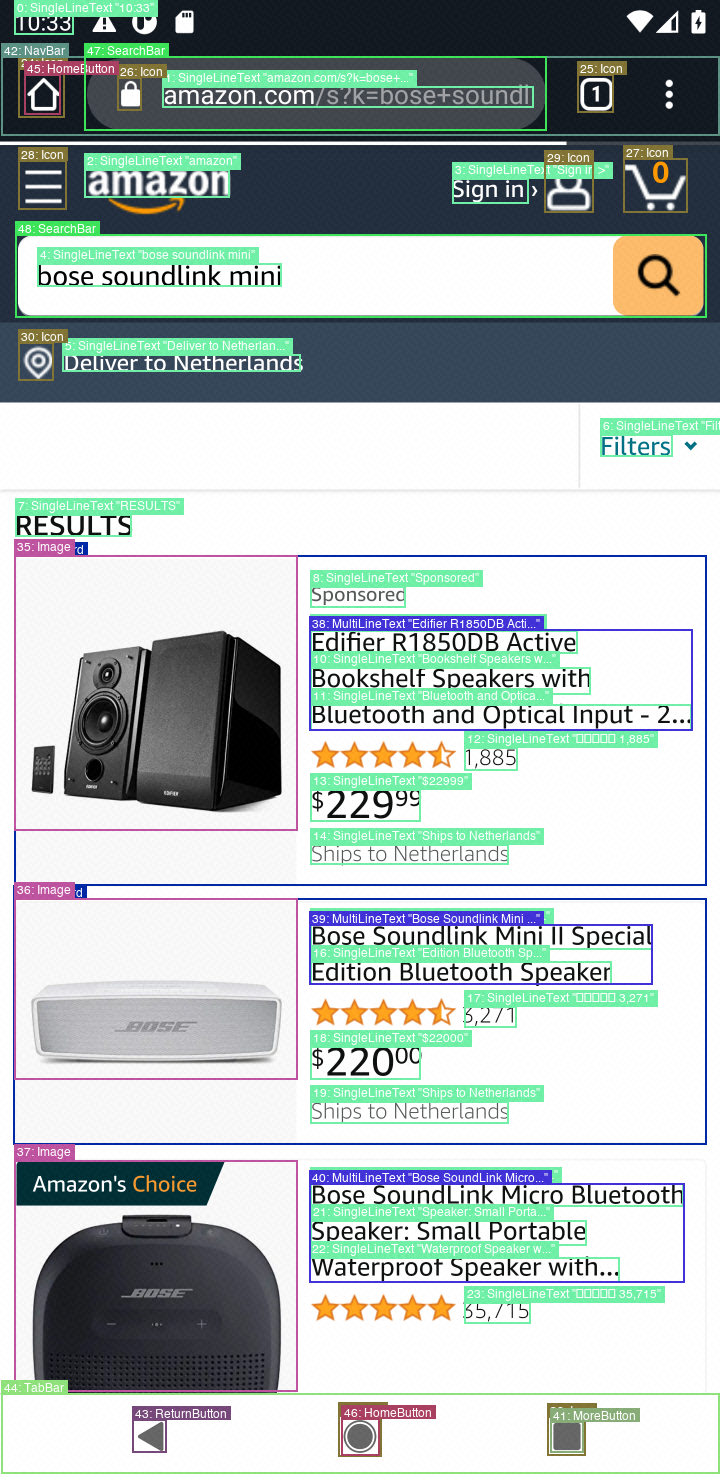}\\[-0.2em]
\scriptsize UI-UG 2-4B
\end{minipage}

\vspace{0.45em}

\begin{minipage}[t]{0.136\textwidth}
\centering
\includegraphics[width=\linewidth]{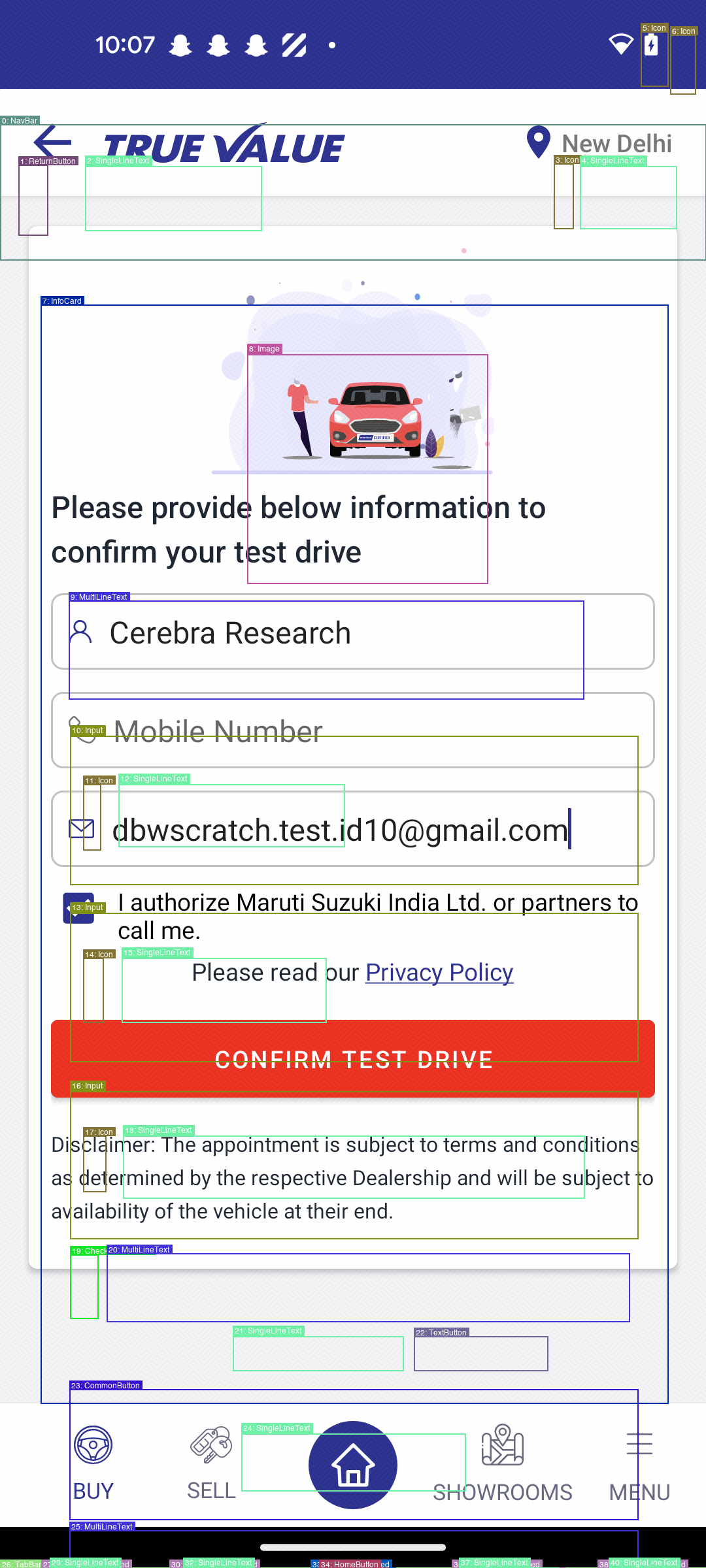}\\[-0.2em]
\scriptsize GPT-5.2
\end{minipage}\hfill
\begin{minipage}[t]{0.136\textwidth}
\centering
\includegraphics[width=\linewidth]{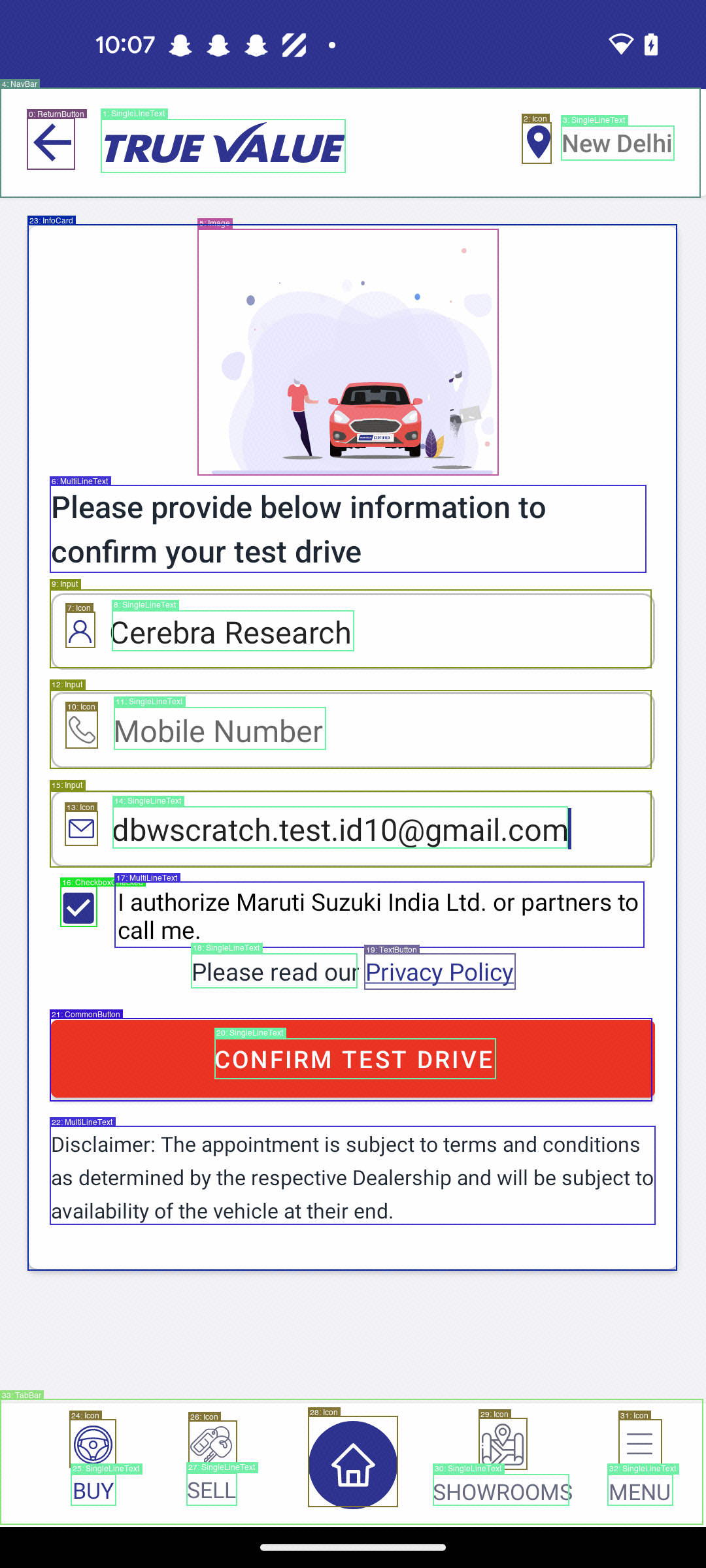}\\[-0.2em]
\scriptsize Gemini~3.1~Pro
\end{minipage}\hfill
\begin{minipage}[t]{0.136\textwidth}
\centering
\includegraphics[width=\linewidth]{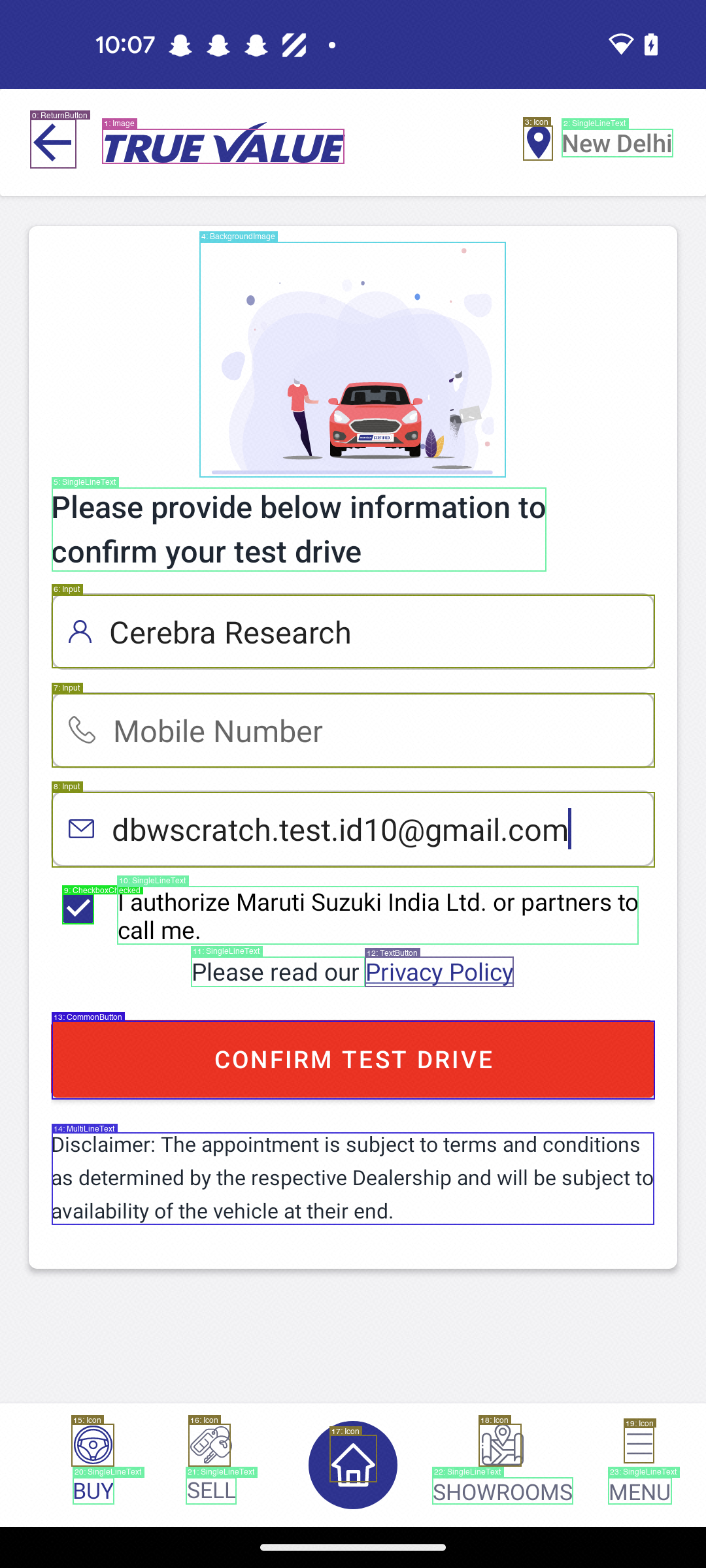}\\[-0.2em]
\scriptsize Claude~Sonnet~4.5
\end{minipage}\hfill
\begin{minipage}[t]{0.136\textwidth}
\centering
\includegraphics[width=\linewidth]{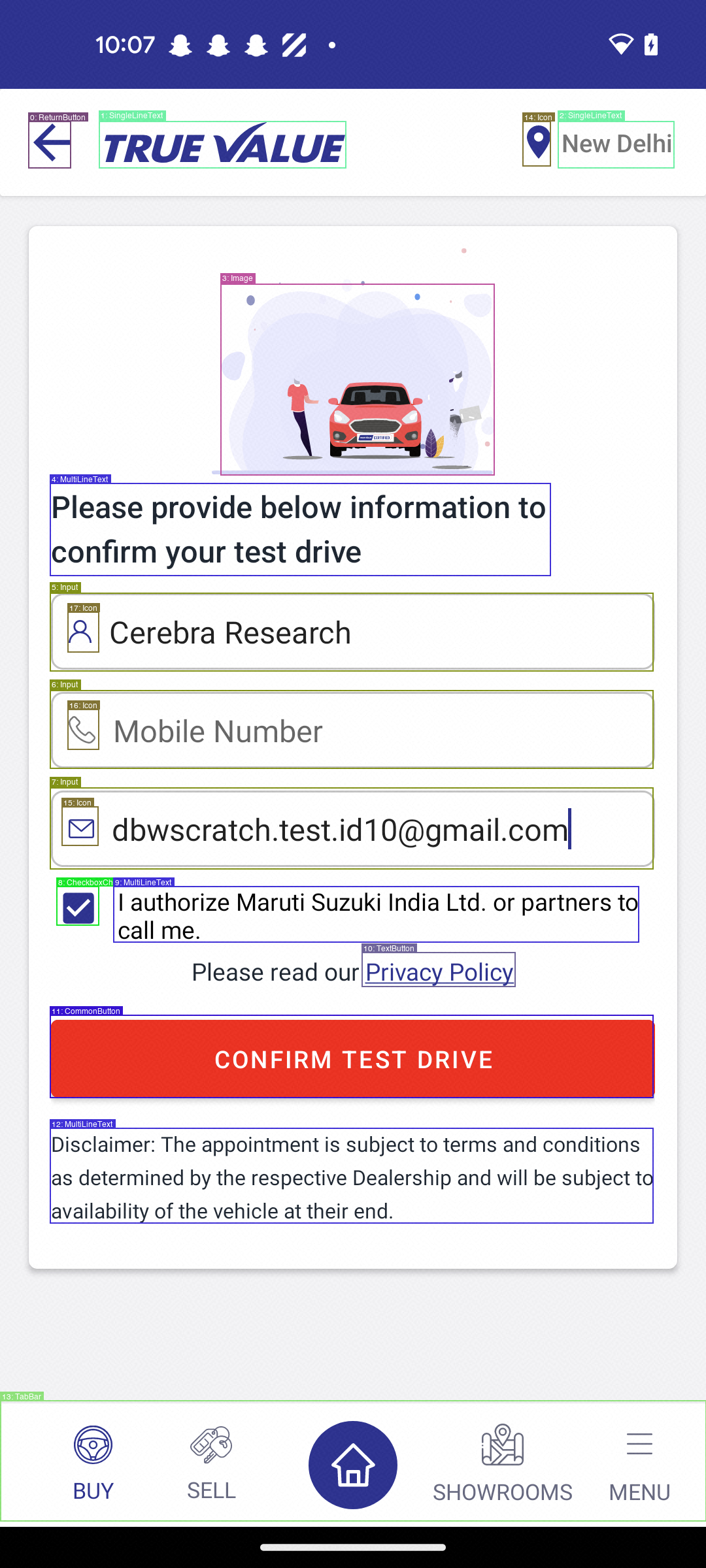}\\[-0.2em]
\scriptsize Qwen3-VL-235B-A22B
\end{minipage}\hfill
\begin{minipage}[t]{0.136\textwidth}
\centering
\includegraphics[width=\linewidth]{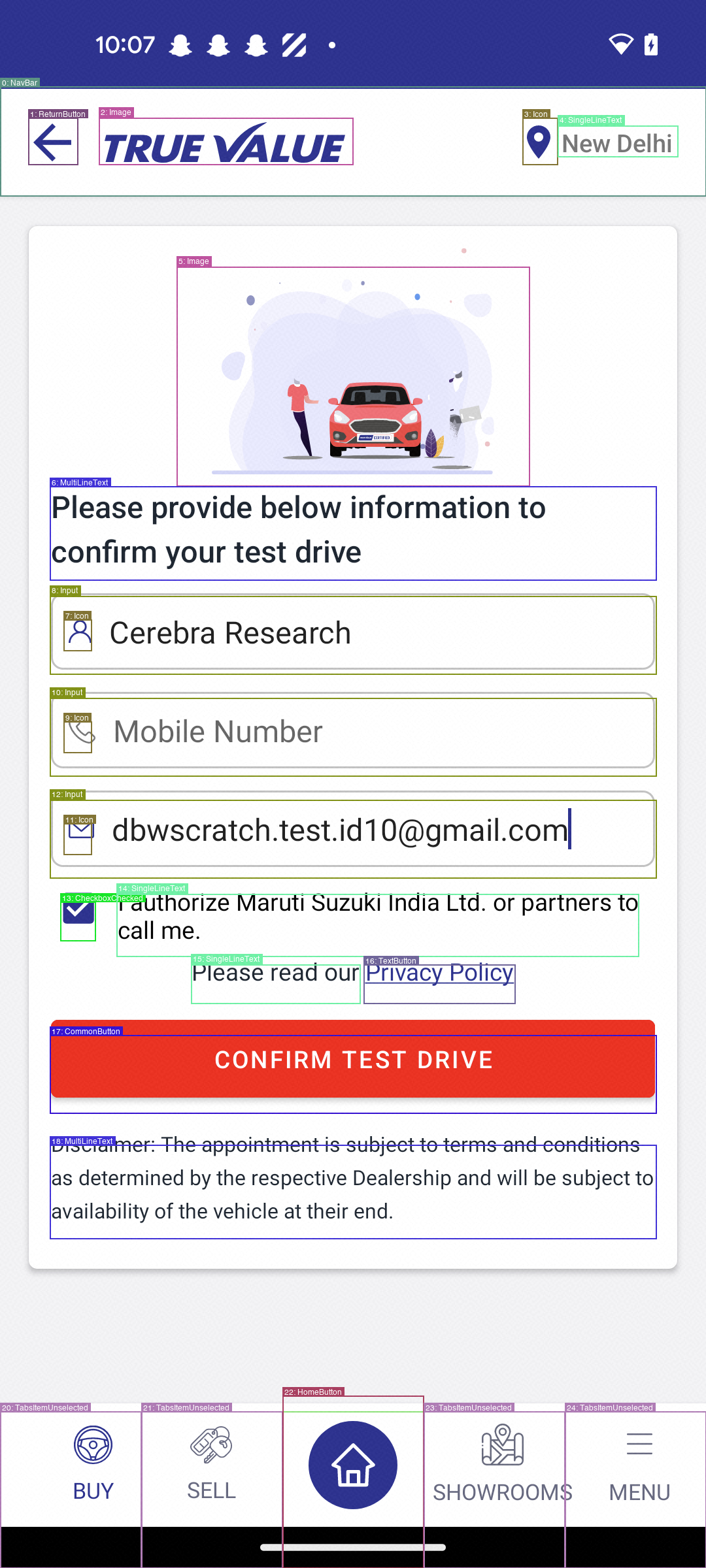}\\[-0.2em]
\scriptsize Kimi~K2.5
\end{minipage}\hfill
\begin{minipage}[t]{0.136\textwidth}
\centering
\includegraphics[width=\linewidth]{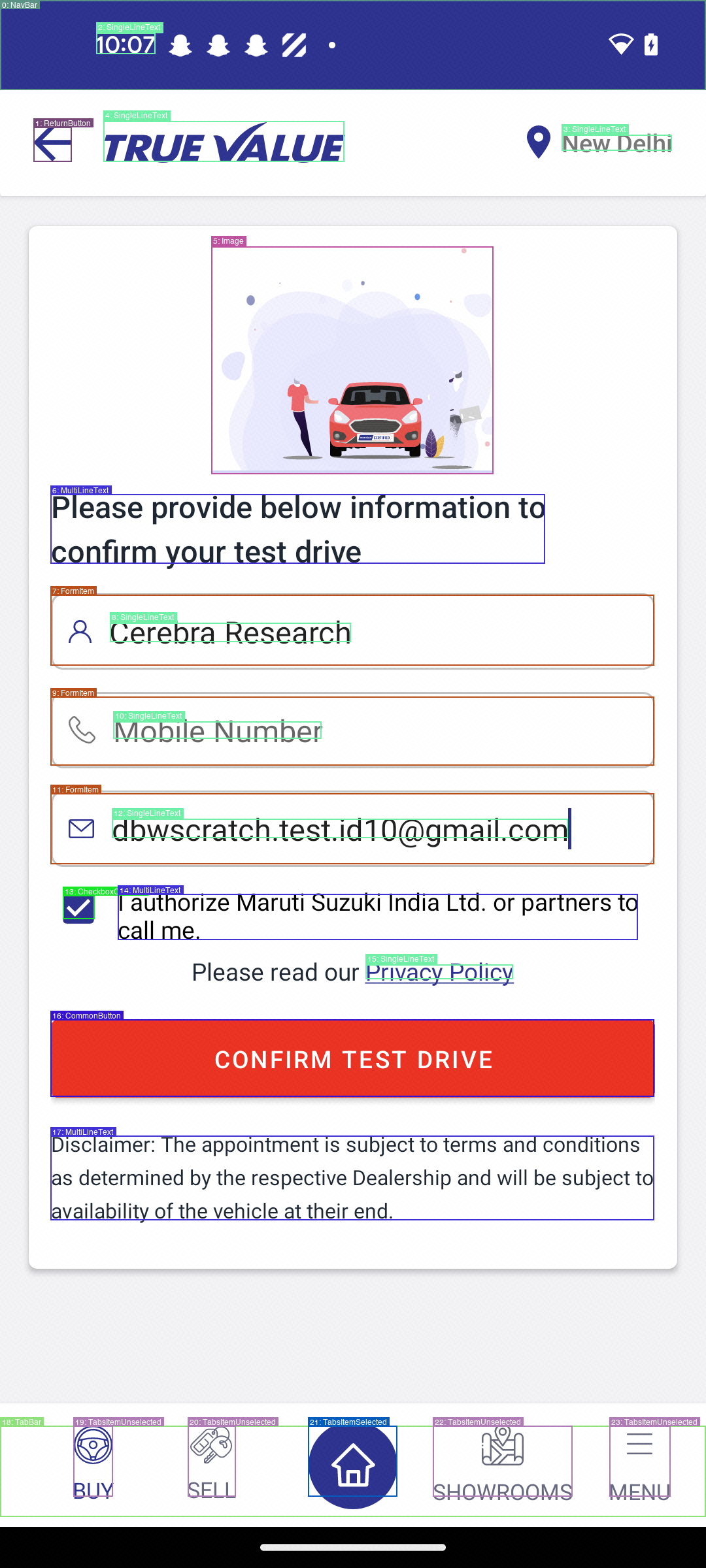}\\[-0.2em]
\scriptsize UI-Venus-Ground-72B
\end{minipage}\hfill
\begin{minipage}[t]{0.136\textwidth}
\centering
\includegraphics[width=\linewidth]{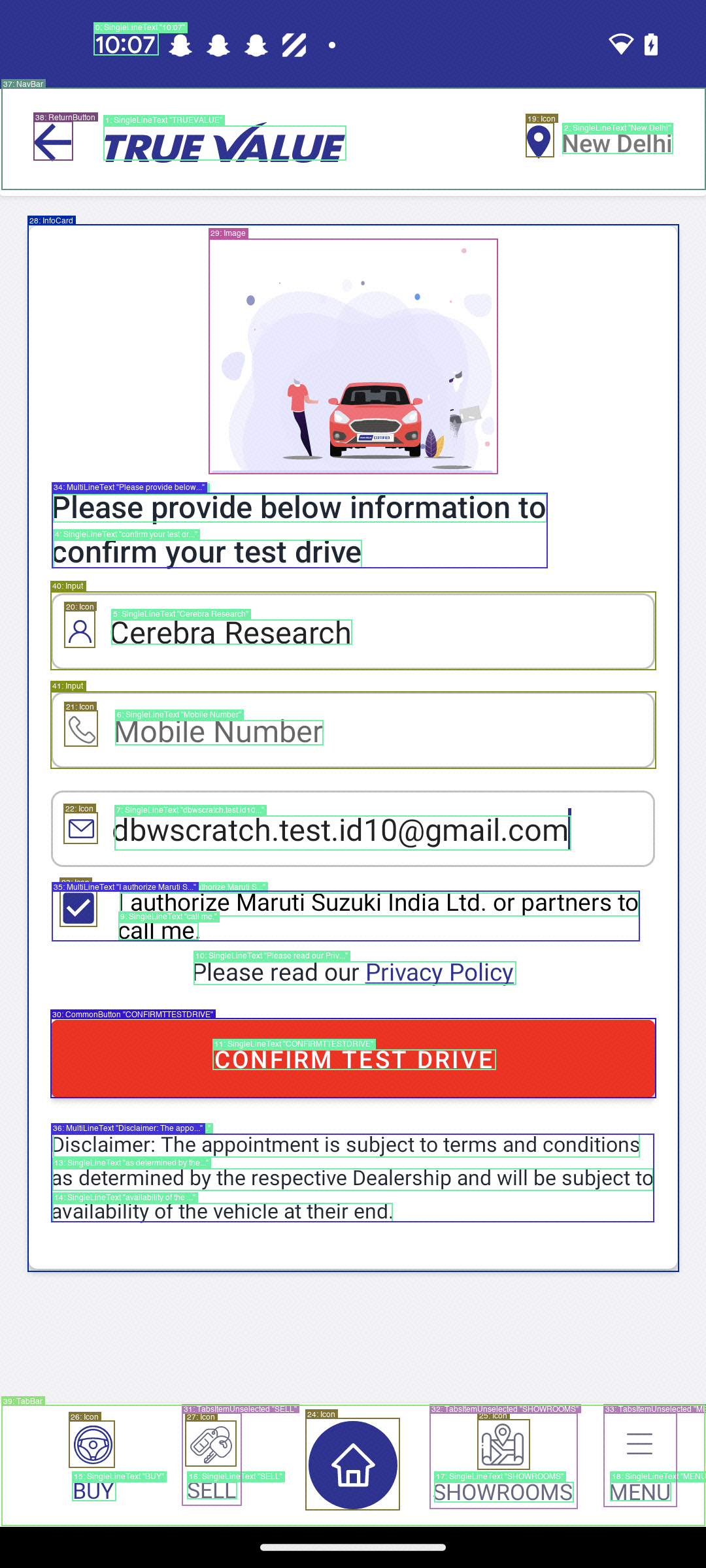}\\[-0.2em]
\scriptsize UI-UG 2-4B
\end{minipage}
\caption{Visual comparison of grounding results on two dense UI examples (2$\times$7). Each row follows the same model order: GPT-5.2, Gemini~3.1~Pro, Claude~Sonnet~4.5, Qwen3-VL-235B-A22B, Kimi~K2.5, UI-Venus-Ground-72B, and UI-UG 2-4B (ours).}
\label{fig:grounding_compare_dense}
\end{figure}

Figure~\ref{fig:grounding_compare_dense} presents qualitative comparisons of grounding results across representative models on two dense UI cases. Our model demonstrates stronger category recognition and tighter bounding boxes across both examples.

From the qualitative comparisons, we observe three consistent patterns. \textbf{First}, several baseline models are weaker in \emph{bounding-box localization precision}: for example, GPT-5.2 often exhibits noticeable spatial offsets, while other models also show imperfect alignment on narrow regions such as top navigation bars and bottom icon areas. \textbf{Second}, baseline predictions show more \emph{missed detections}, including both high-level composite components (e.g., outer boxes of product cards) and fine-grained inner elements (e.g., small icons inside cards). \textbf{Third}, there are recurring \emph{category confusions} for function-oriented controls, such as failing to correctly detect or classify return/back buttons. These failure modes are reduced in UI-UG 2-4B, which yields more complete and structurally consistent grounding results.

\subsection{Grounding: GUI Grounding Per-Category Analysis}
\label{app:gui_category}

Table~\ref{tab:gui_category} reports UI-UG 2-4B's GUI grounding results on both ScreenSpot-V1 Mobile and ScreenSpot-V2 Mobile. We group the results into Icon and Text categories and report the corresponding Overall accuracy per dataset. Text is consistently slightly higher than Icon, and the overall accuracy is 0.902 on V1 and 0.929 on V2.

\begin{table}[htbp]
\centering
\small
\caption{UI-UG 2-4B GUI grounding accuracy on ScreenSpot mobile benchmarks by element type.}
\label{tab:gui_category}
\begin{tabular}{lccc}
\toprule
\textbf{Dataset} & \textbf{Icon} & \textbf{Text} & \textbf{Overall} \\
\midrule
ScreenSpot-V1 Mobile & 0.887 & 0.916 & 0.902 \\
ScreenSpot-V2 Mobile & 0.912 & 0.944 & 0.929 \\
\bottomrule
\end{tabular}
\end{table}

\begin{figure}[htbp]
\centering
\includegraphics[width=0.95\linewidth,trim=0 200 0 0,clip]{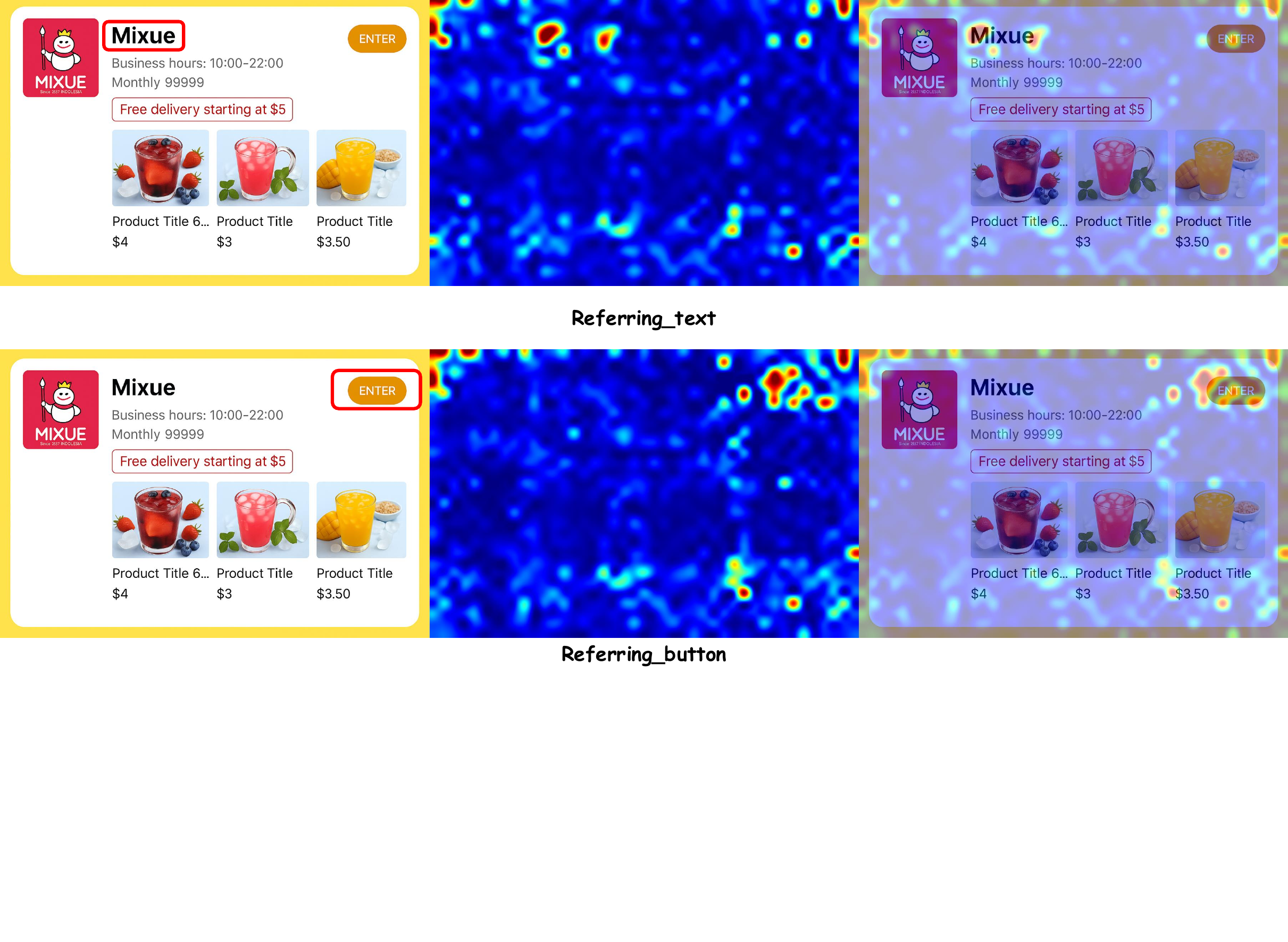}
\caption{Attention maps demonstrate task-dependent focus, with highlights concentrated on the corresponding target regions.}
\label{fig:attn_case}
\end{figure}

\subsection{Attention Visualization Analysis}
\label{app:attn_referring}

In the main paper, we already presented an attention-mechanism experiment. To provide a clearer and more interpretable view of that behavior, we further conduct attention visualization analysis in this appendix.

Figure~\ref{fig:attn_case} focuses on \textbf{referring-task} cases and shows a clear task-conditioned attention shift. In referring-text examples, high-response regions are concentrated around the target text spans; in referring-button examples, attention moves to the corresponding interactive button areas. This qualitative evidence supports the same conclusion as our quantitative results: the model can adapt its visual focus according to referring intent, which is critical for stable category grounding and precise target localization in UI understanding.

\FloatBarrier

%% file: appendices/07_generation.tex
\section{Additional Generation Results}
\label{app:generation_results}

This section provides supporting analyses for generation quality, judge reliability, and qualitative comparisons beyond the main-paper table.

\subsection{LLM-as-a-Judge Reliability Analysis}
\label{app:judge_validity}

We evaluate judge reliability on 100 generated UI cases. For each case, three professional UI designers provide scores on four dimensions (Color \& Aesthetic, Structure \& Alignment, Textual \& Content Consistency, and Interface Usability \& Interactivity), and the averaged human score is used as the reference value. The designers were recruited based on prior mobile UI design experience, scored anonymized model outputs independently under the same rubric as the MLLM judges, and were compensated according to their standard annotation rate. Table~\ref{tab:judge_reliability} compares four MLLM judges (Qwen3-VL-235B-A22B, GPT-5.2, Claude~Sonnet~4.5, and Gemini~3.1~Pro) against this reference using score deviation ($\Delta$) and variance.

\begin{table}[htbp]
\centering
\scriptsize
\caption{LLM-as-a-Judge reliability on 100 samples. $\Delta$ is the mean absolute deviation from human reference (lower is better). Variance is computed on residuals against the human reference.}
\label{tab:judge_reliability}
\setlength{\tabcolsep}{2pt}
\begin{tabular}{lcccccc}
\toprule
\textbf{Judge Model} & \textbf{$\Delta_{\text{Col}}$} & \textbf{$\Delta_{\text{Str}}$} & \textbf{$\Delta_{\text{Txt}}$} & \textbf{$\Delta_{\text{Int}}$} & \textbf{$\Delta_{\text{Overall}}$} & \textbf{Variance} \\
\midrule
Gemini~3.1~Pro & \textbf{0.07} & \textbf{0.08} & \textbf{0.09} & \textbf{0.08} & \textbf{0.08} & \textbf{0.032} \\
Qwen3-VL-235B-A22B & 0.15 & 0.17 & 0.13 & 0.20 & 0.17 & 0.046 \\
GPT-5.2 & 0.12 & 0.11 & 0.14 & 0.12 & 0.13 & 0.049 \\
Claude~Sonnet~4.5 & 0.21 & 0.24 & 0.23 & 0.26 & 0.24 & 0.050 \\
\bottomrule
\end{tabular}
\end{table}

Overall, the four judges show similar reliability ranges. Their absolute deviations are moderate and residual variance remains controlled, indicating that MLLM-based aesthetic evaluation is reasonably stable. In raw-score inspection, judges tend to assign slightly higher scores than human references. Gemini~3.1~Pro performs best. GPT-5.2 and Qwen3-VL-235B-A22B are close in overall deviation with comparable variance, while Claude~Sonnet~4.5 shows a larger average deviation in this evaluation.

Although Gemini~3.1~Pro obtains the best evaluation performance, we finally use Qwen3-VL-235B-A22B for GRPO for practical repeated evaluation through the separately provisioned judge service.

%% file: appendices/08_efficiency.tex
\clearpage
\section{Inference Efficiency}
\label{app:inference}

To complement quality results, this section reports practical latency-oriented efficiency analysis with measured values distinguished from scaling estimates.

\subsection{Inference Time Estimation for UI Generation}

To provide a size-aware efficiency reference, Table~\ref{tab:latency_1000token} reports estimated relative latency across representative open-source model scales for UI generation. We do not include closed-source API models in this comparison for two reasons: (1) the goal here is to analyze the relationship between model size and speed under controlled assumptions, and (2) API serving speed from closed-source providers is often unstable due to external service variance.

UI-UG 2-4B latency is measured on 100 sampled UI-generation cases with an average output length of 894 tokens. For larger open-source baselines, we report a size-aware estimate under a 16$\times$A100 inference setting, using \textbf{active parameters} for MoE models and normalizing UI-UG 2-4B (4B) to \textbf{1.00$\times$}. These estimates contextualize scaling trends rather than replace hardware-specific deployment benchmarking.

\begin{table}[htbp]
\centering
\footnotesize
\caption{Estimated UI generation latency using a size-based scaling factor for larger open-source baselines.}
\label{tab:latency_1000token}
\begin{tabular}{lccc}
\toprule
\textbf{Model} & \textbf{Size} & \textbf{Size-Based Factor} & \textbf{Estimated Time (s)} \\
\midrule
\textbf{UI-UG 2-4B} & 4B & 1.00$\times$ & 0.9 \\
Qwen3-VL-30B-A3B & A3B & 0.75$\times$ & 0.7 \\
UI-Venus-Ground-7B & 7B & 1.75$\times$ & 1.6 \\
Qwen3-VL-8B & 8B & 2.00$\times$ & 1.8 \\
UI-Venus-Ground-72B & 72B & 18.00$\times$ & 16.0 \\
Qwen3-VL-235B-A22B & A22B & 5.50$\times$ & 4.9 \\
\bottomrule
\end{tabular}
\end{table}

Although Qwen3-VL-30B-A3B has a much larger total parameter count than UI-UG 2-4B, its active parameter size (A3B) is smaller than the dense 4B baseline, which explains the lower estimated latency under the active-parameter approximation.

\subsection{Progressive Rendering Illustration}

\begin{figure}[!htbp]
\centering
\includegraphics[width=0.78\linewidth]{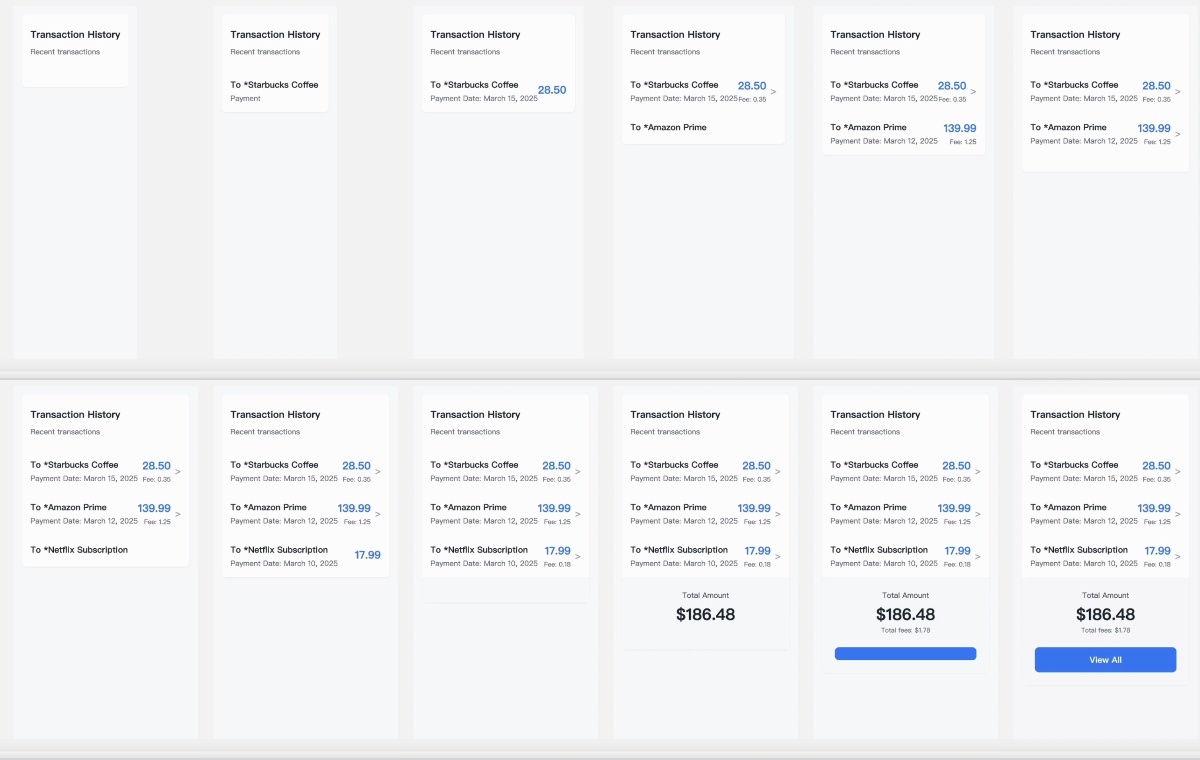}
\caption{Progressive rendering of generated UI specifications.}
\label{fig:thumbnail}
\end{figure}

Figure~\ref{fig:thumbnail} illustrates progressive rendering of generated UI specifications.

